\documentclass{article}
\newif\ifchecklist
\newif\ifcustomstyle\customstylefalse
\newif\ifconfidential\confidentialfalse

\usepackage[final,nonatbib]{neurips_2024}\checklistfalse

\usepackage[utf8]{inputenc}
\usepackage[T1]{fontenc}
\usepackage{hyperref}
\usepackage{url}
\usepackage{booktabs}
\usepackage{amsfonts}
\usepackage{nicefrac}
\usepackage{microtype}
\usepackage{color}
\usepackage{graphicx}
\usepackage{tikz}
\usepackage{amsmath,amssymb}
\usepackage{mathtools}
\usepackage[normalem]{ulem}
\usepackage{tabu}
\usepackage{multirow}
\usepackage{pgfplots,pgfplotstable}
\usepgfplotslibrary{fillbetween}
\usepackage{calc}
\usepackage{pbox}
\usepackage{algorithm}
\usepackage[noend]{algpseudocode}
\usepackage{caption}
\usepackage{afterpage}
\usepackage{subcaption}
\usepackage{listings}

\pgfplotsset{compat=1.14}
\pgfplotsset{compat/show suggested version=false}
\graphicspath{{./figures/}}
\definecolor{olive}{rgb}{0.5, 0.5, 0.0}
\definecolor{maroon}{rgb}{0.69, 0.19, 0.38}
\definecolor{celestialblue}{rgb}{0.29, 0.59, 0.82}
\definecolor{darkgreen}{rgb}{0.0, 0.6, 0.0}
\definecolor{grey}{rgb}{0.5,0.5,0.5}
\definecolor{darkblue}{rgb}{0.19, 0.19, 0.62}
\definecolor{silver}{rgb}{0.7,0.7,0.7}
\definecolor{darkcyan}{rgb}{0.0, 0.55, 0.55}

\newcommand{\DINO}{FD\textsubscript{DINOv2}}

\newcommand{\norm}[1]{\left\lVert#1\right\rVert}

\newcommand{\tstrut}{\vphantom{$\hat{A}$}}

\newcommand{\NN}{\mathcal{N}}

\newcommand{\yy}{\boldsymbol{y}}

\newcommand{\boldzero}{\mathbf{0}}
\newcommand{\boldI}{\mathbf{I}}
\newcommand{\boldx}{\mathbf{x}}

\newcommand{\boldy}{\mathbf{y}}

\newcommand{\boldc}{\mathbf{c}}
\newcommand{\boldn}{\mathbf{n}}

\newcommand{\nablax}{\smash{\nabla_{\hspace*{-.12em}\boldx}\hspace*{-.05em}}}
\newcommand{\nablaxs}{\smash{\nabla_{\hspace*{-.12em}\boldx_\sigma}\hspace*{-.25em}}}
\newcommand{\dnet}{D_{\theta}}

\newcommand\undefcolumntype[1]{\expandafter\let\csname NC@find@#1\endcsname\relax}

\newcommand{\diff}{\mathrm{d}}

\DeclareMathOperator{\argmin}{arg\,min}

\newcommand{\guidance}{w}
\newcommand{\ema}{EMA}
\newcommand{\mainEMA}{EMA\textsubscript{m}}
\newcommand{\rudderEMA}{EMA\textsubscript{g}}

\newcommand\dmain{D_1}
\newcommand\drud{D_0}
\newcommand\dguid{D_\guidance}
\newcommand\pmain{p_1}
\newcommand\prud{p_0}
\newcommand\pguid{p_\guidance}

\newcommand{\datalabel}[1]{\scalebox{0.9}{\scriptsize #1}}
\pgfplotsset{tick label style={font=\scriptsize}}
\pgfplotsset{legend style={font=\scriptsize}}
\pgfplotsset{tick style={draw=none}}
\pgfplotsset{major grid style={gray!40}}
\pgfplotsset{every axis plot/.append style={line width=0.6pt, mark size=1.1pt}}
\pgfplotsset{legend image code/.code={\draw[mark repeat=2, mark phase=2] plot coordinates {(0mm, 0.2mm) (1.5mm, 0.2mm) (3mm, 0.2mm)};}}

\definecolor{C0}{rgb}{0.121569, 0.466667, 0.705882}
\definecolor{C1}{rgb}{1.000000, 0.498039, 0.054902}
\definecolor{C2}{rgb}{0.172549, 0.627451, 0.172549}
\definecolor{C3}{rgb}{0.839216, 0.152941, 0.156863}
\definecolor{C4}{rgb}{0.580392, 0.403922, 0.741176}
\definecolor{C5}{rgb}{0.549020, 0.337255, 0.294118}
\definecolor{C6}{rgb}{0.890196, 0.466667, 0.760784}
\definecolor{C7}{rgb}{0.498039, 0.498039, 0.498039}
\definecolor{C8}{rgb}{0.737255, 0.741176, 0.133333}
\definecolor{C9}{rgb}{0.090196, 0.745098, 0.811765}
\definecolor{C10}{rgb}{1,1,1}
\definecolor{C11}{rgb}{0,0,0}

\newcommand{\hs}[1]{\hspace{#1mm}}
\newcommand{\fillbetween}[3][]{\addplot+[name path=A, draw=none, mark=none, forget plot] #2; \addplot+[name path=B, draw=none, mark=none, forget plot] #3; \addplot[#1] fill between[of=A and B]}

\newcommand{\defdata}[2]{\expandafter\newcommand\csname data-#1\endcsname{#2}}
\newcommand{\data}[2][???]{\ifdata{#2}\rawdata{#2}\else#1\fi}
\newcommand{\ifdata}[1]{\ifcsname data-#1\endcsname}
\newcommand{\rawdata}[1]{\csname data-#1\endcsname}

\newcommand{\cidBestS}{img512-S-XS-128Mi}
\newcommand{\cidBestXXL}{img512-XXL-M-256Mi}
\newcommand{\cidBestUncond}{img512-Su-XSu-128Mi}
\newcommand{\cidBestSixtyFour}{img64-S-XS-128Mi}

\newcommand{\cidSynDropout}{img512-S-d0.05r1-S-d0.10r1-fixp0.130}
\newcommand{\cidSynInputNoise}{img512-S-i0.10r1-S-i0.20r1-fixp0.130}

\defdata{img512-S-XSu-fid}{2.26}
\defdata{img512-S-XSu-fid-precise}{2.2639}
\defdata{img512-S-XSu-fid-p}{0.025}
\defdata{img512-S-XSu-fid-r}{0.025}
\defdata{img512-S-XSu-fid-g}{1.40}
\defdata{img512-S-XSu-dino}{51.77}
\defdata{img512-S-XSu-dino-precise}{51.7667}
\defdata{img512-S-XSu-dino-p}{0.085}
\defdata{img512-S-XSu-dino-r}{0.085}
\defdata{img512-S-XSu-dino-g}{1.90}
\defdata{img512-S-XSu-g1.00-fid}{3.27}
\defdata{img512-S-XSu-g1.00-fid-precise}{3.2654}
\defdata{img512-S-XSu-g1.00-fid-p}{0.085}
\defdata{img512-S-XSu-g1.00-fid-r}{0.085}
\defdata{img512-S-XSu-g1.00-fid-g}{1.00}
\defdata{img512-S-XSu-g1.00-dino}{122.80}
\defdata{img512-S-XSu-g1.00-dino-precise}{122.7996}
\defdata{img512-S-XSu-g1.00-dino-p}{0.085}
\defdata{img512-S-XSu-g1.00-dino-r}{0.085}
\defdata{img512-S-XSu-g1.00-dino-g}{1.00}

\defdata{img512-S-XSu-glo0.28-ghi2.90-fid}{1.65}
\defdata{img512-S-XSu-glo0.28-ghi2.90-fid-precise}{1.6502}
\defdata{img512-S-XSu-glo0.28-ghi2.90-fid-p}{0.025}
\defdata{img512-S-XSu-glo0.28-ghi2.90-fid-r}{0.025}
\defdata{img512-S-XSu-glo0.28-ghi2.90-fid-g}{2.10}
\defdata{img512-S-XSu-glo0.28-ghi2.90-dino}{49.16}
\defdata{img512-S-XSu-glo0.28-ghi2.90-dino-precise}{49.1610}
\defdata{img512-S-XSu-glo0.28-ghi2.90-dino-p}{0.120}
\defdata{img512-S-XSu-glo0.28-ghi2.90-dino-r}{0.120}
\defdata{img512-S-XSu-glo0.28-ghi2.90-dino-g}{2.70}
\defdata{img512-S-XSu-glo0.28-ghi2.90-g1.00-fid}{3.35}
\defdata{img512-S-XSu-glo0.28-ghi2.90-g1.00-fid-precise}{3.3534}
\defdata{img512-S-XSu-glo0.28-ghi2.90-g1.00-fid-p}{0.085}
\defdata{img512-S-XSu-glo0.28-ghi2.90-g1.00-fid-r}{0.085}
\defdata{img512-S-XSu-glo0.28-ghi2.90-g1.00-fid-g}{1.00}
\defdata{img512-S-XSu-glo0.28-ghi2.90-g1.00-dino}{122.40}
\defdata{img512-S-XSu-glo0.28-ghi2.90-g1.00-dino-precise}{122.3982}
\defdata{img512-S-XSu-glo0.28-ghi2.90-g1.00-dino-p}{0.085}
\defdata{img512-S-XSu-glo0.28-ghi2.90-g1.00-dino-r}{0.085}
\defdata{img512-S-XSu-glo0.28-ghi2.90-g1.00-dino-g}{1.00}

\defdata{img512-S-nog-fid}{2.56}
\defdata{img512-S-nog-fid-precise}{2.5560}
\defdata{img512-S-nog-fid-p}{0.120}
\defdata{img512-S-nog-fid-r}{0.120}
\defdata{img512-S-nog-fid-g}{1.00}
\defdata{img512-S-nog-dino}{68.43}
\defdata{img512-S-nog-dino-precise}{68.4308}
\defdata{img512-S-nog-dino-p}{0.195}
\defdata{img512-S-nog-dino-r}{0.195}
\defdata{img512-S-nog-dino-g}{1.00}
\defdata{img512-S-nog-g1.00-fid}{2.56}
\defdata{img512-S-nog-g1.00-fid-precise}{2.5560}
\defdata{img512-S-nog-g1.00-fid-p}{0.120}
\defdata{img512-S-nog-g1.00-fid-r}{0.120}
\defdata{img512-S-nog-g1.00-fid-g}{1.00}
\defdata{img512-S-nog-g1.00-dino}{68.43}
\defdata{img512-S-nog-g1.00-dino-precise}{68.4308}
\defdata{img512-S-nog-g1.00-dino-p}{0.195}
\defdata{img512-S-nog-g1.00-dino-r}{0.195}
\defdata{img512-S-nog-g1.00-dino-g}{1.00}

\defdata{img512-XXL-XSu-fid}{1.80}
\defdata{img512-XXL-XSu-fid-precise}{1.7964}
\defdata{img512-XXL-XSu-fid-p}{0.025}
\defdata{img512-XXL-XSu-fid-r}{0.025}
\defdata{img512-XXL-XSu-fid-g}{1.15}
\defdata{img512-XXL-XSu-dino}{32.67}
\defdata{img512-XXL-XSu-dino-precise}{32.6703}
\defdata{img512-XXL-XSu-dino-p}{0.015}
\defdata{img512-XXL-XSu-dino-r}{0.015}
\defdata{img512-XXL-XSu-dino-g}{1.85}
\defdata{img512-XXL-XSu-g1.00-fid}{2.09}
\defdata{img512-XXL-XSu-g1.00-fid-precise}{2.0897}
\defdata{img512-XXL-XSu-g1.00-fid-p}{0.045}
\defdata{img512-XXL-XSu-g1.00-fid-r}{0.045}
\defdata{img512-XXL-XSu-g1.00-fid-g}{1.00}
\defdata{img512-XXL-XSu-g1.00-dino}{69.42}
\defdata{img512-XXL-XSu-g1.00-dino-precise}{69.4187}
\defdata{img512-XXL-XSu-g1.00-dino-p}{0.045}
\defdata{img512-XXL-XSu-g1.00-dino-r}{0.045}
\defdata{img512-XXL-XSu-g1.00-dino-g}{1.00}

\defdata{img512-XXL-nog-fid}{1.92}
\defdata{img512-XXL-nog-fid-precise}{1.9169}
\defdata{img512-XXL-nog-fid-p}{0.075}
\defdata{img512-XXL-nog-fid-r}{0.075}
\defdata{img512-XXL-nog-fid-g}{1.00}
\defdata{img512-XXL-nog-dino}{42.12}
\defdata{img512-XXL-nog-dino-precise}{42.1155}
\defdata{img512-XXL-nog-dino-p}{0.145}
\defdata{img512-XXL-nog-dino-r}{0.145}
\defdata{img512-XXL-nog-dino-g}{1.00}
\defdata{img512-XXL-nog-g1.00-fid}{1.92}
\defdata{img512-XXL-nog-g1.00-fid-precise}{1.9169}
\defdata{img512-XXL-nog-g1.00-fid-p}{0.075}
\defdata{img512-XXL-nog-g1.00-fid-r}{0.075}
\defdata{img512-XXL-nog-g1.00-fid-g}{1.00}
\defdata{img512-XXL-nog-g1.00-dino}{42.12}
\defdata{img512-XXL-nog-g1.00-dino-precise}{42.1155}
\defdata{img512-XXL-nog-g1.00-dino-p}{0.145}
\defdata{img512-XXL-nog-g1.00-dino-r}{0.145}
\defdata{img512-XXL-nog-g1.00-dino-g}{1.00}

\defdata{img512-S-XS-128Mi-sameEMA-fid}{1.53}
\defdata{img512-S-XS-128Mi-sameEMA-fid-precise}{1.5342}
\defdata{img512-S-XS-128Mi-sameEMA-fid-p}{0.050}
\defdata{img512-S-XS-128Mi-sameEMA-fid-r}{0.050}
\defdata{img512-S-XS-128Mi-sameEMA-fid-g}{1.95}
\defdata{img512-S-XS-128Mi-sameEMA-dino}{40.81}
\defdata{img512-S-XS-128Mi-sameEMA-dino-precise}{40.8147}
\defdata{img512-S-XS-128Mi-sameEMA-dino-p}{0.115}
\defdata{img512-S-XS-128Mi-sameEMA-dino-r}{0.115}
\defdata{img512-S-XS-128Mi-sameEMA-dino-g}{2.25}

\defdata{img512-S-XS-128Mi-fid}{1.34}
\defdata{img512-S-XS-128Mi-fid-precise}{1.3397}
\defdata{img512-S-XS-128Mi-fid-p}{0.070}
\defdata{img512-S-XS-128Mi-fid-r}{0.125}
\defdata{img512-S-XS-128Mi-fid-g}{2.10}
\defdata{img512-S-XS-128Mi-dino}{36.67}
\defdata{img512-S-XS-128Mi-dino-precise}{36.6668}
\defdata{img512-S-XS-128Mi-dino-p}{0.120}
\defdata{img512-S-XS-128Mi-dino-r}{0.165}
\defdata{img512-S-XS-128Mi-dino-g}{2.45}
\defdata{img512-S-XS-128Mi-g1.00-fid}{2.60}
\defdata{img512-S-XS-128Mi-g1.00-fid-precise}{2.6048}
\defdata{img512-S-XS-128Mi-g1.00-fid-p}{0.120}
\defdata{img512-S-XS-128Mi-g1.00-fid-r}{0.165}
\defdata{img512-S-XS-128Mi-g1.00-fid-g}{1.00}
\defdata{img512-S-XS-128Mi-g1.00-dino}{96.30}
\defdata{img512-S-XS-128Mi-g1.00-dino-precise}{96.2984}
\defdata{img512-S-XS-128Mi-g1.00-dino-p}{0.120}
\defdata{img512-S-XS-128Mi-g1.00-dino-r}{0.165}
\defdata{img512-S-XS-128Mi-g1.00-dino-g}{1.00}

\defdata{img512-XXL-M-256Mi-fid}{1.25}
\defdata{img512-XXL-M-256Mi-fid-precise}{1.2503}
\defdata{img512-XXL-M-256Mi-fid-p}{0.075}
\defdata{img512-XXL-M-256Mi-fid-r}{0.155}
\defdata{img512-XXL-M-256Mi-fid-g}{2.05}
\defdata{img512-XXL-M-256Mi-dino}{24.18}
\defdata{img512-XXL-M-256Mi-dino-precise}{24.1792}
\defdata{img512-XXL-M-256Mi-dino-p}{0.130}
\defdata{img512-XXL-M-256Mi-dino-r}{0.205}
\defdata{img512-XXL-M-256Mi-dino-g}{2.30}

\defdata{img512-S-XSu-glo0.60-ghi5.00-fid}{1.88}
\defdata{img512-S-XSu-glo0.60-ghi5.00-fid-precise}{1.8780}
\defdata{img512-S-XSu-glo0.60-ghi5.00-fid-p}{0.030}
\defdata{img512-S-XSu-glo0.60-ghi5.00-fid-r}{0.030}
\defdata{img512-S-XSu-glo0.60-ghi5.00-fid-g}{1.75}
\defdata{img512-S-XSu-glo0.60-ghi5.00-dino}{46.26}
\defdata{img512-S-XSu-glo0.60-ghi5.00-dino-precise}{46.2560}
\defdata{img512-S-XSu-glo0.60-ghi5.00-dino-p}{0.085}
\defdata{img512-S-XSu-glo0.60-ghi5.00-dino-r}{0.085}
\defdata{img512-S-XSu-glo0.60-ghi5.00-dino-g}{3.20}
\defdata{img512-S-XSu-glo0.60-ghi5.00-g1.00-fid}{3.33}
\defdata{img512-S-XSu-glo0.60-ghi5.00-g1.00-fid-precise}{3.3306}
\defdata{img512-S-XSu-glo0.60-ghi5.00-g1.00-fid-p}{0.085}
\defdata{img512-S-XSu-glo0.60-ghi5.00-g1.00-fid-r}{0.085}
\defdata{img512-S-XSu-glo0.60-ghi5.00-g1.00-fid-g}{1.00}
\defdata{img512-S-XSu-glo0.60-ghi5.00-g1.00-dino}{122.36}
\defdata{img512-S-XSu-glo0.60-ghi5.00-g1.00-dino-precise}{122.3640}
\defdata{img512-S-XSu-glo0.60-ghi5.00-g1.00-dino-p}{0.085}
\defdata{img512-S-XSu-glo0.60-ghi5.00-g1.00-dino-r}{0.085}
\defdata{img512-S-XSu-glo0.60-ghi5.00-g1.00-dino-g}{1.00}

\defdata{img512-S-S-128Mi-fid}{1.51}
\defdata{img512-S-S-128Mi-fid-precise}{1.5085}
\defdata{img512-S-S-128Mi-fid-p}{0.090}
\defdata{img512-S-S-128Mi-fid-r}{0.130}
\defdata{img512-S-S-128Mi-fid-g}{2.20}
\defdata{img512-S-S-128Mi-dino}{42.27}
\defdata{img512-S-S-128Mi-dino-precise}{42.2742}
\defdata{img512-S-S-128Mi-dino-p}{0.130}
\defdata{img512-S-S-128Mi-dino-r}{0.170}
\defdata{img512-S-S-128Mi-dino-g}{2.55}
\defdata{img512-S-S-128Mi-g1.00-fid}{3.17}
\defdata{img512-S-S-128Mi-g1.00-fid-precise}{3.1675}
\defdata{img512-S-S-128Mi-g1.00-fid-p}{0.090}
\defdata{img512-S-S-128Mi-g1.00-fid-r}{0.130}
\defdata{img512-S-S-128Mi-g1.00-fid-g}{1.00}
\defdata{img512-S-S-128Mi-g1.00-dino}{118.66}
\defdata{img512-S-S-128Mi-g1.00-dino-precise}{118.6642}
\defdata{img512-S-S-128Mi-g1.00-dino-p}{0.090}
\defdata{img512-S-S-128Mi-g1.00-dino-r}{0.130}
\defdata{img512-S-S-128Mi-g1.00-dino-g}{1.00}

\defdata{img512-S-XS-fid}{2.13}
\defdata{img512-S-XS-fid-precise}{2.1307}
\defdata{img512-S-XS-fid-p}{0.120}
\defdata{img512-S-XS-fid-r}{0.160}
\defdata{img512-S-XS-fid-g}{1.80}
\defdata{img512-S-XS-dino}{59.89}
\defdata{img512-S-XS-dino-precise}{59.8858}
\defdata{img512-S-XS-dino-p}{0.140}
\defdata{img512-S-XS-dino-r}{0.085}
\defdata{img512-S-XS-dino-g}{1.90}

\defdata{img512-S-w0.030s0-S-w0.040s0-fixp0.130-fid}{2.64}
\defdata{img512-S-w0.030s0-S-w0.040s0-fixp0.130-fid-precise}{2.6450}
\defdata{img512-S-w0.030s0-S-w0.040s0-fixp0.130-fid-p}{0.130}
\defdata{img512-S-w0.030s0-S-w0.040s0-fixp0.130-fid-r}{0.130}
\defdata{img512-S-w0.030s0-S-w0.040s0-fixp0.130-fid-g}{3.05}
\defdata{img512-S-w0.030s0-S-w0.040s0-fixp0.130-dino}{82.38}
\defdata{img512-S-w0.030s0-S-w0.040s0-fixp0.130-dino-precise}{82.3787}
\defdata{img512-S-w0.030s0-S-w0.040s0-fixp0.130-dino-p}{0.130}
\defdata{img512-S-w0.030s0-S-w0.040s0-fixp0.130-dino-r}{0.130}
\defdata{img512-S-w0.030s0-S-w0.040s0-fixp0.130-dino-g}{3.00}
\defdata{img512-S-w0.030s0-S-w0.040s0-fixp0.130-g0.00-fid}{3.86}
\defdata{img512-S-w0.030s0-S-w0.040s0-fixp0.130-g0.00-fid-precise}{3.8571}
\defdata{img512-S-w0.030s0-S-w0.040s0-fixp0.130-g0.00-fid-p}{0.130}
\defdata{img512-S-w0.030s0-S-w0.040s0-fixp0.130-g0.00-fid-r}{0.130}
\defdata{img512-S-w0.030s0-S-w0.040s0-fixp0.130-g0.00-fid-g}{0.00}
\defdata{img512-S-w0.030s0-S-w0.040s0-fixp0.130-g0.00-dino}{107.80}
\defdata{img512-S-w0.030s0-S-w0.040s0-fixp0.130-g0.00-dino-precise}{107.7993}
\defdata{img512-S-w0.030s0-S-w0.040s0-fixp0.130-g0.00-dino-p}{0.130}
\defdata{img512-S-w0.030s0-S-w0.040s0-fixp0.130-g0.00-dino-r}{0.130}
\defdata{img512-S-w0.030s0-S-w0.040s0-fixp0.130-g0.00-dino-g}{0.00}
\defdata{img512-S-w0.030s0-S-w0.040s0-fixp0.130-g1.00-fid}{3.15}
\defdata{img512-S-w0.030s0-S-w0.040s0-fixp0.130-g1.00-fid-precise}{3.1498}
\defdata{img512-S-w0.030s0-S-w0.040s0-fixp0.130-g1.00-fid-p}{0.130}
\defdata{img512-S-w0.030s0-S-w0.040s0-fixp0.130-g1.00-fid-r}{0.130}
\defdata{img512-S-w0.030s0-S-w0.040s0-fixp0.130-g1.00-fid-g}{1.00}
\defdata{img512-S-w0.030s0-S-w0.040s0-fixp0.130-g1.00-dino}{97.20}
\defdata{img512-S-w0.030s0-S-w0.040s0-fixp0.130-g1.00-dino-precise}{97.2022}
\defdata{img512-S-w0.030s0-S-w0.040s0-fixp0.130-g1.00-dino-p}{0.130}
\defdata{img512-S-w0.030s0-S-w0.040s0-fixp0.130-g1.00-dino-r}{0.130}
\defdata{img512-S-w0.030s0-S-w0.040s0-fixp0.130-g1.00-dino-g}{1.00}

\defdata{img512-S-d0.05r1-S-d0.10r1-fixp0.130-fid}{2.55}
\defdata{img512-S-d0.05r1-S-d0.10r1-fixp0.130-fid-precise}{2.5527}
\defdata{img512-S-d0.05r1-S-d0.10r1-fixp0.130-fid-p}{0.130}
\defdata{img512-S-d0.05r1-S-d0.10r1-fixp0.130-fid-r}{0.130}
\defdata{img512-S-d0.05r1-S-d0.10r1-fixp0.130-fid-g}{2.25}
\defdata{img512-S-d0.05r1-S-d0.10r1-fixp0.130-dino}{71.47}
\defdata{img512-S-d0.05r1-S-d0.10r1-fixp0.130-dino-precise}{71.4684}
\defdata{img512-S-d0.05r1-S-d0.10r1-fixp0.130-dino-p}{0.130}
\defdata{img512-S-d0.05r1-S-d0.10r1-fixp0.130-dino-r}{0.130}
\defdata{img512-S-d0.05r1-S-d0.10r1-fixp0.130-dino-g}{3.05}
\defdata{img512-S-d0.05r1-S-d0.10r1-fixp0.130-g0.00-fid}{15.00}
\defdata{img512-S-d0.05r1-S-d0.10r1-fixp0.130-g0.00-fid-precise}{15.0047}
\defdata{img512-S-d0.05r1-S-d0.10r1-fixp0.130-g0.00-fid-p}{0.130}
\defdata{img512-S-d0.05r1-S-d0.10r1-fixp0.130-g0.00-fid-r}{0.130}
\defdata{img512-S-d0.05r1-S-d0.10r1-fixp0.130-g0.00-fid-g}{0.00}
\defdata{img512-S-d0.05r1-S-d0.10r1-fixp0.130-g0.00-dino}{270.82}
\defdata{img512-S-d0.05r1-S-d0.10r1-fixp0.130-g0.00-dino-precise}{270.8229}
\defdata{img512-S-d0.05r1-S-d0.10r1-fixp0.130-g0.00-dino-p}{0.130}
\defdata{img512-S-d0.05r1-S-d0.10r1-fixp0.130-g0.00-dino-r}{0.130}
\defdata{img512-S-d0.05r1-S-d0.10r1-fixp0.130-g0.00-dino-g}{0.00}
\defdata{img512-S-d0.05r1-S-d0.10r1-fixp0.130-g1.00-fid}{4.98}
\defdata{img512-S-d0.05r1-S-d0.10r1-fixp0.130-g1.00-fid-precise}{4.9802}
\defdata{img512-S-d0.05r1-S-d0.10r1-fixp0.130-g1.00-fid-p}{0.130}
\defdata{img512-S-d0.05r1-S-d0.10r1-fixp0.130-g1.00-fid-r}{0.130}
\defdata{img512-S-d0.05r1-S-d0.10r1-fixp0.130-g1.00-fid-g}{1.00}
\defdata{img512-S-d0.05r1-S-d0.10r1-fixp0.130-g1.00-dino}{149.11}
\defdata{img512-S-d0.05r1-S-d0.10r1-fixp0.130-g1.00-dino-precise}{149.1101}
\defdata{img512-S-d0.05r1-S-d0.10r1-fixp0.130-g1.00-dino-p}{0.130}
\defdata{img512-S-d0.05r1-S-d0.10r1-fixp0.130-g1.00-dino-r}{0.130}
\defdata{img512-S-d0.05r1-S-d0.10r1-fixp0.130-g1.00-dino-g}{1.00}

\defdata{img512-S-i0.10r1-S-i0.20r1-fixp0.130-fid}{2.56}
\defdata{img512-S-i0.10r1-S-i0.20r1-fixp0.130-fid-precise}{2.5612}
\defdata{img512-S-i0.10r1-S-i0.20r1-fixp0.130-fid-p}{0.130}
\defdata{img512-S-i0.10r1-S-i0.20r1-fixp0.130-fid-r}{0.130}
\defdata{img512-S-i0.10r1-S-i0.20r1-fixp0.130-fid-g}{2.00}
\defdata{img512-S-i0.10r1-S-i0.20r1-fixp0.130-dino}{88.79}
\defdata{img512-S-i0.10r1-S-i0.20r1-fixp0.130-dino-precise}{88.7907}
\defdata{img512-S-i0.10r1-S-i0.20r1-fixp0.130-dino-p}{0.130}
\defdata{img512-S-i0.10r1-S-i0.20r1-fixp0.130-dino-r}{0.130}
\defdata{img512-S-i0.10r1-S-i0.20r1-fixp0.130-dino-g}{2.00}
\defdata{img512-S-i0.10r1-S-i0.20r1-fixp0.130-g0.00-fid}{9.73}
\defdata{img512-S-i0.10r1-S-i0.20r1-fixp0.130-g0.00-fid-precise}{9.7331}
\defdata{img512-S-i0.10r1-S-i0.20r1-fixp0.130-g0.00-fid-p}{0.130}
\defdata{img512-S-i0.10r1-S-i0.20r1-fixp0.130-g0.00-fid-r}{0.130}
\defdata{img512-S-i0.10r1-S-i0.20r1-fixp0.130-g0.00-fid-g}{0.00}
\defdata{img512-S-i0.10r1-S-i0.20r1-fixp0.130-g0.00-dino}{167.97}
\defdata{img512-S-i0.10r1-S-i0.20r1-fixp0.130-g0.00-dino-precise}{167.9702}
\defdata{img512-S-i0.10r1-S-i0.20r1-fixp0.130-g0.00-dino-p}{0.130}
\defdata{img512-S-i0.10r1-S-i0.20r1-fixp0.130-g0.00-dino-r}{0.130}
\defdata{img512-S-i0.10r1-S-i0.20r1-fixp0.130-g0.00-dino-g}{0.00}
\defdata{img512-S-i0.10r1-S-i0.20r1-fixp0.130-g1.00-fid}{3.96}
\defdata{img512-S-i0.10r1-S-i0.20r1-fixp0.130-g1.00-fid-precise}{3.9585}
\defdata{img512-S-i0.10r1-S-i0.20r1-fixp0.130-g1.00-fid-p}{0.130}
\defdata{img512-S-i0.10r1-S-i0.20r1-fixp0.130-g1.00-fid-r}{0.130}
\defdata{img512-S-i0.10r1-S-i0.20r1-fixp0.130-g1.00-fid-g}{1.00}
\defdata{img512-S-i0.10r1-S-i0.20r1-fixp0.130-g1.00-dino}{113.47}
\defdata{img512-S-i0.10r1-S-i0.20r1-fixp0.130-g1.00-dino-precise}{113.4736}
\defdata{img512-S-i0.10r1-S-i0.20r1-fixp0.130-g1.00-dino-p}{0.130}
\defdata{img512-S-i0.10r1-S-i0.20r1-fixp0.130-g1.00-dino-r}{0.130}
\defdata{img512-S-i0.10r1-S-i0.20r1-fixp0.130-g1.00-dino-g}{1.00}

\defdata{img512-S-XS-64Mi-fid}{1.40}
\defdata{img512-S-XS-64Mi-fid-precise}{1.4046}
\defdata{img512-S-XS-64Mi-fid-p}{0.040}
\defdata{img512-S-XS-64Mi-fid-r}{0.135}
\defdata{img512-S-XS-64Mi-fid-g}{2.00}
\defdata{img512-S-XS-64Mi-dino}{37.95}
\defdata{img512-S-XS-64Mi-dino-precise}{37.9546}
\defdata{img512-S-XS-64Mi-dino-p}{0.105}
\defdata{img512-S-XS-64Mi-dino-r}{0.195}
\defdata{img512-S-XS-64Mi-dino-g}{2.15}
\defdata{img512-S-XS-64Mi-g1.00-fid}{5.52}
\defdata{img512-S-XS-64Mi-g1.00-fid-precise}{5.5213}
\defdata{img512-S-XS-64Mi-g1.00-fid-p}{0.040}
\defdata{img512-S-XS-64Mi-g1.00-fid-r}{0.140}
\defdata{img512-S-XS-64Mi-g1.00-fid-g}{1.00}
\defdata{img512-S-XS-64Mi-g1.00-dino}{166.06}
\defdata{img512-S-XS-64Mi-g1.00-dino-precise}{166.0602}
\defdata{img512-S-XS-64Mi-g1.00-dino-p}{0.040}
\defdata{img512-S-XS-64Mi-g1.00-dino-r}{0.140}
\defdata{img512-S-XS-64Mi-g1.00-dino-g}{1.00}

\defdata{img512-S-d0.05r10-S-d0.10r10-fixp0.130-fid}{2.53}
\defdata{img512-S-d0.05r10-S-d0.10r10-fixp0.130-fid-precise}{2.5343}
\defdata{img512-S-d0.05r10-S-d0.10r10-fixp0.130-fid-p}{0.130}
\defdata{img512-S-d0.05r10-S-d0.10r10-fixp0.130-fid-r}{0.130}
\defdata{img512-S-d0.05r10-S-d0.10r10-fixp0.130-fid-g}{2.10}
\defdata{img512-S-d0.05r10-S-d0.10r10-fixp0.130-dino}{72.17}
\defdata{img512-S-d0.05r10-S-d0.10r10-fixp0.130-dino-precise}{72.1710}
\defdata{img512-S-d0.05r10-S-d0.10r10-fixp0.130-dino-p}{0.130}
\defdata{img512-S-d0.05r10-S-d0.10r10-fixp0.130-dino-r}{0.130}
\defdata{img512-S-d0.05r10-S-d0.10r10-fixp0.130-dino-g}{2.90}
\defdata{img512-S-d0.05r10-S-d0.10r10-fixp0.130-g0.00-fid}{15.10}
\defdata{img512-S-d0.05r10-S-d0.10r10-fixp0.130-g0.00-fid-precise}{15.1044}
\defdata{img512-S-d0.05r10-S-d0.10r10-fixp0.130-g0.00-fid-p}{0.130}
\defdata{img512-S-d0.05r10-S-d0.10r10-fixp0.130-g0.00-fid-r}{0.130}
\defdata{img512-S-d0.05r10-S-d0.10r10-fixp0.130-g0.00-fid-g}{0.00}
\defdata{img512-S-d0.05r10-S-d0.10r10-fixp0.130-g0.00-dino}{275.64}
\defdata{img512-S-d0.05r10-S-d0.10r10-fixp0.130-g0.00-dino-precise}{275.6417}
\defdata{img512-S-d0.05r10-S-d0.10r10-fixp0.130-g0.00-dino-p}{0.130}
\defdata{img512-S-d0.05r10-S-d0.10r10-fixp0.130-g0.00-dino-r}{0.130}
\defdata{img512-S-d0.05r10-S-d0.10r10-fixp0.130-g0.00-dino-g}{0.00}
\defdata{img512-S-d0.05r10-S-d0.10r10-fixp0.130-g1.00-fid}{5.07}
\defdata{img512-S-d0.05r10-S-d0.10r10-fixp0.130-g1.00-fid-precise}{5.0674}
\defdata{img512-S-d0.05r10-S-d0.10r10-fixp0.130-g1.00-fid-p}{0.130}
\defdata{img512-S-d0.05r10-S-d0.10r10-fixp0.130-g1.00-fid-r}{0.130}
\defdata{img512-S-d0.05r10-S-d0.10r10-fixp0.130-g1.00-fid-g}{1.00}
\defdata{img512-S-d0.05r10-S-d0.10r10-fixp0.130-g1.00-dino}{152.62}
\defdata{img512-S-d0.05r10-S-d0.10r10-fixp0.130-g1.00-dino-precise}{152.6220}
\defdata{img512-S-d0.05r10-S-d0.10r10-fixp0.130-g1.00-dino-p}{0.130}
\defdata{img512-S-d0.05r10-S-d0.10r10-fixp0.130-g1.00-dino-r}{0.130}
\defdata{img512-S-d0.05r10-S-d0.10r10-fixp0.130-g1.00-dino-g}{1.00}

\defdata{img512-S-XS-256Mi-fid}{1.47}
\defdata{img512-S-XS-256Mi-fid-precise}{1.4738}
\defdata{img512-S-XS-256Mi-fid-p}{0.085}
\defdata{img512-S-XS-256Mi-fid-r}{0.110}
\defdata{img512-S-XS-256Mi-fid-g}{2.20}
\defdata{img512-S-XS-256Mi-dino}{38.22}
\defdata{img512-S-XS-256Mi-dino-precise}{38.2178}
\defdata{img512-S-XS-256Mi-dino-p}{0.155}
\defdata{img512-S-XS-256Mi-dino-r}{0.195}
\defdata{img512-S-XS-256Mi-dino-g}{2.25}
\defdata{img512-S-XS-256Mi-g1.00-fid}{3.25}
\defdata{img512-S-XS-256Mi-g1.00-fid-precise}{3.2544}
\defdata{img512-S-XS-256Mi-g1.00-fid-p}{0.085}
\defdata{img512-S-XS-256Mi-g1.00-fid-r}{0.110}
\defdata{img512-S-XS-256Mi-g1.00-fid-g}{1.00}
\defdata{img512-S-XS-256Mi-g1.00-dino}{122.35}
\defdata{img512-S-XS-256Mi-g1.00-dino-precise}{122.3471}
\defdata{img512-S-XS-256Mi-g1.00-dino-p}{0.085}
\defdata{img512-S-XS-256Mi-g1.00-dino-r}{0.110}
\defdata{img512-S-XS-256Mi-g1.00-dino-g}{1.00}

\defdata{img512-S-XXS-128Mi-fid}{1.41}
\defdata{img512-S-XXS-128Mi-fid-precise}{1.4145}
\defdata{img512-S-XXS-128Mi-fid-p}{0.050}
\defdata{img512-S-XXS-128Mi-fid-r}{0.140}
\defdata{img512-S-XXS-128Mi-fid-g}{1.80}
\defdata{img512-S-XXS-128Mi-dino}{36.25}
\defdata{img512-S-XXS-128Mi-dino-precise}{36.2487}
\defdata{img512-S-XXS-128Mi-dino-p}{0.110}
\defdata{img512-S-XXS-128Mi-dino-r}{0.180}
\defdata{img512-S-XXS-128Mi-dino-g}{2.05}
\defdata{img512-S-XXS-128Mi-g1.00-fid}{5.00}
\defdata{img512-S-XXS-128Mi-g1.00-fid-precise}{4.9959}
\defdata{img512-S-XXS-128Mi-g1.00-fid-p}{0.050}
\defdata{img512-S-XXS-128Mi-g1.00-fid-r}{0.140}
\defdata{img512-S-XXS-128Mi-g1.00-fid-g}{1.00}
\defdata{img512-S-XXS-128Mi-g1.00-dino}{155.68}
\defdata{img512-S-XXS-128Mi-g1.00-dino-precise}{155.6825}
\defdata{img512-S-XXS-128Mi-g1.00-dino-p}{0.050}
\defdata{img512-S-XXS-128Mi-g1.00-dino-r}{0.140}
\defdata{img512-S-XXS-128Mi-g1.00-dino-g}{1.00}

\defdata{img64-S-XS-128Mi-fid}{1.01}
\defdata{img64-S-XS-128Mi-fid-precise}{1.0148}
\defdata{img64-S-XS-128Mi-fid-p}{0.045}
\defdata{img64-S-XS-128Mi-fid-r}{0.110}
\defdata{img64-S-XS-128Mi-fid-g}{1.70}
\defdata{img64-S-XS-128Mi-dino}{31.85}
\defdata{img64-S-XS-128Mi-dino-precise}{31.8532}
\defdata{img64-S-XS-128Mi-dino-p}{0.105}
\defdata{img64-S-XS-128Mi-dino-r}{0.175}
\defdata{img64-S-XS-128Mi-dino-g}{2.20}

\defdata{img64-S-XSu-fid}{1.48}
\defdata{img64-S-XSu-fid-precise}{1.4798}
\defdata{img64-S-XSu-fid-p}{0.030}
\defdata{img64-S-XSu-fid-r}{0.030}
\defdata{img64-S-XSu-fid-g}{1.15}
\defdata{img64-S-XSu-dino}{41.84}
\defdata{img64-S-XSu-dino-precise}{41.8435}
\defdata{img64-S-XSu-dino-p}{0.040}
\defdata{img64-S-XSu-dino-r}{0.040}
\defdata{img64-S-XSu-dino-g}{1.85}
\defdata{img64-S-XSu-g1.00-fid}{1.63}
\defdata{img64-S-XSu-g1.00-fid-precise}{1.6342}
\defdata{img64-S-XSu-g1.00-fid-p}{0.070}
\defdata{img64-S-XSu-g1.00-fid-r}{0.070}
\defdata{img64-S-XSu-g1.00-fid-g}{1.00}
\defdata{img64-S-XSu-g1.00-dino}{96.93}
\defdata{img64-S-XSu-g1.00-dino-precise}{96.9319}
\defdata{img64-S-XSu-g1.00-dino-p}{0.070}
\defdata{img64-S-XSu-g1.00-dino-r}{0.070}
\defdata{img64-S-XSu-g1.00-dino-g}{1.00}

\defdata{img64-S-nog-fid}{1.62}
\defdata{img64-S-nog-fid-precise}{1.6197}
\defdata{img64-S-nog-fid-p}{0.075}
\defdata{img64-S-nog-fid-r}{0.075}
\defdata{img64-S-nog-fid-g}{1.00}
\defdata{img64-S-nog-dino}{58.52}
\defdata{img64-S-nog-dino-precise}{58.5197}
\defdata{img64-S-nog-dino-p}{0.160}
\defdata{img64-S-nog-dino-r}{0.160}
\defdata{img64-S-nog-dino-g}{1.00}
\defdata{img64-S-nog-g1.00-fid}{1.62}
\defdata{img64-S-nog-g1.00-fid-precise}{1.6197}
\defdata{img64-S-nog-g1.00-fid-p}{0.075}
\defdata{img64-S-nog-g1.00-fid-r}{0.075}
\defdata{img64-S-nog-g1.00-fid-g}{1.00}
\defdata{img64-S-nog-g1.00-dino}{58.52}
\defdata{img64-S-nog-g1.00-dino-precise}{58.5197}
\defdata{img64-S-nog-g1.00-dino-p}{0.160}
\defdata{img64-S-nog-g1.00-dino-r}{0.160}
\defdata{img64-S-nog-g1.00-dino-g}{1.00}

\defdata{img512-Su-nog-fid}{11.67}
\defdata{img512-Su-nog-fid-precise}{11.6657}
\defdata{img512-Su-nog-fid-p}{0.145}
\defdata{img512-Su-nog-fid-r}{0.145}
\defdata{img512-Su-nog-fid-g}{1.00}
\defdata{img512-Su-nog-dino}{209.53}
\defdata{img512-Su-nog-dino-precise}{209.5333}
\defdata{img512-Su-nog-dino-p}{0.170}
\defdata{img512-Su-nog-dino-r}{0.170}
\defdata{img512-Su-nog-dino-g}{1.00}
\defdata{img512-Su-nog-g1.00-fid}{11.67}
\defdata{img512-Su-nog-g1.00-fid-precise}{11.6657}
\defdata{img512-Su-nog-g1.00-fid-p}{0.145}
\defdata{img512-Su-nog-g1.00-fid-r}{0.145}
\defdata{img512-Su-nog-g1.00-fid-g}{1.00}
\defdata{img512-Su-nog-g1.00-dino}{209.53}
\defdata{img512-Su-nog-g1.00-dino-precise}{209.5333}
\defdata{img512-Su-nog-g1.00-dino-p}{0.170}
\defdata{img512-Su-nog-g1.00-dino-r}{0.170}
\defdata{img512-Su-nog-g1.00-dino-g}{1.00}

\defdata{img512-Su-XSu-128Mi-fid}{3.86}
\defdata{img512-Su-XSu-128Mi-fid-precise}{3.8609}
\defdata{img512-Su-XSu-128Mi-fid-p}{0.070}
\defdata{img512-Su-XSu-128Mi-fid-r}{0.110}
\defdata{img512-Su-XSu-128Mi-fid-g}{2.85}
\defdata{img512-Su-XSu-128Mi-dino}{90.39}
\defdata{img512-Su-XSu-128Mi-dino-precise}{90.3914}
\defdata{img512-Su-XSu-128Mi-dino-p}{0.090}
\defdata{img512-Su-XSu-128Mi-dino-r}{0.125}
\defdata{img512-Su-XSu-128Mi-dino-g}{2.90}

\defdata{img512-S-XSu-fid-plo}{(0.025, 2.2639)}
\defdata{img512-S-XSu-fid-phi}{(0.025, 2.3233)}
\defdata{img512-S-XSu-fid-rlo}{(0.025, 2.2639)}
\defdata{img512-S-XSu-fid-rhi}{(0.025, 2.3233)}
\defdata{img512-S-XSu-fid-glo}{(1.00, 6.8217) (1.05, 5.3265) (1.10, 4.3915) (1.15, 3.5566) (1.20, 3.0280) (1.25, 2.6612) (1.30, 2.4312) (1.35, 2.2933) (1.40, 2.2639) (1.45, 2.2684) (1.50, 2.3746) (1.55, 2.4564) (1.60, 2.6135) (1.65, 2.7598) (1.70, 2.9621) (1.75, 3.2797) (1.80, 3.4798) (1.85, 3.5946) (1.90, 3.8183) (1.95, 4.0985) (2.00, 4.1994) (2.05, 4.3736) (2.10, 4.5828) (2.15, 4.7965) (2.20, 4.9635) (2.25, 5.1494) (2.30, 5.3103) (2.35, 5.5336) (2.40, 5.6326) (2.45, 5.7738) (2.50, 5.9795) (2.55, 6.0928) (2.60, 6.1809) (2.65, 6.2666) (2.70, 6.4726) (2.75, 6.5671) (2.80, 6.6130) (2.85, 6.7586) (2.90, 6.8477) (2.95, 6.8946) (3.00, 6.9765) (3.05, 7.0065) (3.10, 7.0865) (3.15, 7.1485) (3.20, 7.2247) (3.25, 7.2141) (3.30, 7.2842) (3.35, 7.2724) (3.40, 7.3673) (3.45, 7.4687) (3.50, 7.5117)}
\defdata{img512-S-XSu-fid-ghi}{(1.00, 6.8781) (1.05, 5.5622) (1.10, 4.4171) (1.15, 3.6222) (1.20, 3.0672) (1.25, 2.7133) (1.30, 2.4932) (1.35, 2.3636) (1.40, 2.3233) (1.45, 2.3435) (1.50, 2.4603) (1.55, 2.5234) (1.60, 2.7369) (1.65, 2.8597) (1.70, 3.0962) (1.75, 3.3090) (1.80, 3.5010) (1.85, 3.6260) (1.90, 3.9586) (1.95, 4.1086) (2.00, 4.4102) (2.05, 4.5624) (2.10, 4.7391) (2.15, 4.9376) (2.20, 5.1418) (2.25, 5.2897) (2.30, 5.4447) (2.35, 5.6127) (2.40, 5.8630) (2.45, 5.8486) (2.50, 6.0377) (2.55, 6.1626) (2.60, 6.2908) (2.65, 6.3413) (2.70, 6.5088) (2.75, 6.7049) (2.80, 6.7238) (2.85, 6.8452) (2.90, 6.9375) (2.95, 6.9722) (3.00, 7.0161) (3.05, 7.0818) (3.10, 7.2019) (3.15, 7.3021) (3.20, 7.2490) (3.25, 7.3551) (3.30, 7.3970) (3.35, 7.4684) (3.40, 7.5778) (3.45, 7.5404) (3.50, 7.5909)}
\defdata{img512-S-XSu-dino-plo}{(0.085, 51.7667)}
\defdata{img512-S-XSu-dino-phi}{(0.085, 52.8050)}
\defdata{img512-S-XSu-dino-rlo}{(0.085, 51.7667)}
\defdata{img512-S-XSu-dino-rhi}{(0.085, 52.8050)}
\defdata{img512-S-XSu-dino-glo}{(1.00, 122.7996) (1.05, 108.8155) (1.10, 97.6740) (1.15, 89.0150) (1.20, 81.6525) (1.25, 76.3732) (1.30, 71.0899) (1.35, 67.1530) (1.40, 63.2378) (1.45, 60.6968) (1.50, 58.6750) (1.55, 56.7224) (1.60, 55.2315) (1.65, 54.4947) (1.70, 53.3049) (1.75, 53.1616) (1.80, 52.4588) (1.85, 52.2857) (1.90, 51.7667) (1.95, 52.0720) (2.00, 52.0143) (2.05, 52.4008) (2.10, 52.7459) (2.15, 53.0474) (2.20, 53.5963) (2.25, 53.6330) (2.30, 54.5378) (2.35, 54.9945) (2.40, 55.7421) (2.45, 56.2479) (2.50, 56.6229) (2.55, 57.1221) (2.60, 58.1007) (2.65, 58.5503) (2.70, 59.8229) (2.75, 60.3953) (2.80, 60.9156) (2.85, 61.9182) (2.90, 61.9039) (2.95, 62.8246) (3.00, 62.9802) (3.05, 64.2126) (3.10, 64.8222) (3.15, 65.3342) (3.20, 66.4375) (3.25, 67.2873) (3.30, 68.0838) (3.35, 68.0878) (3.40, 69.3305) (3.45, 70.2201) (3.50, 70.8836)}
\defdata{img512-S-XSu-dino-ghi}{(1.00, 123.2115) (1.05, 109.7223) (1.10, 99.3152) (1.15, 90.4769) (1.20, 82.6916) (1.25, 77.1002) (1.30, 71.3513) (1.35, 67.5886) (1.40, 64.1751) (1.45, 61.6909) (1.50, 58.7433) (1.55, 56.9219) (1.60, 56.0486) (1.65, 54.8206) (1.70, 54.2778) (1.75, 53.2224) (1.80, 52.9669) (1.85, 52.9190) (1.90, 52.8050) (1.95, 52.0968) (2.00, 52.9253) (2.05, 52.9352) (2.10, 52.9837) (2.15, 53.6902) (2.20, 54.0279) (2.25, 54.3967) (2.30, 54.8357) (2.35, 55.3378) (2.40, 56.3199) (2.45, 57.0602) (2.50, 57.2979) (2.55, 57.8037) (2.60, 58.7190) (2.65, 59.1631) (2.70, 59.9291) (2.75, 60.6141) (2.80, 61.7443) (2.85, 62.2786) (2.90, 62.5436) (2.95, 63.2428) (3.00, 64.4864) (3.05, 64.6608) (3.10, 65.1231) (3.15, 65.8913) (3.20, 66.9893) (3.25, 67.5678) (3.30, 68.3469) (3.35, 69.5418) (3.40, 69.9465) (3.45, 70.7621) (3.50, 71.1916)}

\defdata{img512-S-XSu-glo0.28-ghi2.90-fid-plo}{(0.025, 1.6502)}
\defdata{img512-S-XSu-glo0.28-ghi2.90-fid-phi}{(0.025, 1.7122)}
\defdata{img512-S-XSu-glo0.28-ghi2.90-fid-rlo}{(0.025, 1.6502)}
\defdata{img512-S-XSu-glo0.28-ghi2.90-fid-rhi}{(0.025, 1.7122)}
\defdata{img512-S-XSu-glo0.28-ghi2.90-fid-glo}{(1.00, 6.7764) (1.05, 6.0179) (1.10, 5.2788) (1.15, 4.7897) (1.20, 4.3100) (1.25, 3.8089) (1.30, 3.4932) (1.35, 3.1803) (1.40, 2.8805) (1.45, 2.6796) (1.50, 2.5008) (1.55, 2.3199) (1.60, 2.1897) (1.65, 2.0449) (1.70, 1.9815) (1.75, 1.8783) (1.80, 1.8610) (1.85, 1.7692) (1.90, 1.7574) (1.95, 1.7114) (2.00, 1.6874) (2.05, 1.7257) (2.10, 1.6502) (2.15, 1.6899) (2.20, 1.6987) (2.25, 1.6774) (2.30, 1.7182) (2.35, 1.6935) (2.40, 1.7061) (2.45, 1.7527) (2.50, 1.7657) (2.55, 1.7850) (2.60, 1.7817) (2.65, 1.8098) (2.70, 1.8733) (2.75, 1.8806) (2.80, 1.9568) (2.85, 1.9237) (2.90, 2.0055) (2.95, 2.0193) (3.00, 2.0103) (3.05, 2.0866) (3.10, 2.1149) (3.15, 2.1355) (3.20, 2.1442) (3.25, 2.1577) (3.30, 2.2229) (3.35, 2.2545) (3.40, 2.3053) (3.45, 2.3171) (3.50, 2.3357)}
\defdata{img512-S-XSu-glo0.28-ghi2.90-fid-ghi}{(1.00, 7.0127) (1.05, 6.1214) (1.10, 5.4677) (1.15, 4.8873) (1.20, 4.3215) (1.25, 3.9250) (1.30, 3.5228) (1.35, 3.2001) (1.40, 2.9840) (1.45, 2.7304) (1.50, 2.5369) (1.55, 2.3687) (1.60, 2.2112) (1.65, 2.1408) (1.70, 2.0864) (1.75, 1.9260) (1.80, 1.8891) (1.85, 1.8012) (1.90, 1.8153) (1.95, 1.7242) (2.00, 1.7200) (2.05, 1.7284) (2.10, 1.7122) (2.15, 1.7252) (2.20, 1.7172) (2.25, 1.7127) (2.30, 1.7562) (2.35, 1.7345) (2.40, 1.7442) (2.45, 1.7681) (2.50, 1.8289) (2.55, 1.8352) (2.60, 1.8520) (2.65, 1.8467) (2.70, 1.9276) (2.75, 1.9335) (2.80, 2.0297) (2.85, 1.9883) (2.90, 2.0438) (2.95, 2.0935) (3.00, 2.1021) (3.05, 2.1056) (3.10, 2.1813) (3.15, 2.1685) (3.20, 2.1953) (3.25, 2.2523) (3.30, 2.2696) (3.35, 2.3495) (3.40, 2.3694) (3.45, 2.3827) (3.50, 2.3619)}
\defdata{img512-S-XSu-glo0.28-ghi2.90-dino-plo}{(0.120, 49.1610)}
\defdata{img512-S-XSu-glo0.28-ghi2.90-dino-phi}{(0.120, 50.1896)}
\defdata{img512-S-XSu-glo0.28-ghi2.90-dino-rlo}{(0.120, 49.1610)}
\defdata{img512-S-XSu-glo0.28-ghi2.90-dino-rhi}{(0.120, 50.1896)}
\defdata{img512-S-XSu-glo0.28-ghi2.90-dino-glo}{(2.40, 49.7731) (2.50, 50.3174) (2.60, 49.3034) (2.65, 49.3916) (2.70, 49.1610) (2.75, 49.4485) (2.80, 50.3611) (2.90, 50.1533) (3.00, 50.3364)}
\defdata{img512-S-XSu-glo0.28-ghi2.90-dino-ghi}{(2.40, 49.7731) (2.50, 50.3174) (2.60, 49.3034) (2.65, 49.9756) (2.70, 50.1896) (2.75, 50.0185) (2.80, 50.3611) (2.90, 50.1533) (3.00, 50.3364)}

\defdata{img512-S-XS-128Mi-fid-plo}{(0.010, 15.2498) (0.015, 6.2518) (0.020, 4.2146) (0.025, 3.3243) (0.030, 2.7510) (0.035, 2.4032) (0.040, 2.0954) (0.045, 1.8735) (0.050, 1.7216) (0.055, 1.5759) (0.060, 1.4628) (0.065, 1.4239) (0.070, 1.3397) (0.075, 1.3887) (0.080, 1.3995) (0.085, 1.5019) (0.090, 1.5735) (0.095, 1.7047) (0.100, 1.8154) (0.105, 2.0374) (0.110, 2.2029) (0.115, 2.4504) (0.120, 2.7331) (0.125, 3.0320) (0.130, 3.3007) (0.135, 3.5540) (0.140, 4.0149) (0.145, 4.2672) (0.150, 4.6300) (0.155, 4.9492) (0.160, 5.3007) (0.165, 5.7446) (0.170, 6.0656) (0.175, 6.5421) (0.180, 6.9627) (0.185, 7.2711) (0.190, 7.6934) (0.195, 8.0517) (0.200, 8.5105) (0.205, 9.0093) (0.210, 9.4534) (0.215, 9.8349) (0.220, 10.3947) (0.225, 10.8843) (0.230, 11.5567) (0.235, 12.0450) (0.240, 12.8460) (0.245, 13.5627) (0.250, 14.4074)}
\defdata{img512-S-XS-128Mi-fid-phi}{(0.010, 15.4919) (0.015, 6.3522) (0.020, 4.2351) (0.025, 3.3715) (0.030, 2.7899) (0.035, 2.4713) (0.040, 2.2041) (0.045, 1.9016) (0.050, 1.7239) (0.055, 1.5816) (0.060, 1.5023) (0.065, 1.4532) (0.070, 1.4062) (0.075, 1.4141) (0.080, 1.4454) (0.085, 1.5259) (0.090, 1.6154) (0.095, 1.7590) (0.100, 1.8792) (0.105, 2.1225) (0.110, 2.2502) (0.115, 2.4841) (0.120, 2.7863) (0.125, 3.0521) (0.130, 3.3689) (0.135, 3.6706) (0.140, 4.0315) (0.145, 4.3521) (0.150, 4.7345) (0.155, 5.1537) (0.160, 5.4534) (0.165, 5.8236) (0.170, 6.1970) (0.175, 6.5855) (0.180, 7.0446) (0.185, 7.3933) (0.190, 7.7889) (0.195, 8.1649) (0.200, 8.6463) (0.205, 9.0667) (0.210, 9.5388) (0.215, 10.0151) (0.220, 10.4572) (0.225, 11.0595) (0.230, 11.6432) (0.235, 12.2470) (0.240, 12.9937) (0.245, 13.5895) (0.250, 14.4448)}
\defdata{img512-S-XS-128Mi-fid-rlo}{(0.010, 3.0265) (0.015, 2.7693) (0.020, 2.5884) (0.025, 2.4392) (0.030, 2.2933) (0.035, 2.2214) (0.040, 2.0836) (0.045, 2.0074) (0.050, 1.9364) (0.055, 1.8983) (0.060, 1.8047) (0.065, 1.7814) (0.070, 1.6856) (0.075, 1.6887) (0.080, 1.5945) (0.085, 1.5740) (0.090, 1.5480) (0.095, 1.4959) (0.100, 1.4493) (0.105, 1.3946) (0.110, 1.3898) (0.115, 1.3894) (0.120, 1.3656) (0.125, 1.3397) (0.130, 1.3972) (0.135, 1.4008) (0.140, 1.4619) (0.145, 1.5076) (0.150, 1.5885) (0.155, 1.6928) (0.160, 1.8194) (0.165, 1.9057) (0.170, 2.1249) (0.175, 2.2659) (0.180, 2.5957) (0.185, 2.8107) (0.190, 3.1725) (0.195, 3.5679) (0.200, 3.9749) (0.205, 4.5256) (0.210, 5.1054) (0.215, 5.6521) (0.220, 6.1613) (0.225, 6.9626) (0.230, 7.8130) (0.235, 8.4937) (0.240, 9.4248) (0.245, 10.0249) (0.250, 11.0261)}
\defdata{img512-S-XS-128Mi-fid-rhi}{(0.010, 3.1859) (0.015, 2.8736) (0.020, 2.6285) (0.025, 2.4489) (0.030, 2.3212) (0.035, 2.2666) (0.040, 2.1408) (0.045, 2.1196) (0.050, 1.9876) (0.055, 1.9687) (0.060, 1.8613) (0.065, 1.8416) (0.070, 1.7207) (0.075, 1.7321) (0.080, 1.6295) (0.085, 1.5991) (0.090, 1.5695) (0.095, 1.5099) (0.100, 1.5170) (0.105, 1.4597) (0.110, 1.4096) (0.115, 1.4170) (0.120, 1.4002) (0.125, 1.4062) (0.130, 1.4335) (0.135, 1.4400) (0.140, 1.4768) (0.145, 1.5442) (0.150, 1.6053) (0.155, 1.7025) (0.160, 1.8262) (0.165, 1.9421) (0.170, 2.1661) (0.175, 2.3437) (0.180, 2.6166) (0.185, 2.8718) (0.190, 3.3749) (0.195, 3.6122) (0.200, 4.0792) (0.205, 4.6236) (0.210, 5.1561) (0.215, 5.7616) (0.220, 6.3882) (0.225, 7.1103) (0.230, 7.8161) (0.235, 8.7604) (0.240, 9.5532) (0.245, 10.2832) (0.250, 11.1098)}
\defdata{img512-S-XS-128Mi-fid-glo}{(1.00, 3.8762) (1.05, 3.6203) (1.10, 3.3475) (1.15, 3.0522) (1.20, 2.9031) (1.25, 2.7021) (1.30, 2.4543) (1.35, 2.3130) (1.40, 2.1909) (1.45, 2.0611) (1.50, 1.9385) (1.55, 1.8375) (1.60, 1.7802) (1.65, 1.6512) (1.70, 1.6356) (1.75, 1.5427) (1.80, 1.5166) (1.85, 1.4518) (1.90, 1.4639) (1.95, 1.4215) (2.00, 1.3839) (2.05, 1.3914) (2.10, 1.3397) (2.15, 1.3841) (2.20, 1.3758) (2.25, 1.3738) (2.30, 1.3786) (2.35, 1.3675) (2.40, 1.3961) (2.45, 1.4109) (2.50, 1.3999) (2.55, 1.4476) (2.60, 1.4811) (2.65, 1.4985) (2.70, 1.5373) (2.75, 1.5483) (2.80, 1.5415) (2.85, 1.5912) (2.90, 1.6120) (2.95, 1.6523) (3.00, 1.7122) (3.05, 1.7226) (3.10, 1.7906) (3.15, 1.8253) (3.20, 1.8567) (3.25, 1.8998) (3.30, 1.9451) (3.35, 1.9438) (3.40, 2.0385) (3.45, 2.0675) (3.50, 2.1100)}
\defdata{img512-S-XS-128Mi-fid-ghi}{(1.00, 4.0321) (1.05, 3.7000) (1.10, 3.4416) (1.15, 3.1641) (1.20, 2.9265) (1.25, 2.7373) (1.30, 2.4758) (1.35, 2.3498) (1.40, 2.2559) (1.45, 2.1161) (1.50, 1.9652) (1.55, 1.9046) (1.60, 1.8724) (1.65, 1.7175) (1.70, 1.6988) (1.75, 1.6037) (1.80, 1.5608) (1.85, 1.4983) (1.90, 1.4730) (1.95, 1.4504) (2.00, 1.3892) (2.05, 1.4147) (2.10, 1.4062) (2.15, 1.3955) (2.20, 1.3967) (2.25, 1.4116) (2.30, 1.4158) (2.35, 1.4347) (2.40, 1.4313) (2.45, 1.4421) (2.50, 1.4478) (2.55, 1.4754) (2.60, 1.5070) (2.65, 1.5173) (2.70, 1.5616) (2.75, 1.5719) (2.80, 1.6054) (2.85, 1.6443) (2.90, 1.6657) (2.95, 1.7030) (3.00, 1.7282) (3.05, 1.7609) (3.10, 1.8405) (3.15, 1.8338) (3.20, 1.9034) (3.25, 1.9376) (3.30, 1.9848) (3.35, 1.9752) (3.40, 2.0762) (3.45, 2.1221) (3.50, 2.2013)}
\defdata{img512-S-XS-128Mi-dino-plo}{(0.100, 39.9229) (0.110, 37.5982) (0.115, 37.3852) (0.120, 36.6668) (0.125, 37.5611) (0.130, 37.8126) (0.135, 38.7164) (0.140, 40.6244)}
\defdata{img512-S-XS-128Mi-dino-phi}{(0.100, 39.9229) (0.110, 37.5982) (0.115, 37.7797) (0.120, 38.1835) (0.125, 37.8135) (0.130, 37.8126) (0.135, 38.7164) (0.140, 40.6244)}
\defdata{img512-S-XS-128Mi-dino-rlo}{(0.145, 39.1617) (0.155, 38.0841) (0.160, 37.9580) (0.165, 36.6668) (0.170, 37.0408) (0.175, 37.1055) (0.180, 37.2988) (0.185, 37.3583)}
\defdata{img512-S-XS-128Mi-dino-rhi}{(0.145, 39.1617) (0.155, 38.0841) (0.160, 38.0754) (0.165, 38.1835) (0.170, 37.5574) (0.175, 37.1055) (0.180, 37.2988) (0.185, 37.3583)}
\defdata{img512-S-XS-128Mi-dino-glo}{(1.00, 96.2984) (1.05, 88.9408) (1.10, 81.5489) (1.15, 76.7873) (1.20, 72.2005) (1.25, 67.7189) (1.30, 64.2785) (1.35, 60.3452) (1.40, 57.2680) (1.45, 55.0050) (1.50, 52.3497) (1.55, 50.3669) (1.60, 48.6609) (1.65, 47.2823) (1.70, 45.4594) (1.75, 43.8948) (1.80, 43.1367) (1.85, 42.0937) (1.90, 41.1836) (1.95, 40.3443) (2.00, 39.8994) (2.05, 39.1409) (2.10, 38.4086) (2.15, 38.4259) (2.20, 37.9063) (2.25, 37.8182) (2.30, 37.7618) (2.35, 37.4860) (2.40, 37.4230) (2.45, 36.6668) (2.50, 37.3633) (2.55, 37.4127) (2.60, 37.3766) (2.65, 37.7703) (2.70, 37.4773) (2.75, 37.9302) (2.80, 38.0549) (2.85, 38.8168) (2.90, 38.2418) (2.95, 38.7916) (3.00, 39.0042) (3.05, 39.8112) (3.10, 40.0621) (3.15, 40.4937) (3.20, 40.8537) (3.25, 40.8915) (3.30, 41.7593) (3.35, 41.9299) (3.40, 42.2901) (3.45, 42.9098) (3.50, 43.3970)}
\defdata{img512-S-XS-128Mi-dino-ghi}{(1.00, 97.0291) (1.05, 90.1031) (1.10, 82.9762) (1.15, 77.5052) (1.20, 73.0241) (1.25, 68.1595) (1.30, 64.3116) (1.35, 60.6511) (1.40, 57.8397) (1.45, 55.2909) (1.50, 52.6708) (1.55, 50.9913) (1.60, 49.0488) (1.65, 47.4780) (1.70, 46.1647) (1.75, 44.6428) (1.80, 43.4388) (1.85, 42.5337) (1.90, 41.6886) (1.95, 40.7808) (2.00, 40.5367) (2.05, 39.3843) (2.10, 39.1394) (2.15, 38.6882) (2.20, 38.3036) (2.25, 38.2944) (2.30, 38.0388) (2.35, 38.0274) (2.40, 37.8061) (2.45, 38.1835) (2.50, 37.8430) (2.55, 37.7621) (2.60, 37.7351) (2.65, 38.0650) (2.70, 38.1364) (2.75, 38.2744) (2.80, 38.4168) (2.85, 39.0722) (2.90, 39.0011) (2.95, 39.6849) (3.00, 39.1974) (3.05, 40.0495) (3.10, 40.8338) (3.15, 40.8214) (3.20, 41.1642) (3.25, 41.7287) (3.30, 41.9928) (3.35, 42.3863) (3.40, 43.0306) (3.45, 43.0874) (3.50, 44.2566)}

\defdata{img512-S-XSu-glo0.60-ghi5.00-fid-plo}{(0.030, 1.8780)}
\defdata{img512-S-XSu-glo0.60-ghi5.00-fid-phi}{(0.030, 1.9879)}
\defdata{img512-S-XSu-glo0.60-ghi5.00-fid-rlo}{(0.030, 1.8780)}
\defdata{img512-S-XSu-glo0.60-ghi5.00-fid-rhi}{(0.030, 1.9879)}
\defdata{img512-S-XSu-glo0.60-ghi5.00-fid-glo}{(1.50, 2.2033) (1.55, 2.1171) (1.60, 2.0212) (1.65, 1.9889) (1.70, 1.9125) (1.75, 1.8780) (1.80, 1.9003) (1.85, 1.9381) (1.90, 1.9302) (1.95, 1.9730) (2.00, 1.9984) (2.10, 2.0629) (2.20, 2.2106) (2.30, 2.3227) (2.40, 2.3923) (2.50, 2.5944) (2.60, 2.5625) (2.70, 2.7719) (2.80, 2.8117)}
\defdata{img512-S-XSu-glo0.60-ghi5.00-fid-ghi}{(1.50, 2.2033) (1.55, 2.1171) (1.60, 2.0212) (1.65, 1.9889) (1.70, 2.0184) (1.75, 1.9879) (1.80, 1.9610) (1.85, 1.9381) (1.90, 1.9302) (1.95, 1.9730) (2.00, 1.9984) (2.10, 2.0629) (2.20, 2.2106) (2.30, 2.3227) (2.40, 2.3923) (2.50, 2.5944) (2.60, 2.5625) (2.70, 2.7719) (2.80, 2.8117)}
\defdata{img512-S-XSu-glo0.60-ghi5.00-dino-plo}{(0.085, 46.2560)}
\defdata{img512-S-XSu-glo0.60-ghi5.00-dino-phi}{(0.085, 46.4723)}
\defdata{img512-S-XSu-glo0.60-ghi5.00-dino-rlo}{(0.085, 46.2560)}
\defdata{img512-S-XSu-glo0.60-ghi5.00-dino-rhi}{(0.085, 46.4723)}
\defdata{img512-S-XSu-glo0.60-ghi5.00-dino-glo}{(1.00, 122.3640) (1.05, 114.4664) (1.10, 107.2830) (1.15, 100.6894) (1.20, 94.5728) (1.25, 88.8950) (1.30, 85.2452) (1.35, 81.0743) (1.40, 77.7065) (1.45, 74.1744) (1.50, 71.2300) (1.55, 68.6246) (1.60, 66.1665) (1.65, 63.9911) (1.70, 62.7626) (1.75, 60.7642) (1.80, 59.5764) (1.85, 58.0356) (1.90, 56.3281) (1.95, 55.3729) (2.00, 54.1070) (2.05, 53.4044) (2.10, 52.5904) (2.15, 51.7870) (2.20, 50.9731) (2.25, 50.1877) (2.30, 49.6455) (2.35, 48.9754) (2.40, 49.0091) (2.45, 48.3136) (2.50, 48.1378) (2.55, 47.6724) (2.60, 47.2323) (2.65, 47.1047) (2.70, 46.7962) (2.75, 46.4511) (2.80, 46.5200) (2.85, 46.9559) (2.90, 45.6330) (2.95, 46.6728) (3.00, 46.3041) (3.05, 46.1097) (3.10, 46.1695) (3.15, 46.2676) (3.20, 46.2560) (3.25, 46.0311) (3.30, 46.4685) (3.35, 46.3722) (3.40, 46.6244) (3.45, 45.8178) (3.50, 46.7397)}
\defdata{img512-S-XSu-glo0.60-ghi5.00-dino-ghi}{(1.00, 123.7094) (1.05, 114.7458) (1.10, 107.5632) (1.15, 101.0402) (1.20, 95.8980) (1.25, 89.4073) (1.30, 86.1679) (1.35, 81.6473) (1.40, 77.8994) (1.45, 75.3470) (1.50, 72.3598) (1.55, 69.7135) (1.60, 67.2240) (1.65, 65.0218) (1.70, 63.1715) (1.75, 61.8319) (1.80, 60.1351) (1.85, 58.5654) (1.90, 56.7917) (1.95, 56.2080) (2.00, 54.9272) (2.05, 53.6246) (2.10, 53.0382) (2.15, 52.3874) (2.20, 51.7369) (2.25, 51.1976) (2.30, 50.4099) (2.35, 50.2849) (2.40, 49.2659) (2.45, 48.8458) (2.50, 48.3641) (2.55, 47.6968) (2.60, 47.4765) (2.65, 47.9049) (2.70, 47.1400) (2.75, 47.0060) (2.80, 47.2109) (2.85, 47.3366) (2.90, 46.9211) (2.95, 46.9328) (3.00, 46.5998) (3.05, 46.5820) (3.10, 47.2067) (3.15, 46.5890) (3.20, 46.4723) (3.25, 46.4974) (3.30, 46.9952) (3.35, 46.8262) (3.40, 46.7085) (3.45, 47.0654) (3.50, 47.2549)}

\defdata{img512-S-S-128Mi-fid-plo}{(0.050, 2.9153) (0.060, 2.3592) (0.070, 1.9072) (0.080, 1.6304) (0.085, 1.5456) (0.090, 1.5085) (0.095, 1.5270) (0.100, 1.6125) (0.110, 1.8431)}
\defdata{img512-S-S-128Mi-fid-phi}{(0.050, 2.9153) (0.060, 2.3592) (0.070, 1.9072) (0.080, 1.6304) (0.085, 1.5793) (0.090, 1.5709) (0.095, 1.6027) (0.100, 1.6125) (0.110, 1.8431)}
\defdata{img512-S-S-128Mi-fid-rlo}{(0.090, 1.7469) (0.100, 1.6863) (0.110, 1.5803) (0.120, 1.5922) (0.125, 1.5441) (0.130, 1.5085) (0.135, 1.5213) (0.140, 1.5683) (0.150, 1.6479)}
\defdata{img512-S-S-128Mi-fid-rhi}{(0.090, 1.7469) (0.100, 1.6863) (0.110, 1.5803) (0.120, 1.5922) (0.125, 1.5900) (0.130, 1.5709) (0.135, 1.5801) (0.140, 1.5683) (0.150, 1.6479)}
\defdata{img512-S-S-128Mi-fid-glo}{(1.00, 3.1675) (1.05, 2.9734) (1.10, 2.8652) (1.15, 2.7080) (1.20, 2.5947) (1.25, 2.4467) (1.30, 2.3005) (1.35, 2.2362) (1.40, 2.1626) (1.45, 2.0715) (1.50, 1.9862) (1.55, 1.9194) (1.60, 1.8719) (1.65, 1.7998) (1.70, 1.7706) (1.75, 1.7085) (1.80, 1.7156) (1.85, 1.6140) (1.90, 1.6357) (1.95, 1.5669) (2.00, 1.5936) (2.05, 1.6037) (2.10, 1.5354) (2.15, 1.5565) (2.20, 1.5085) (2.25, 1.5215) (2.30, 1.5453) (2.35, 1.5121) (2.40, 1.5621) (2.45, 1.5332) (2.50, 1.5869) (2.55, 1.5756) (2.60, 1.5930) (2.65, 1.6284) (2.70, 1.6482) (2.75, 1.6380) (2.80, 1.7151) (2.85, 1.7056) (2.90, 1.7406) (2.95, 1.7311) (3.00, 1.8201) (3.05, 1.8095) (3.10, 1.8669) (3.15, 1.8712) (3.20, 1.8991) (3.25, 1.9422) (3.30, 1.9935) (3.35, 2.0370) (3.40, 2.0972) (3.45, 2.1097) (3.50, 2.1621)}
\defdata{img512-S-S-128Mi-fid-ghi}{(1.00, 3.1946) (1.05, 3.0627) (1.10, 2.9599) (1.15, 2.7872) (1.20, 2.6297) (1.25, 2.5040) (1.30, 2.4135) (1.35, 2.2564) (1.40, 2.2171) (1.45, 2.1322) (1.50, 2.0364) (1.55, 1.9427) (1.60, 1.8950) (1.65, 1.8382) (1.70, 1.7954) (1.75, 1.7447) (1.80, 1.7270) (1.85, 1.6771) (1.90, 1.6537) (1.95, 1.6359) (2.00, 1.6130) (2.05, 1.6228) (2.10, 1.6005) (2.15, 1.5717) (2.20, 1.5709) (2.25, 1.5675) (2.30, 1.5657) (2.35, 1.5731) (2.40, 1.5749) (2.45, 1.6042) (2.50, 1.6135) (2.55, 1.6117) (2.60, 1.6288) (2.65, 1.6752) (2.70, 1.6794) (2.75, 1.6695) (2.80, 1.7468) (2.85, 1.7244) (2.90, 1.7546) (2.95, 1.7668) (3.00, 1.8309) (3.05, 1.8362) (3.10, 1.8853) (3.15, 1.9045) (3.20, 1.9787) (3.25, 2.0159) (3.30, 2.0027) (3.35, 2.0831) (3.40, 2.1068) (3.45, 2.1361) (3.50, 2.1899)}
\defdata{img512-S-S-128Mi-dino-plo}{(0.120, 44.2810) (0.125, 43.3386) (0.130, 42.2742) (0.135, 43.0267) (0.140, 43.7847)}
\defdata{img512-S-S-128Mi-dino-phi}{(0.120, 44.2810) (0.125, 43.7814) (0.130, 43.2285) (0.135, 43.4927) (0.140, 43.7847)}
\defdata{img512-S-S-128Mi-dino-rlo}{(0.155, 43.6388) (0.160, 42.9739) (0.165, 42.7490) (0.170, 42.2742) (0.175, 43.3146)}
\defdata{img512-S-S-128Mi-dino-rhi}{(0.155, 43.6388) (0.160, 43.4492) (0.165, 42.8069) (0.170, 43.2285) (0.175, 43.3146)}
\defdata{img512-S-S-128Mi-dino-glo}{(2.20, 44.3848) (2.30, 43.7582) (2.35, 43.1901) (2.40, 43.0102) (2.45, 42.8184) (2.50, 43.0621) (2.55, 42.2742) (2.60, 42.9770) (2.70, 43.9709)}
\defdata{img512-S-S-128Mi-dino-ghi}{(2.20, 44.3848) (2.30, 43.7582) (2.35, 43.1901) (2.40, 43.0102) (2.45, 43.2385) (2.50, 43.5163) (2.55, 43.2285) (2.60, 42.9770) (2.70, 43.9709)}

\defdata{img512-S-XS-64Mi-fid-plo}{(0.020, 1.9020) (0.030, 1.4931) (0.035, 1.4295) (0.040, 1.4046) (0.045, 1.4293) (0.050, 1.4743) (0.060, 1.6485)}
\defdata{img512-S-XS-64Mi-fid-phi}{(0.020, 1.9020) (0.030, 1.4931) (0.035, 1.4722) (0.040, 1.4566) (0.045, 1.4689) (0.050, 1.4743) (0.060, 1.6485)}
\defdata{img512-S-XS-64Mi-fid-rlo}{(0.110, 1.5340) (0.115, 1.4635) (0.120, 1.4780) (0.125, 1.4582) (0.130, 1.4217) (0.135, 1.4046) (0.140, 1.4465) (0.145, 1.4348) (0.150, 1.4536) (0.155, 1.4273) (0.160, 1.4601)}
\defdata{img512-S-XS-64Mi-fid-rhi}{(0.110, 1.5340) (0.115, 1.4635) (0.120, 1.4780) (0.125, 1.4582) (0.130, 1.4492) (0.135, 1.4566) (0.140, 1.4663) (0.145, 1.4348) (0.150, 1.4536) (0.155, 1.4273) (0.160, 1.4601)}
\defdata{img512-S-XS-64Mi-fid-glo}{(1.00, 5.5336) (1.05, 5.0999) (1.10, 4.5125) (1.15, 4.0741) (1.20, 3.6700) (1.25, 3.2531) (1.30, 2.9076) (1.35, 2.6337) (1.40, 2.4379) (1.45, 2.1903) (1.50, 2.0412) (1.55, 1.8939) (1.60, 1.8094) (1.65, 1.6684) (1.70, 1.5975) (1.75, 1.5360) (1.80, 1.4623) (1.85, 1.4702) (1.90, 1.4263) (1.95, 1.4341) (2.00, 1.4046) (2.05, 1.4130) (2.10, 1.4528) (2.15, 1.4348) (2.20, 1.4610) (2.25, 1.5331) (2.30, 1.5108) (2.35, 1.5750) (2.40, 1.5942) (2.45, 1.6404) (2.50, 1.6565) (2.55, 1.7215) (2.60, 1.8038) (2.65, 1.7649) (2.70, 1.8252) (2.75, 1.8555) (2.80, 1.9342) (2.85, 2.0174) (2.90, 2.0877) (2.95, 2.0864) (3.00, 2.1533) (3.05, 2.1884) (3.10, 2.2538) (3.15, 2.3015) (3.20, 2.3692) (3.25, 2.4082) (3.30, 2.4330) (3.35, 2.5249) (3.40, 2.5772) (3.45, 2.6215) (3.50, 2.7261)}
\defdata{img512-S-XS-64Mi-fid-ghi}{(1.00, 5.6424) (1.05, 5.1403) (1.10, 4.6556) (1.15, 4.1254) (1.20, 3.7119) (1.25, 3.3000) (1.30, 2.9387) (1.35, 2.7787) (1.40, 2.4935) (1.45, 2.2774) (1.50, 2.0740) (1.55, 1.9472) (1.60, 1.8456) (1.65, 1.7331) (1.70, 1.6255) (1.75, 1.5743) (1.80, 1.5596) (1.85, 1.4821) (1.90, 1.4595) (1.95, 1.4742) (2.00, 1.4566) (2.05, 1.4270) (2.10, 1.4744) (2.15, 1.4853) (2.20, 1.5020) (2.25, 1.5525) (2.30, 1.5970) (2.35, 1.6211) (2.40, 1.6729) (2.45, 1.6634) (2.50, 1.6980) (2.55, 1.7440) (2.60, 1.8079) (2.65, 1.8090) (2.70, 1.9070) (2.75, 1.9304) (2.80, 2.0076) (2.85, 2.0394) (2.90, 2.1129) (2.95, 2.1471) (3.00, 2.2018) (3.05, 2.2685) (3.10, 2.3060) (3.15, 2.3308) (3.20, 2.4470) (3.25, 2.4379) (3.30, 2.4797) (3.35, 2.5391) (3.40, 2.5934) (3.45, 2.7021) (3.50, 2.7567)}
\defdata{img512-S-XS-64Mi-dino-plo}{(0.085, 39.7300) (0.095, 38.4550) (0.100, 38.3040) (0.105, 37.9546) (0.110, 38.7182) (0.115, 39.2407) (0.120, 40.1976) (0.125, 40.5345)}
\defdata{img512-S-XS-64Mi-dino-phi}{(0.085, 39.7300) (0.095, 38.4550) (0.100, 38.9668) (0.105, 38.6526) (0.110, 38.8071) (0.115, 39.2407) (0.120, 40.1976) (0.125, 40.5345)}
\defdata{img512-S-XS-64Mi-dino-rlo}{(0.175, 39.2499) (0.185, 38.8425) (0.190, 38.3101) (0.195, 37.9546) (0.200, 38.6461) (0.205, 38.4238) (0.210, 38.6989) (0.215, 39.3293)}
\defdata{img512-S-XS-64Mi-dino-rhi}{(0.175, 39.2499) (0.185, 38.8425) (0.190, 38.5059) (0.195, 38.6526) (0.200, 38.9465) (0.205, 38.4238) (0.210, 38.6989) (0.215, 39.3293)}
\defdata{img512-S-XS-64Mi-dino-glo}{(1.95, 39.2989) (2.05, 39.2918) (2.10, 38.5962) (2.15, 37.9546) (2.20, 38.3214) (2.25, 38.3617) (2.30, 38.8306) (2.35, 38.4557)}
\defdata{img512-S-XS-64Mi-dino-ghi}{(1.95, 39.2989) (2.05, 39.2918) (2.10, 39.2173) (2.15, 38.6526) (2.20, 38.9439) (2.25, 38.3617) (2.30, 38.8306) (2.35, 38.4557)}

\defdata{img512-S-XS-256Mi-fid-plo}{(0.065, 1.9098) (0.070, 1.7223) (0.075, 1.5690) (0.080, 1.4967) (0.085, 1.4738) (0.090, 1.5364) (0.095, 1.5753) (0.100, 1.6448) (0.105, 1.8746) (0.110, 1.9248)}
\defdata{img512-S-XS-256Mi-fid-phi}{(0.065, 1.9098) (0.070, 1.7223) (0.075, 1.5690) (0.080, 1.5290) (0.085, 1.5142) (0.090, 1.5456) (0.095, 1.5753) (0.100, 1.6448) (0.105, 1.8746) (0.110, 1.9248)}
\defdata{img512-S-XS-256Mi-fid-rlo}{(0.090, 1.6053) (0.100, 1.5631) (0.105, 1.5088) (0.110, 1.4738) (0.115, 1.5219) (0.120, 1.5606) (0.125, 1.5042) (0.130, 1.6103)}
\defdata{img512-S-XS-256Mi-fid-rhi}{(0.090, 1.6053) (0.100, 1.5631) (0.105, 1.5523) (0.110, 1.5142) (0.115, 1.5509) (0.120, 1.5606) (0.125, 1.5042) (0.130, 1.6103)}
\defdata{img512-S-XS-256Mi-fid-glo}{(1.00, 3.2544) (1.05, 3.1778) (1.10, 2.8851) (1.15, 2.7649) (1.20, 2.6017) (1.25, 2.4360) (1.30, 2.2816) (1.35, 2.2082) (1.40, 2.1234) (1.45, 1.9966) (1.50, 1.9446) (1.55, 1.8349) (1.60, 1.8110) (1.65, 1.7585) (1.70, 1.7107) (1.75, 1.6499) (1.80, 1.6346) (1.85, 1.5790) (1.90, 1.5707) (1.95, 1.5263) (2.00, 1.5218) (2.05, 1.5048) (2.10, 1.5042) (2.15, 1.5229) (2.20, 1.4738) (2.25, 1.5055) (2.30, 1.5215) (2.35, 1.4802) (2.40, 1.5449) (2.45, 1.5283) (2.50, 1.5563) (2.55, 1.5947) (2.60, 1.6240) (2.65, 1.6451) (2.70, 1.6321) (2.75, 1.6760) (2.80, 1.6889) (2.85, 1.7591) (2.90, 1.7639) (2.95, 1.8238) (3.00, 1.8753) (3.05, 1.8635) (3.10, 1.8864) (3.15, 1.9803) (3.20, 2.0200) (3.25, 2.0756) (3.30, 2.0927) (3.35, 2.1793) (3.40, 2.2266) (3.45, 2.2597) (3.50, 2.3253)}
\defdata{img512-S-XS-256Mi-fid-ghi}{(1.00, 3.3469) (1.05, 3.1966) (1.10, 2.9626) (1.15, 2.8385) (1.20, 2.6763) (1.25, 2.5292) (1.30, 2.3822) (1.35, 2.2518) (1.40, 2.1753) (1.45, 2.1004) (1.50, 1.9938) (1.55, 1.9545) (1.60, 1.8435) (1.65, 1.7919) (1.70, 1.7356) (1.75, 1.6828) (1.80, 1.6489) (1.85, 1.6287) (1.90, 1.5851) (1.95, 1.5571) (2.00, 1.5553) (2.05, 1.5329) (2.10, 1.5311) (2.15, 1.5398) (2.20, 1.5142) (2.25, 1.5363) (2.30, 1.5440) (2.35, 1.5289) (2.40, 1.5874) (2.45, 1.5460) (2.50, 1.6019) (2.55, 1.6234) (2.60, 1.6528) (2.65, 1.7006) (2.70, 1.6711) (2.75, 1.7186) (2.80, 1.7242) (2.85, 1.7745) (2.90, 1.8123) (2.95, 1.8387) (3.00, 1.8956) (3.05, 1.9307) (3.10, 1.9688) (3.15, 2.0431) (3.20, 2.0473) (3.25, 2.1002) (3.30, 2.1255) (3.35, 2.1986) (3.40, 2.2420) (3.45, 2.2793) (3.50, 2.3339)}
\defdata{img512-S-XS-256Mi-dino-plo}{(0.135, 43.6549) (0.140, 41.1944) (0.145, 39.7742) (0.150, 39.0524) (0.155, 38.2178) (0.160, 38.7980) (0.165, 39.2940) (0.175, 41.9563)}
\defdata{img512-S-XS-256Mi-dino-phi}{(0.135, 43.6549) (0.140, 41.1944) (0.145, 39.7742) (0.150, 39.5709) (0.155, 39.2471) (0.160, 39.2555) (0.165, 39.2940) (0.175, 41.9563)}
\defdata{img512-S-XS-256Mi-dino-rlo}{(0.175, 39.5111) (0.180, 39.3117) (0.185, 39.1230) (0.190, 38.7077) (0.195, 38.2178) (0.200, 38.5678) (0.205, 39.9584) (0.215, 41.2753)}
\defdata{img512-S-XS-256Mi-dino-rhi}{(0.175, 39.5111) (0.180, 39.3117) (0.185, 39.1230) (0.190, 39.2915) (0.195, 39.2471) (0.200, 39.0675) (0.205, 39.9584) (0.215, 41.2753)}
\defdata{img512-S-XS-256Mi-dino-glo}{(2.05, 39.3149) (2.15, 38.6171) (2.20, 38.5709) (2.25, 38.2178) (2.30, 38.6841) (2.35, 38.4640) (2.40, 38.7374) (2.45, 39.0830)}
\defdata{img512-S-XS-256Mi-dino-ghi}{(2.05, 39.3149) (2.15, 38.6171) (2.20, 38.9345) (2.25, 39.2471) (2.30, 39.3473) (2.35, 38.4640) (2.40, 38.7374) (2.45, 39.0830)}

\defdata{img512-S-XXS-128Mi-fid-plo}{(0.010, 6.9353) (0.020, 1.9937) (0.030, 1.5749) (0.040, 1.4517) (0.045, 1.4219) (0.050, 1.4145) (0.055, 1.4969) (0.060, 1.5615) (0.070, 1.8016)}
\defdata{img512-S-XXS-128Mi-fid-phi}{(0.010, 6.9353) (0.020, 1.9937) (0.030, 1.5749) (0.040, 1.4517) (0.045, 1.4439) (0.050, 1.5088) (0.055, 1.5539) (0.060, 1.5615) (0.070, 1.8016)}
\defdata{img512-S-XXS-128Mi-fid-rlo}{(0.110, 1.5590) (0.120, 1.5277) (0.130, 1.4760) (0.135, 1.4479) (0.140, 1.4145) (0.145, 1.4829) (0.150, 1.4879) (0.160, 1.5725) (0.170, 1.6796)}
\defdata{img512-S-XXS-128Mi-fid-rhi}{(0.110, 1.5590) (0.120, 1.5277) (0.130, 1.4760) (0.135, 1.5021) (0.140, 1.5088) (0.145, 1.5099) (0.150, 1.4879) (0.160, 1.5725) (0.170, 1.6796)}
\defdata{img512-S-XXS-128Mi-fid-glo}{(1.00, 4.9959) (1.05, 4.2831) (1.10, 3.6413) (1.15, 3.1632) (1.20, 2.8395) (1.25, 2.5019) (1.30, 2.2597) (1.35, 2.0515) (1.40, 1.8163) (1.45, 1.7173) (1.50, 1.6432) (1.55, 1.5787) (1.60, 1.4822) (1.65, 1.4658) (1.70, 1.4461) (1.75, 1.4238) (1.80, 1.4145) (1.85, 1.4639) (1.90, 1.4975) (1.95, 1.5561) (2.00, 1.6010) (2.05, 1.6054) (2.10, 1.7026) (2.15, 1.7272) (2.20, 1.7804) (2.25, 1.8404) (2.30, 1.9349) (2.35, 2.0108) (2.40, 2.0446) (2.45, 2.0768) (2.50, 2.1316) (2.55, 2.1910) (2.60, 2.3491) (2.65, 2.3723) (2.70, 2.4541) (2.75, 2.5069) (2.80, 2.5621) (2.85, 2.7013) (2.90, 2.7228) (2.95, 2.7699) (3.00, 2.8376) (3.05, 2.9548) (3.10, 2.9839) (3.15, 3.1274) (3.20, 3.1483) (3.25, 3.2501) (3.30, 3.3313) (3.35, 3.4330) (3.40, 3.4593) (3.45, 3.5745) (3.50, 3.6886)}
\defdata{img512-S-XXS-128Mi-fid-ghi}{(1.00, 5.1649) (1.05, 4.3465) (1.10, 3.7783) (1.15, 3.2709) (1.20, 2.9030) (1.25, 2.5369) (1.30, 2.2959) (1.35, 2.0947) (1.40, 1.8758) (1.45, 1.7586) (1.50, 1.6543) (1.55, 1.6117) (1.60, 1.5300) (1.65, 1.4906) (1.70, 1.4689) (1.75, 1.4659) (1.80, 1.5088) (1.85, 1.5388) (1.90, 1.5627) (1.95, 1.5985) (2.00, 1.6284) (2.05, 1.7150) (2.10, 1.7511) (2.15, 1.8239) (2.20, 1.8833) (2.25, 1.9085) (2.30, 1.9827) (2.35, 2.0238) (2.40, 2.0803) (2.45, 2.1370) (2.50, 2.2484) (2.55, 2.3048) (2.60, 2.3872) (2.65, 2.3887) (2.70, 2.4708) (2.75, 2.5427) (2.80, 2.5718) (2.85, 2.7211) (2.90, 2.7993) (2.95, 2.8525) (3.00, 2.9058) (3.05, 3.0056) (3.10, 3.0931) (3.15, 3.1963) (3.20, 3.2250) (3.25, 3.2773) (3.30, 3.3492) (3.35, 3.5414) (3.40, 3.5361) (3.45, 3.6436) (3.50, 3.7529)}
\defdata{img512-S-XXS-128Mi-dino-plo}{(0.090, 37.3523) (0.100, 36.6857) (0.105, 36.4525) (0.110, 36.2487) (0.115, 36.8050) (0.120, 37.6404) (0.130, 39.0227)}
\defdata{img512-S-XXS-128Mi-dino-phi}{(0.090, 37.3523) (0.100, 36.6857) (0.105, 36.6942) (0.110, 37.3685) (0.115, 37.3439) (0.120, 37.6404) (0.130, 39.0227)}
\defdata{img512-S-XXS-128Mi-dino-rlo}{(0.160, 37.8923) (0.170, 36.8661) (0.175, 36.9380) (0.180, 36.2487) (0.185, 36.2944) (0.190, 37.2728) (0.200, 37.8335)}
\defdata{img512-S-XXS-128Mi-dino-rhi}{(0.160, 37.8923) (0.170, 36.8661) (0.175, 37.2680) (0.180, 37.3685) (0.185, 36.6822) (0.190, 37.2728) (0.200, 37.8335)}
\defdata{img512-S-XXS-128Mi-dino-glo}{(1.80, 37.8205) (1.85, 37.7485) (1.90, 37.3119) (1.95, 37.1619) (2.00, 36.5368) (2.05, 36.2487) (2.10, 36.7086) (2.15, 36.9260) (2.20, 36.9225) (2.25, 37.3855) (2.30, 37.0109)}
\defdata{img512-S-XXS-128Mi-dino-ghi}{(1.80, 37.8205) (1.85, 37.7485) (1.90, 37.3119) (1.95, 37.1619) (2.00, 37.4399) (2.05, 37.3685) (2.10, 37.3138) (2.15, 36.9260) (2.20, 36.9225) (2.25, 37.3855) (2.30, 37.0109)}

\defdata{img512-S-nog-edm2-fid}{2.56}
\defdata{img512-S-nog-edm2-fid-p}{0.130}
\defdata{img512-S-nog-edm2-fid-r}{0.130}
\defdata{img512-S-nog-edm2-fid-g}{1.00}
\defdata{img512-S-nog-edm2-dino}{68.64}
\defdata{img512-S-nog-edm2-dino-p}{0.190}
\defdata{img512-S-nog-edm2-dino-r}{0.190}
\defdata{img512-S-nog-edm2-dino-g}{1.00}

\defdata{img512-S-XSu-edm2-fid}{2.23}
\defdata{img512-S-XSu-edm2-fid-p}{0.025}
\defdata{img512-S-XSu-edm2-fid-r}{0.025}
\defdata{img512-S-XSu-edm2-fid-g}{1.40}
\defdata{img512-S-XSu-edm2-dino}{52.32}
\defdata{img512-S-XSu-edm2-dino-p}{0.085}
\defdata{img512-S-XSu-edm2-dino-r}{0.085}
\defdata{img512-S-XSu-edm2-dino-g}{1.90}

\defdata{img512-S-XSu-limit-tuomas-fid}{1.68}
\defdata{img512-S-XSu-limit-tuomas-fid-p}{0.025}
\defdata{img512-S-XSu-limit-tuomas-fid-r}{0.025}
\defdata{img512-S-XSu-limit-tuomas-fid-g}{2.10}
\defdata{img512-S-XSu-limit-tuomas-dino}{46.25}
\defdata{img512-S-XSu-limit-tuomas-dino-p}{0.085}
\defdata{img512-S-XSu-limit-tuomas-dino-r}{0.085}
\defdata{img512-S-XSu-limit-tuomas-dino-g}{3.20}

\defdata{img512-XXL-nog-edm2-fid}{1.91}
\defdata{img512-XXL-nog-edm2-fid-p}{0.070}
\defdata{img512-XXL-nog-edm2-fid-r}{0.070}
\defdata{img512-XXL-nog-edm2-fid-g}{1.00}
\defdata{img512-XXL-nog-edm2-dino}{42.84}
\defdata{img512-XXL-nog-edm2-dino-p}{0.150}
\defdata{img512-XXL-nog-edm2-dino-r}{0.150}
\defdata{img512-XXL-nog-edm2-dino-g}{1.00}

\defdata{img512-XXL-XSu-edm2-fid}{1.81}
\defdata{img512-XXL-XSu-edm2-fid-p}{0.015}
\defdata{img512-XXL-XSu-edm2-fid-r}{0.015}
\defdata{img512-XXL-XSu-edm2-fid-g}{1.20}
\defdata{img512-XXL-XSu-edm2-dino}{33.09}
\defdata{img512-XXL-XSu-edm2-dino-p}{0.015}
\defdata{img512-XXL-XSu-edm2-dino-r}{0.015}
\defdata{img512-XXL-XSu-edm2-dino-g}{1.70}

\defdata{img512-XXL-XSu-limit-tuomas-fid}{1.40}
\defdata{img512-XXL-XSu-limit-tuomas-fid-p}{0.015}
\defdata{img512-XXL-XSu-limit-tuomas-fid-r}{0.015}
\defdata{img512-XXL-XSu-limit-tuomas-fid-g}{2.00}
\defdata{img512-XXL-XSu-limit-tuomas-dino}{29.16}
\defdata{img512-XXL-XSu-limit-tuomas-dino-p}{0.015}
\defdata{img512-XXL-XSu-limit-tuomas-dino-r}{0.015}
\defdata{img512-XXL-XSu-limit-tuomas-dino-g}{2.90}

\defdata{img64-S-nog-edm2-fid}{1.58}
\defdata{img64-S-nog-edm2-fid-p}{0.075}
\defdata{img64-S-nog-edm2-fid-r}{0.075}
\defdata{img64-S-nog-edm2-fid-g}{1.00}
\defdata{img64-S-nog-edm2-dino}{\data{img64-S-nog-dino}}
\defdata{img64-S-nog-edm2-dino-p}{\data{img64-S-nog-dino-p}}
\defdata{img64-S-nog-edm2-dino-r}{\data{img64-S-nog-dino-r}}
\defdata{img64-S-nog-edm2-dino-g}{\data{img64-S-nog-dino-g}}

\defdata{img64-rin-fid}{1.23}
\defdata{img64-rin-fid-p}{0.033}
\defdata{img64-rin-fid-r}{0.033}
\defdata{img64-rin-fid-g}{1.00}

\title{Guiding a Diffusion Model with a Bad Version of Itself}

\ifcustomstyle
  \author{ Anonymous Author(s)\\Affiliation\\Address\\\texttt{email}}
\else
  \author{
    Tero Karras \\ NVIDIA \And
    Miika Aittala \\ NVIDIA \And
    Tuomas Kynk\"a\"anniemi \\ Aalto University \AND
    Jaakko Lehtinen \\ NVIDIA, Aalto University \And
    Timo Aila \\ NVIDIA \And
    Samuli Laine \\ NVIDIA
  }
\fi

\begin{document}

\maketitle

\newcommand{\figToyExample}{%
\begin{figure}[t]%
\def\w{0.195\linewidth}%
\def\panel##1##2##3{%
  \centering\setlength{\fboxsep}{0pt}\fbox{\includegraphics[width=0.996\linewidth]{toy2d/##1.jpg}}%
  \caption{\small\parbox[t]{##2\linewidth}{##3\strut}}%
}%
\begin{subfigure}{\w}\panel{gt}{.59}{Ground truth\!\!\\}\label{fig:toyGT}\end{subfigure}\hfill%
\begin{subfigure}{\w}\panel{nog}{.6}{No guidance\!\\}\label{fig:toyNog}\end{subfigure}\hfill%
\begin{subfigure}{\w}\panel{cfg-insetbox}{.67}{Classifier-free\!\\guidance}\label{fig:toyCFG}\end{subfigure}\hfill%
\begin{subfigure}{\w}\panel{naive}{.77}{Naive truncation\!\!\\}\label{fig:toyNaive}\end{subfigure}\hfill%
\begin{subfigure}{\w}\panel{ours}{.65}{Autoguidance\\(ours)}\label{fig:toyOurs}\end{subfigure}%
\vspace*{-1mm}%
\caption{\label{figToyExample}%
A fractal-like 2D distribution with two classes indicated with gray and orange regions.
Approximately~99\% of the probability mass is inside the shown contours.
\textbf{(a)} Ground truth samples drawn directly from the orange class distribution.
\textbf{(b)} Conditional sampling using a small denoising diffusion model generates outliers.
\textbf{(c)} Classifier-free guidance ($\guidance=4$) eliminates outliers but reduces diversity by over-emphasizing the class.
\textbf{(d)} Naive truncation via lengthening the score vectors.
\textbf{(e)} Our method concentrates samples on high-probability regions without reducing diversity.
}\vspace*{0mm}%
\end{figure}
}%

\newcommand{\figToyDetail}{%
\begin{figure}[t]%
\def\w{0.195\linewidth}%
\def\panel##1{\centering\setlength{\fboxsep}{0pt}\fbox{\includegraphics[width=0.996\linewidth, trim={100 0 130 0}, clip]{toy2d/##1.pdf}}}
\begin{subfigure}{\w}\panel{pmain}\caption{$p_1(\boldx | \boldc; \sigma_\text{mid})$}\label{fig:toyPMain}\end{subfigure}\hfill%
\begin{subfigure}{\w}\panel{prudder}\caption{$p_0(\boldx; \sigma_\text{mid})$}\label{fig:toyPRudder}\end{subfigure}\hfill%
\begin{subfigure}{\w}\panel{pratio}\caption{Ratio $p_1 / p_0$}\label{fig:toyPRatio}\end{subfigure}\hfill%
\begin{subfigure}{\w}\panel{tmain-shade}\caption{No guidance}\label{fig:toyTMain}\end{subfigure}\hfill%
\begin{subfigure}{\w}\panel{tguid-shade}\caption{CFG with $\guidance=4$}\label{fig:toyTGuid}\end{subfigure}%
\caption{\label{figToyDetail}%
Closeup of the region highlighted in Figure~\ref{fig:toyCFG}.
\textbf{(a)} The implied learned density $\pmain(\boldx | \boldc; \sigma_\text{mid})$ (green) at an intermediate noise level $\sigma_\text{mid}$ and its score vectors (log-gradients), plotted at representative sample points. The learned density approximates the underlying ground truth $p(\boldx | \boldc; \sigma_\text{mid})$ (orange) but fails to replicate its sharper details.
\textbf{(b)} The weaker unconditional model learns a further spread-out density $\prud(\boldx; \sigma_\text{mid})$ (red) with a looser fit to the data.
\textbf{(c)} Guidance moves the points according to the gradient of the (log) ratio of the two learned densities (blue). As the higher-quality model is more sharply concentrated at the data, this field tends inward towards the data distribution. The corresponding gradient is simply the difference of respective gradients in (a) and (b), illustrated at selected points.
\textbf{(d)} Sampling trajectories taken by standard unguided diffusion following the learned score $\nabla_\boldx \log \pmain(\boldx | \boldc; \sigma)$, from noise level $\sigma_\text{mid}$ to $0$. The contours (orange) represent the ground truth noise-free density.
\textbf{(e)} Guidance introduces an additional force shown in (c), causing the points to concentrate at~the core of the data density during sampling.
}%
\end{figure}
}%

\newcommand{\figToyProgress}{%
\begin{figure}[p]%
\def\ww{0.027\linewidth}%
\def\w{0.190\linewidth}%
\def\col##1{\begin{subfigure}{\w}\footnotesize\centering##1\end{subfigure}}%
\def\row##1{\begin{subfigure}{\ww}\rotatebox{90}{\footnotesize##1}\end{subfigure}}%
\def\panel##1{\begin{subfigure}{\w}\centering\setlength{\fboxsep}{0pt}\fbox{\includegraphics[width=0.996\linewidth, trim={200 150 130 0}, clip]{toy2d/progress/##1.pdf}}\end{subfigure}}
\row{\vphantom{$\sigma=0$}}%
\col{(a) $\pmain(\boldx | \boldc; \sigma)$}\hfill%
\col{(b) CFG \\ $\prud(\boldx; \sigma)$}\hfill%
\col{(c) Autoguidance \\ $\prud(\boldx | \boldc; \sigma)$}\hfill%
\col{(d) CFG \\ ratio $\pmain / \prud$}\hfill%
\col{(e) Autoguidance \\ ratio $\pmain / \prud$}\\[.3ex]%
\row{\makebox[32mm][c]{$\sigma=0.5 $}}\panel{0-pmain}\hfill\panel{0-prudder}\hfill\panel{0-prudder-ag}\hfill\panel{0-pratio}\hfill\panel{0-pratio-ag}\\
\row{\makebox[32mm][c]{$\sigma=0.25$}}\panel{1-pmain}\hfill\panel{1-prudder}\hfill\panel{1-prudder-ag}\hfill\panel{1-pratio}\hfill\panel{1-pratio-ag}\\
\row{\makebox[32mm][c]{$\sigma=0.08$}}\panel{2-pmain}\hfill\panel{2-prudder}\hfill\panel{2-prudder-ag}\hfill\panel{2-pratio}\hfill\panel{2-pratio-ag}\\
\row{\makebox[32mm][c]{$\sigma=0.03$}}\panel{3-pmain}\hfill\panel{3-prudder}\hfill\panel{3-prudder-ag}\hfill\panel{3-pratio}\hfill\panel{3-pratio-ag}\\
\row{\makebox[32mm][c]{$\sigma=0.01$}}\panel{4-pmain}\hfill\panel{4-prudder}\hfill\panel{4-prudder-ag}\hfill\panel{4-pratio}\hfill\panel{4-pratio-ag}%
\caption{\label{figToyProgress}%
Progression of implied learned densities during sampling over various $\sigma$ in a setup similar to Figure~\ref{figToyDetail}.
Contours of the corresponding ground truth distributions are also shown.
\textbf{(a)}~Main model density $\pmain(\boldx | \boldc; \sigma)$.
\textbf{(b)}~Unconditional guiding model density $\prud(\boldx; \sigma)$ in CFG.
\textbf{(c)}~Conditional guiding model density $\prud(\boldx | \boldc; \sigma)$ in autoguidance.
\textbf{(d)}~With CFG, guidance towards higher ratio $\pmain/\prud$ pushes samples towards top right, especially at high $\sigma$ (top rows).
\textbf{(e)}~With autoguidance, this anomalous effect does not occur and samples cover the entire class $\boldc$.
}%
\end{figure}
}%

\newcommand{\tabMainResults}{%
\begin{table}[t]%
\centering\footnotesize%
\def\tab{\hs{2.8}}%
\def\fade##1{\scalebox{0.95}{\color{black!60}##1}}%
\begin{tabu}{|@{\hs{0.9}}l@{\hs{0.7}}|l@{\hs{1.9}}c@{\hs{1.9}}|@{\hs{3.0}}cccc|@{\hs{1.4}}c@{\hs{2.1}}ccc|}
\tabucline{2-}\multicolumn{1}{l|}{}\\[-3.65mm]%
\multicolumn{1}{l|}{} & \multicolumn{2}{l|@{\hs{3.0}}}{\tstrut\bf Method} & {FID} & {$\guidance$} & \makebox[0mm][c]{\fade{\mainEMA}} & \makebox[0mm][c]{\fade{\rudderEMA}} & {\DINO} & {$\guidance$} & \makebox[0mm][c]{\fade{\mainEMA}} & \makebox[0mm][c]{\fade{\rudderEMA}} \\
\tabucline{-}\tstrut%
\multirow{13}{*}{\rotatebox{90}{\hs{4.2}\bf 512$\times$512}}
\gdef\cid{img512-S-nog-edm2}&           {EDM2-S}                              & {\cite{Karras2024edm2}}           & {\data{\cid-fid}}          & {--}                & \fade{\data{\cid-fid-p}} & \fade{--}                & {\data{\cid-dino}}          & {--}                 & \fade{\data{\cid-dino-p}} & \fade{--}                 \\
\gdef\cid{img512-S-XSu-edm2}&           {+ Classifier-free guidance}          & {\cite{Karras2024edm2}}           & {\data{\cid-fid}}          & {\data{\cid-fid-g}} & \fade{\data{\cid-fid-p}} & \fade{\data{\cid-fid-r}} & {\data{\cid-dino}}          & {\data{\cid-dino-g}} & \fade{\data{\cid-dino-p}} & \fade{\data{\cid-dino-r}} \\
\gdef\cid{img512-S-XSu-limit-tuomas}&   {\tab+ Guidance interval}             & {\cite{Kynkaanniemi2024guidance}} & {\data{\cid-fid}}          & {\data{\cid-fid-g}} & \fade{\data{\cid-fid-p}} & \fade{\data{\cid-fid-r}} & {\data{\cid-dino}}          & {\data{\cid-dino-g}} & \fade{\data{\cid-dino-p}} & \fade{\data{\cid-dino-r}} \\
\gdef\cid{\cidBestS}&                   {+ Autoguidance (XS, $T/16$)}         & {Ours}                            & {\bf\data{\cid-fid}}       & {\data{\cid-fid-g}} & \fade{\data{\cid-fid-p}} & \fade{\data{\cid-fid-r}} & {\bf\data{\cid-dino}}       & {\data{\cid-dino-g}} & \fade{\data{\cid-dino-p}} & \fade{\data{\cid-dino-r}} \\
\gdef\cid{\cidBestS-sameEMA}&           {\tab\textminus~Same EMA for both}    & {}                                & {\data{\cid-fid}}          & {\data{\cid-fid-g}} & \fade{\data{\cid-fid-p}} & \fade{\data{\cid-fid-r}} & {\data{\cid-dino}}          & {\data{\cid-dino-g}} & \fade{\data{\cid-dino-p}} & \fade{\data{\cid-dino-r}} \\
\gdef\cid{img512-S-S-128Mi}&            {\tab\textminus~Reduce training only} & {}                                & {\data{\cid-fid}}          & {\data{\cid-fid-g}} & \fade{\data{\cid-fid-p}} & \fade{\data{\cid-fid-r}} & {\data{\cid-dino}}          & {\data{\cid-dino-g}} & \fade{\data{\cid-dino-p}} & \fade{\data{\cid-dino-r}} \\
\gdef\cid{img512-S-XS}&                 {\tab\textminus~Reduce capacity only} & {}                                & {\data{\cid-fid}}          & {\data{\cid-fid-g}} & \fade{\data{\cid-fid-p}} & \fade{\data{\cid-fid-r}} & {\data{\cid-dino}}          & {\data{\cid-dino-g}} & \fade{\data{\cid-dino-p}} & \fade{\data{\cid-dino-r}} \\
\tabucline{2-}\\[-3.66mm]\tstrut%
\gdef\cid{img512-XXL-nog-edm2}&         {EDM2-XXL}                            & {\cite{Karras2024edm2}}           & {\data{\cid-fid}}          & {--}                & \fade{\data{\cid-fid-p}} & \fade{--}                & {\data{\cid-dino}}          & {--}                 & \fade{\data{\cid-dino-p}} & \fade{--}                 \\
\gdef\cid{img512-XXL-XSu-edm2}&         {+ Classifier-free guidance}          & {\cite{Karras2024edm2}}           & {\data{\cid-fid}}          & {\data{\cid-fid-g}} & \fade{\data{\cid-fid-p}} & \fade{\data{\cid-fid-r}} & {\data{\cid-dino}}          & {\data{\cid-dino-g}} & \fade{\data{\cid-dino-p}} & \fade{\data{\cid-dino-r}} \\
\gdef\cid{img512-XXL-XSu-limit-tuomas}& {\tab+ Guidance interval}             & {\cite{Kynkaanniemi2024guidance}} & {\data{\cid-fid}}          & {\data{\cid-fid-g}} & \fade{\data{\cid-fid-p}} & \fade{\data{\cid-fid-r}} & {\data{\cid-dino}}          & {\data{\cid-dino-g}} & \fade{\data{\cid-dino-p}} & \fade{\data{\cid-dino-r}} \\
\gdef\cid{\cidBestXXL}&                 {+ Autoguidance (M, $T/3.5$)}         & {Ours}                            & {\bf\data{\cid-fid}}       & {\data{\cid-fid-g}} & \fade{\data{\cid-fid-p}} & \fade{\data{\cid-fid-r}} & {\bf\data{\cid-dino}}       & {\data{\cid-dino-g}} & \fade{\data{\cid-dino-p}} & \fade{\data{\cid-dino-r}} \\
\tabucline{2-}\\[-3.66mm]\tstrut%
\gdef\cid{img512-Su-nog}&               {EDM2-S, unconditional}               & {}                                & {\hs{-1.6}\data{\cid-fid}} & {--}                & \fade{\data{\cid-fid-p}} & \fade{--}                & {\hs{-1.6}\data{\cid-dino}} & {--}                 & \fade{\data{\cid-dino-p}} & \fade{--}                 \\
\gdef\cid{\cidBestUncond}&              {+ Autoguidance (XS, $T/16$)}         & {Ours}                            & {\bf\data{\cid-fid}}       & {\data{\cid-fid-g}} & \fade{\data{\cid-fid-p}} & \fade{\data{\cid-fid-r}} & {\bf\data{\cid-dino}}       & {\data{\cid-dino-g}} & \fade{\data{\cid-dino-p}} & \fade{\data{\cid-dino-r}} \\
\tabucline{-}\tstrut%
\multirow{4}{*}{\rotatebox{90}{\hs{0.5}\bf 64$\times$64}}
\gdef\cid{img64-rin}&                   {RIN}                                 & {\cite{Jabri2023}}                & {\data{\cid-fid}}          & {--}                & \fade{\data{\cid-fid-p}} & \fade{--}                & {--}                        & {--}                 & \fade{--}                 & \fade{--}                 \\
\gdef\cid{img64-S-nog-edm2}&            {EDM2-S}                              & {\cite{Karras2024edm2}}           & {\data{\cid-fid}}          & {--}                & \fade{\data{\cid-fid-p}} & \fade{--}                & {\data{\cid-dino}}          & {--}                 & \fade{\data{\cid-dino-p}} & \fade{--}                 \\
\gdef\cid{img64-S-XSu}&                 {+ Classifier-free guidance}          & {}                                & {\data{\cid-fid}}          & {\data{\cid-fid-g}} & \fade{\data{\cid-fid-p}} & \fade{\data{\cid-fid-r}} & {\data{\cid-dino}}          & {\data{\cid-dino-g}} & \fade{\data{\cid-dino-p}} & \fade{\data{\cid-dino-r}} \\
\gdef\cid{\cidBestSixtyFour}&           {+ Autoguidance (XS, $T/8$)}          & {Ours}                            & {\bf\data{\cid-fid}}       & {\data{\cid-fid-g}} & \fade{\data{\cid-fid-p}} & \fade{\data{\cid-fid-r}} & {\bf\data{\cid-dino}}       & {\data{\cid-dino-g}} & \fade{\data{\cid-dino-p}} & \fade{\data{\cid-dino-r}} \\
\tabucline{-}
\end{tabu}%
\vspace{2.5mm}%
\caption{\label{tabMainResults}%
Results on ImageNet-512 and ImageNet-64.
The parameters of autoguidance refer to the capacity and amount training received by the guiding model. The latter is given relative to the number of training images shown to the main model ($T$).
The columns \mainEMA{} and \rudderEMA{} indicate the length parameter of the post-hoc EMA technique~\cite{Karras2024edm2} for the main and guiding model, respectively.
}\vspace*{0mm}%
\end{table}
}%

\newcommand{\plotMainGuidanceA}{%
\centering\footnotesize%
\begin{tikzpicture}
\def\colorA{C1}\def\cidA{img512-S-XS-64Mi}
\def\colorB{C2}\def\cidB{img512-S-XS-128Mi}
\def\colorC{C4}\def\cidC{img512-S-XS-256Mi}
\begin{axis}[
  width={1.20\linewidth}, height={50mm}, grid={major},
  xmin={1.00}, xmax={3.20}, xmode={linear}, xtick={1.00, 1.50, 2.00, 2.50, 3.00}, xticklabels={\makebox[0mm][r]{$\guidance{=}$}$1.0$, $1.5$, $2.0$, $2.5$, $3.0$},
  ymin={1.0}, ymax={4.0}, ymode={linear}, ytick={1.5, 2.0, 2.5, 3.0, 3.5, 4.0}, yticklabels={$1.5$, $2.0$, $2.5$, $3.0$, $3.5$, \raisebox{-1.5ex}[0ex][0ex]{FID}},
  legend pos={north east}, legend cell align={left}, legend columns={3}, legend style={/tikz/every even column/.append style={column sep=0.7mm}},
]
\fillbetween[\colorA, opacity=0.2, forget plot]{coordinates {\rawdata{\cidA-fid-glo}}}{coordinates {\rawdata{\cidA-fid-ghi}}};
\fillbetween[\colorB, opacity=0.2, forget plot]{coordinates {\rawdata{\cidB-fid-glo}}}{coordinates {\rawdata{\cidB-fid-ghi}}};
\fillbetween[\colorC, opacity=0.2, forget plot]{coordinates {\rawdata{\cidC-fid-glo}}}{coordinates {\rawdata{\cidC-fid-ghi}}};
\addplot[\colorA, forget plot] coordinates {\rawdata{\cidA-fid-glo}};
\addplot[\colorB, forget plot] coordinates {\rawdata{\cidB-fid-glo}};
\addplot[\colorC, forget plot] coordinates {\rawdata{\cidC-fid-glo}};
\addplot[\colorA, mark=*, forget plot, nodes near coords align={north}, nodes near coords=\datalabel{\hs{-3.5}\data{\cidA-fid}}] coordinates {(\rawdata{\cidA-fid-g}, \rawdata{\cidA-fid-precise})};
\addplot[\colorB, mark=*, forget plot, nodes near coords align={north}, nodes near coords=\datalabel{\hs{2}\data{\cidB-fid}}] coordinates {(\rawdata{\cidB-fid-g}, \rawdata{\cidB-fid-precise})};
\addplot[\colorC, mark=*, forget plot, nodes near coords align={south}, nodes near coords=\datalabel{\hs{-1}\data{\cidC-fid}}] coordinates {(\rawdata{\cidC-fid-g}, \rawdata{\cidC-fid-precise})};
\addlegendimage{\colorA, mark=*}\addlegendentry{$T/32$}
\addlegendimage{\colorB, mark=*}\addlegendentry{$T/16^\star$}
\addlegendimage{\colorC, mark=*}\addlegendentry{$T/8$}
\end{axis}
\end{tikzpicture}%
}%

\newcommand{\plotMainGuidanceB}{%
\centering\footnotesize%
\begin{tikzpicture}
\def\colorA{C0}\def\cidA{img512-S-XXS-128Mi}
\def\colorB{C2}\def\cidB{img512-S-XS-128Mi}
\def\colorC{C6}\def\cidC{img512-S-S-128Mi}
\begin{axis}[
  width={1.20\linewidth}, height={50mm}, grid={major},
  xmin={1.00}, xmax={3.20}, xmode={linear}, xtick={1.00, 1.50, 2.00, 2.50, 3.00}, xticklabels={\makebox[0mm][r]{$\guidance{=}$}$1.0$, $1.5$, $2.0$, $2.5$, $3.0$},
  ymin={1.0}, ymax={4.0}, ymode={linear}, ytick={1.5, 2.0, 2.5, 3.0, 3.5, 4.0}, yticklabels={$1.5$, $2.0$, $2.5$, $3.0$, $3.5$, \raisebox{-1.5ex}[0ex][0ex]{FID}},
  legend pos={north east}, legend cell align={left}, legend columns={3}, legend style={/tikz/every even column/.append style={column sep=1.0mm}},
]
\fillbetween[\colorA, opacity=0.2, forget plot]{coordinates {\rawdata{\cidA-fid-glo}}}{coordinates {\rawdata{\cidA-fid-ghi}}};
\fillbetween[\colorB, opacity=0.2, forget plot]{coordinates {\rawdata{\cidB-fid-glo}}}{coordinates {\rawdata{\cidB-fid-ghi}}};
\fillbetween[\colorC, opacity=0.2, forget plot]{coordinates {\rawdata{\cidC-fid-glo}}}{coordinates {\rawdata{\cidC-fid-ghi}}};
\addplot[\colorA, forget plot] coordinates {\rawdata{\cidA-fid-glo}};
\addplot[\colorB, forget plot] coordinates {\rawdata{\cidB-fid-glo}};
\addplot[\colorC, forget plot] coordinates {\rawdata{\cidC-fid-glo}};
\addplot[\colorA, mark=*, forget plot, nodes near coords align={north}, nodes near coords=\datalabel{\data{\cidA-fid}}] coordinates {(\rawdata{\cidA-fid-g}, \rawdata{\cidA-fid-precise})};
\addplot[\colorB, mark=*, forget plot, nodes near coords align={north}, nodes near coords=\datalabel{\data{\cidB-fid}}] coordinates {(\rawdata{\cidB-fid-g}, \rawdata{\cidB-fid-precise})};
\addplot[\colorC, mark=*, forget plot, nodes near coords align={south}, nodes near coords=\datalabel{\hs{2.5}\data{\cidC-fid}}] coordinates {(\rawdata{\cidC-fid-g}, \rawdata{\cidC-fid-precise})};
\addlegendimage{\colorA, mark=*}\addlegendentry{XXS}
\addlegendimage{\colorB, mark=*}\addlegendentry{XS$^\star$}
\addlegendimage{\colorC, mark=*}\addlegendentry{S}
\end{axis}
\end{tikzpicture}%
}%

\newcommand{\plotMainEMA}{%
\centering\footnotesize%
\begin{tikzpicture}
\def\colorA{C2!70!black}\def\colorB{C2!70!white}\def\cid{img512-S-XS-128Mi}
\begin{axis}[
  width={1.20\linewidth}, height={50mm}, grid={major},
  xmin={0.015}, xmax={0.180}, xmode={linear}, xtick={0.050, 0.100, 0.150}, xticklabels={\makebox[0mm][r]{\ema{}$\,=$\hs{0.7}}$0.05$, $0.10$, $0.15$},
  ymin={1.0}, ymax={4.0}, ymode={linear}, ytick={1.5, 2.0, 2.5, 3.0, 3.5, 4.0}, yticklabels={$1.5$, $2.0$, $2.5$, $3.0$, $3.5$, \raisebox{-1.5ex}[0ex][0ex]{FID}},
  legend pos={north east}, legend cell align={left}, legend columns={2}, legend style={/tikz/every even column/.append style={column sep=1.0mm}},
]
\fillbetween[\colorA, opacity=0.2, forget plot]{coordinates {\rawdata{\cid-fid-plo}}}{coordinates {\rawdata{\cid-fid-phi}}};
\fillbetween[\colorB, opacity=0.2, forget plot]{coordinates {\rawdata{\cid-fid-rlo}}}{coordinates {\rawdata{\cid-fid-rhi}}};
\addplot[\colorA, forget plot] coordinates {\rawdata{\cid-fid-plo}};
\addplot[\colorB, forget plot] coordinates {\rawdata{\cid-fid-rlo}};
\addplot[\colorA, mark=*, forget plot, nodes near coords align={north}, nodes near coords=\datalabel{\data{\cid-fid}}] coordinates {(\rawdata{\cid-fid-p}, \rawdata{\cid-fid-precise})};
\addplot[\colorB, mark=*, forget plot, nodes near coords align={north}, nodes near coords=\datalabel{\data{\cid-fid}}] coordinates {(\rawdata{\cid-fid-r}, \rawdata{\cid-fid-precise})};
\addlegendimage{\colorA, mark=*}\addlegendentry{\mainEMA}
\addlegendimage{\colorB, mark=*}\addlegendentry{\rudderEMA}
\end{axis}
\end{tikzpicture}%
}%

\newcommand{\figMainPlots}{%
\begin{figure}[t]%
\def\w{0.33\linewidth}%
\begin{subfigure}{\w}\plotMainGuidanceA\caption{\small Guidance weight \& training}\end{subfigure}\hfill%
\begin{subfigure}{\w}\plotMainGuidanceB\caption{\small Guidance weight \& capacity}\end{subfigure}\hfill%
\begin{subfigure}{\w}\plotMainEMA\caption{\small EMA length parameters}\end{subfigure}%
\caption{\label{figMainPlots}%
Sensitivity w.r.t.~autoguidance parameters, using EDM2-S on ImageNet-512.
The shaded regions indicate the min/max FID over 3 evaluations.
\textbf{(a)}
Sweep over guidance weight $\guidance$ while keeping all other parameters unchanged.
The curves correspond to how much the guiding model was  trained relative to the number of images shown to the main model.
\textbf{(b)}
Sweep over guidance weight for different guiding model capacities.
\textbf{(c)}
Sweep over the two EMA length parameters for our best configuration, denoted with~$^\star$ in (a) and (b).
}%
\end{figure}
}%

\newcommand{\figDeepFloydMain}{%
\usetikzlibrary {arrows.meta}
\begin{figure}[t]%
\def\w{0.935\linewidth}%
\centering\footnotesize%
\rotatebox{90}{\hs{40}Increasing CFG weight}\hfill%
\begin{tikzpicture}
  \draw[use as bounding box, draw=none] (0,0) rectangle (0,10);
  \draw[arrows = {-Stealth[scale=1.25]}] (0,10) -- (0,0.5);
\end{tikzpicture}\hfill\hfill%
\includegraphics[width=\w]{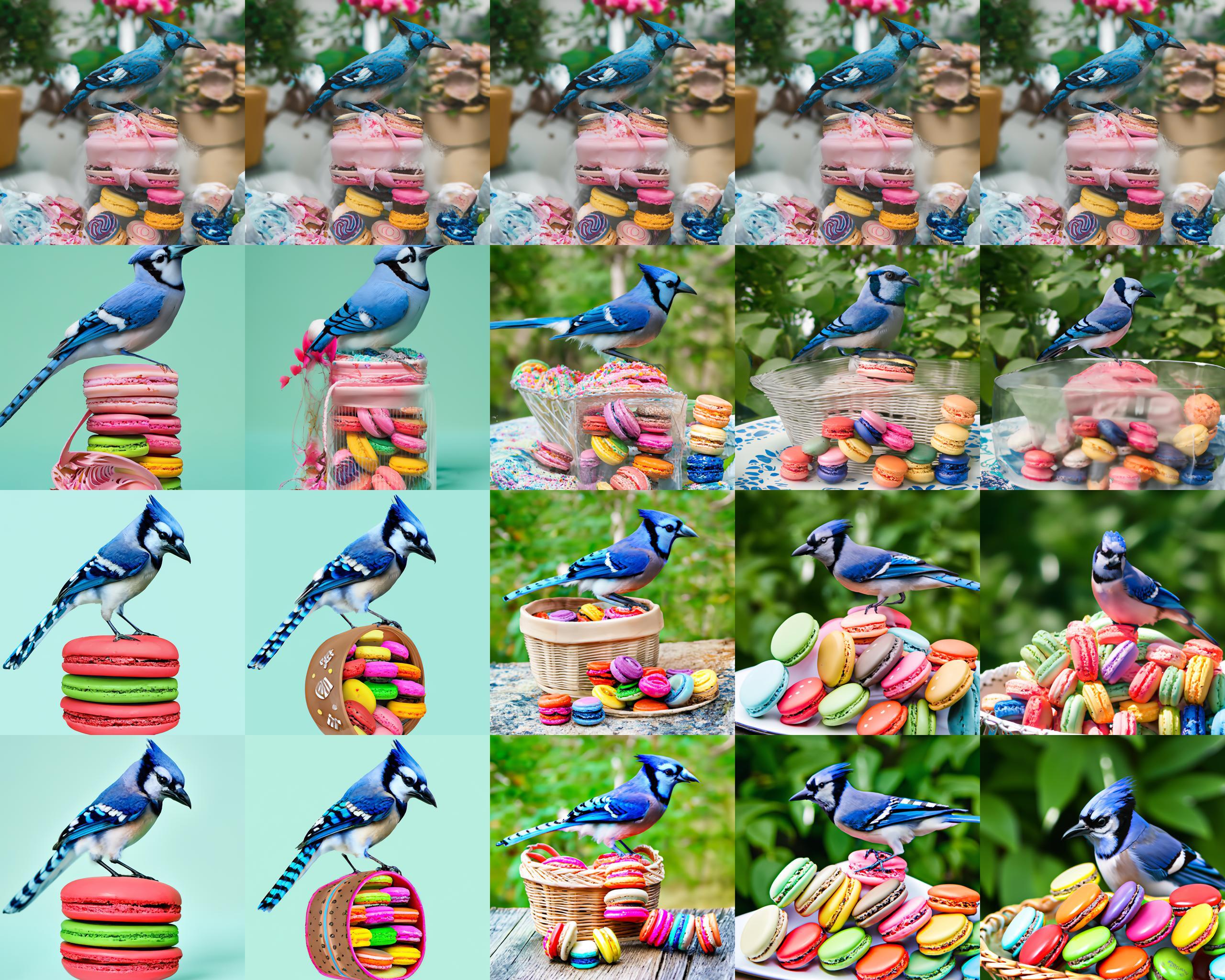}\hfill\hfill%
\begin{tikzpicture}
  \draw[use as bounding box, draw=none] (0,0) rectangle (0,10);
  \draw[arrows = {-Stealth[scale=1.25]}] (0,10) -- (0,0.5);
\end{tikzpicture}\hfill%
\rotatebox{90}{\hs{25}Increasing guidance weight for our method}\\[-2mm]%
\begin{tikzpicture}
  \draw[use as bounding box, draw=none] (0,0) rectangle (13,0);
  \draw[arrows = {Stealth[scale=1.25]-Stealth[scale=1.25]}] (1,0) -- (12,0);
\end{tikzpicture}\\%
Interpolation between CFG and our method\\[0mm]
\caption{\label{figDeepFloydMain}%
Results for DeepFloyd~IF~\cite{DeepFloyd} using the prompt \emph{``A blue jay standing on a large basket of rainbow macarons''}.
The rows correspond to guidance weights $w \in \{1, 2, 3, 4\}$.
The leftmost column shows results for CFG and the rightmost for autoguidance (XL-sized model guided by M-sized one).
The middle columns correspond to blending between the two.
See Appendix~\ref{app:deepFloydExamples} for more examples.
}\vspace*{0mm}%
\end{figure}
}%

\newcommand{\figImageNetMain}{%
\begin{figure}[t]%
\def\w{0.4845\linewidth}%
\def\ww{0.1615\linewidth}%
\centering\footnotesize%
\rotatebox{90}{\vphantom{INVISIBLE}}\hfill%
\makebox[\ww][c]{$\guidance = 1$}%
\makebox[\ww][c]{$\guidance = 2$}%
\makebox[\ww][c]{$\guidance = 3$}\hfill%
\makebox[\ww][c]{$\guidance = 1$}%
\makebox[\ww][c]{$\guidance = 2$}%
\makebox[\ww][c]{$\guidance = 3$}\\[0.5mm]
\rotatebox{90}{\makebox[\ww][c]{Ours}\makebox[\ww][c]{CFG}}\hfill%
\includegraphics[width=\w]{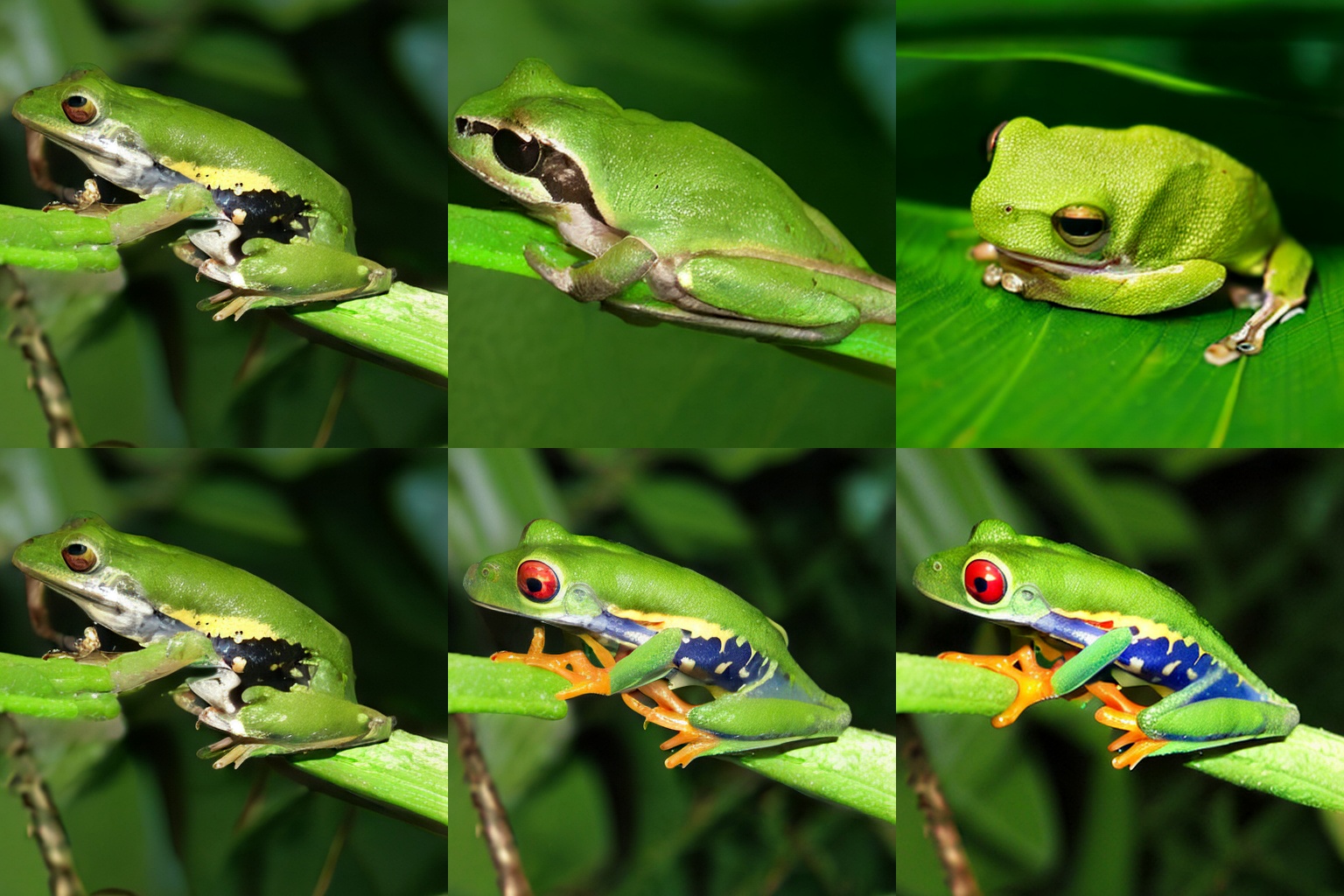}\hfill%
\includegraphics[width=\w]{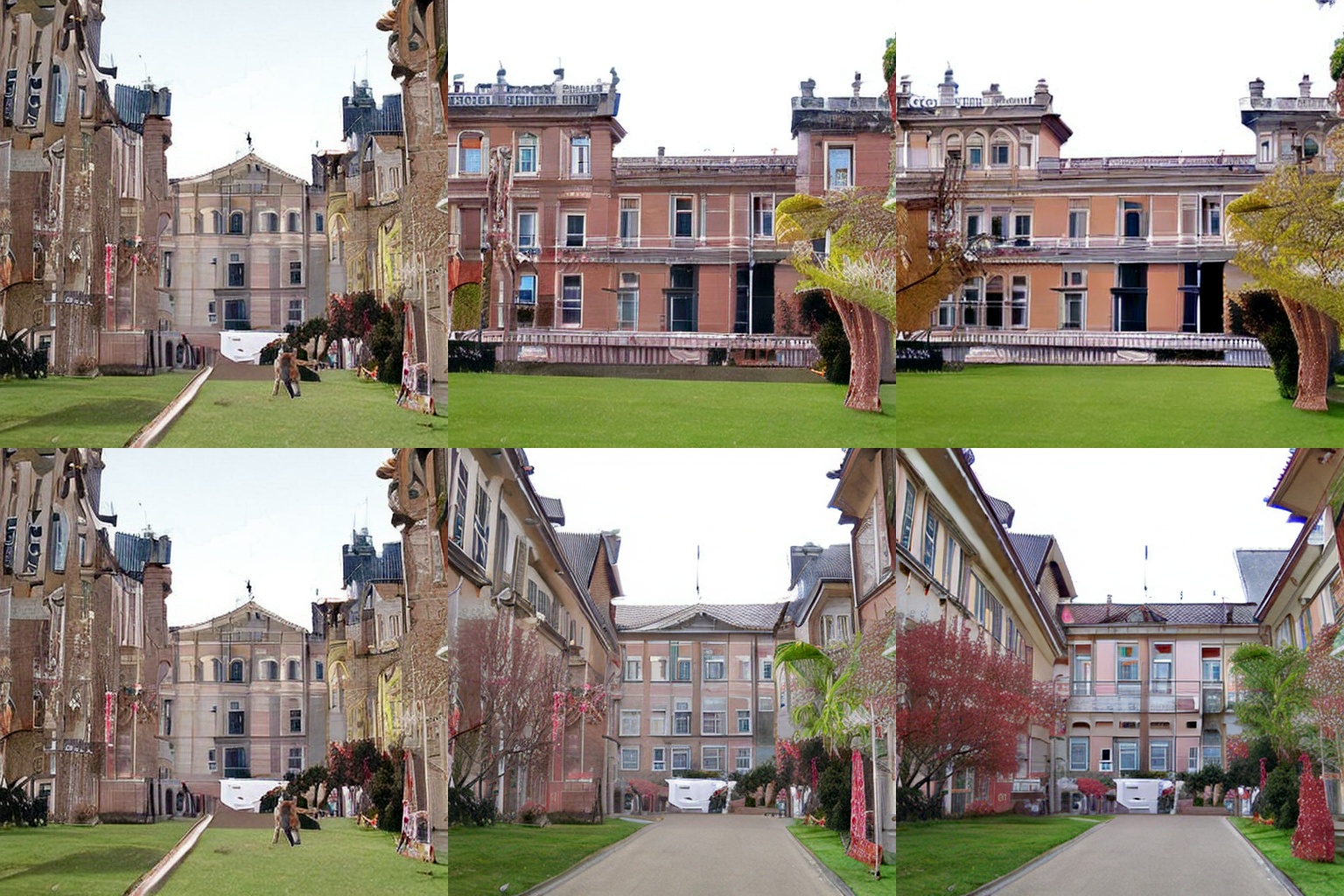}\\[0.8mm]
\rotatebox{90}{\makebox[\ww][c]{Ours}\makebox[\ww][c]{CFG}}\hfill%
\includegraphics[width=\w]{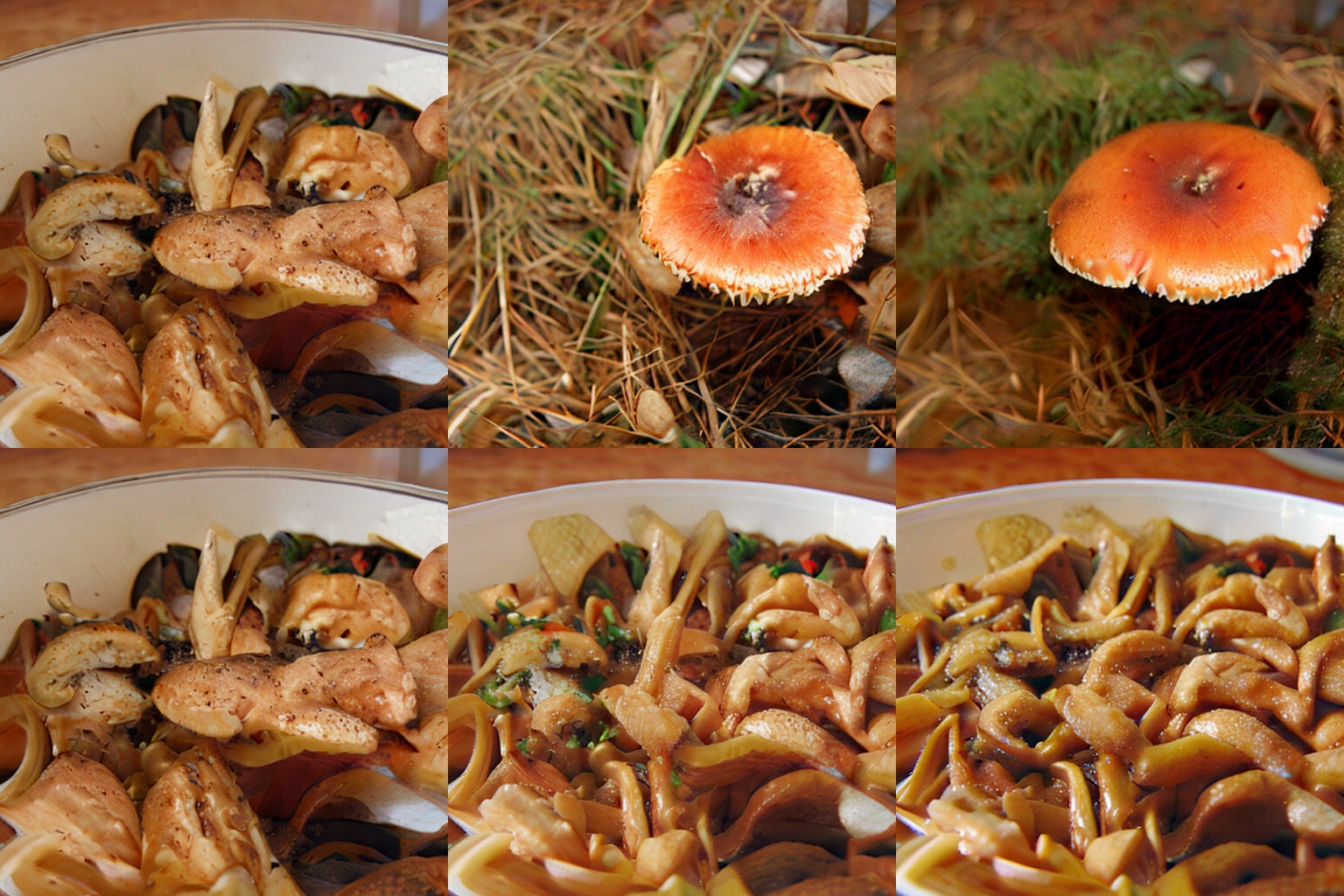}\hfill%
\includegraphics[width=\w]{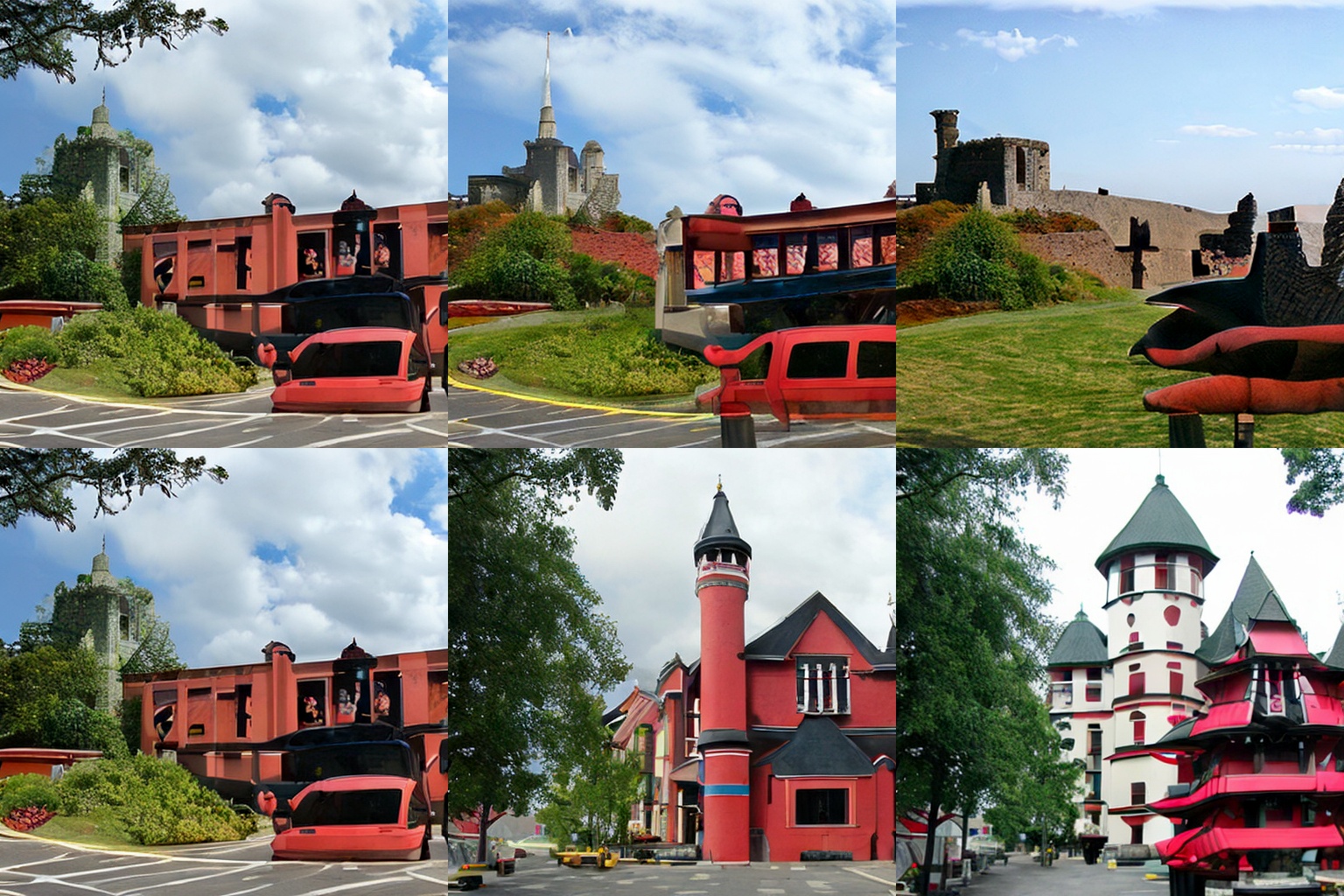}%
\caption{\label{figImageNetMain}%
Example results for the \emph{Tree frog, Palace, Mushroom, Castle} classes of ImageNet-512 using EDM2-S.
Guidance weight increases to the right; rows are classifier-free guidance and our method.
}\vspace*{0mm}%
\end{figure}
}%

\newcommand{\figDeepFloydAppendix}{%
\usetikzlibrary {arrows.meta}
\begin{figure}[H]%
\def\w{0.935\linewidth}%
\centering\footnotesize%
Interpolation between CFG and our method\\[-1mm]%
\begin{tikzpicture}
  \draw[use as bounding box, draw=none] (0,-0.1) rectangle (13,0.1);
  \draw[arrows = {Stealth[scale=1.25]-Stealth[scale=1.25]}] (1,0) -- (12,0);
\end{tikzpicture}\\[0mm]%
\rotatebox{90}{\hs{30}Increasing CFG weight}\hfill%
\begin{tikzpicture}
  \draw[use as bounding box, draw=none] (0,0) rectangle (0,8.5);
  \draw[arrows = {-Stealth[scale=1.25]}] (0,8.25) -- (0,0.4);
\end{tikzpicture}\hfill\hfill%
\includegraphics[width=\w]{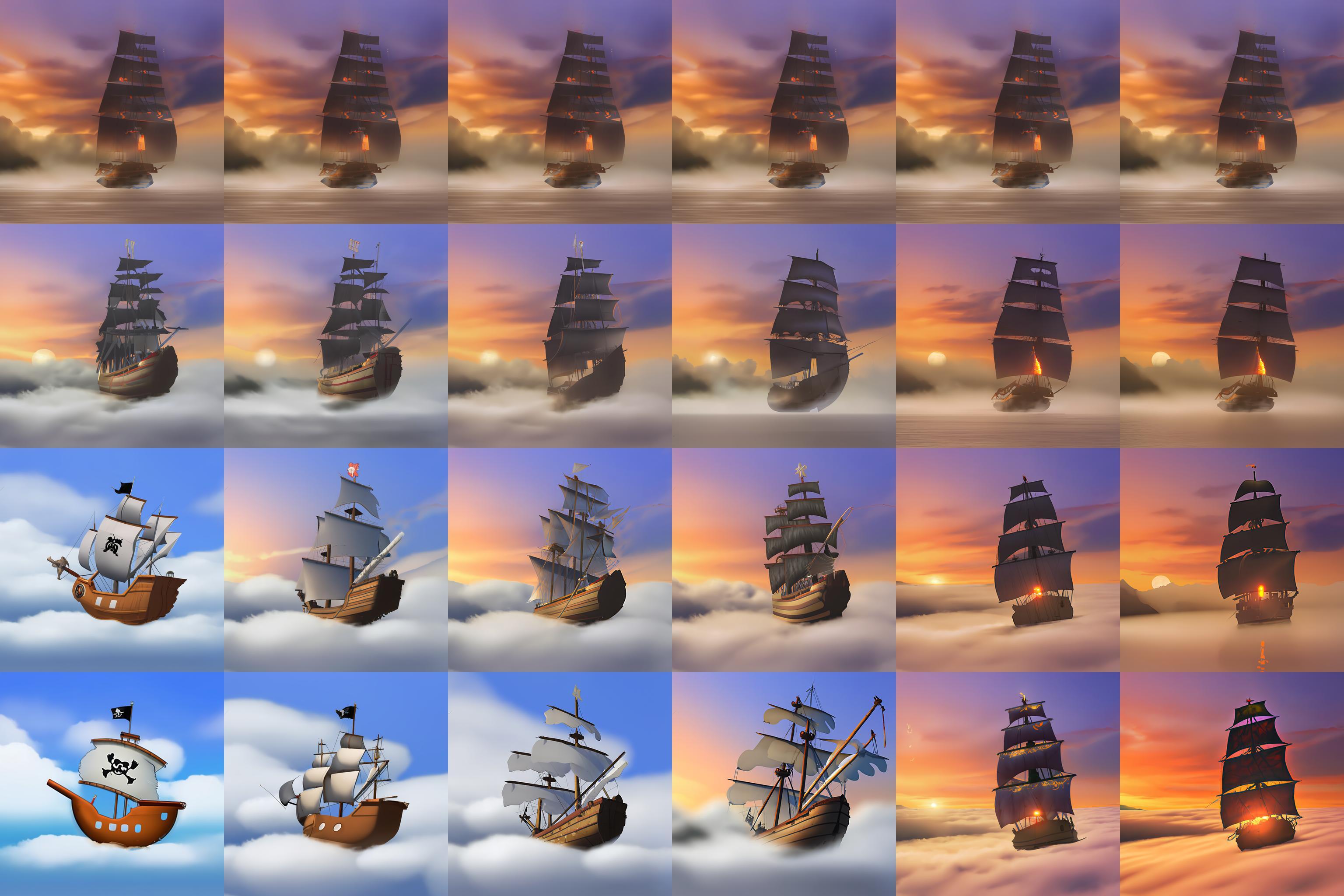}\hfill\hfill%
\begin{tikzpicture}
  \draw[use as bounding box, draw=none] (0,0) rectangle (0,8.5);
  \draw[arrows = {-Stealth[scale=1.25]}] (0,8.25) -- (0,0.4);
\end{tikzpicture}\hfill%
\rotatebox{90}{\hs{15}Increasing guidance weight for our method}\\%
\emph{``A pirate ship sailing through clouds instead of the ocean''}\\[1.5mm]
\rotatebox{90}{\hs{30}Increasing CFG weight}\hfill%
\begin{tikzpicture}
  \draw[use as bounding box, draw=none] (0,0) rectangle (0,8.5);
  \draw[arrows = {-Stealth[scale=1.25]}] (0,8.25) -- (0,0.4);
\end{tikzpicture}\hfill\hfill%
\includegraphics[width=\w]{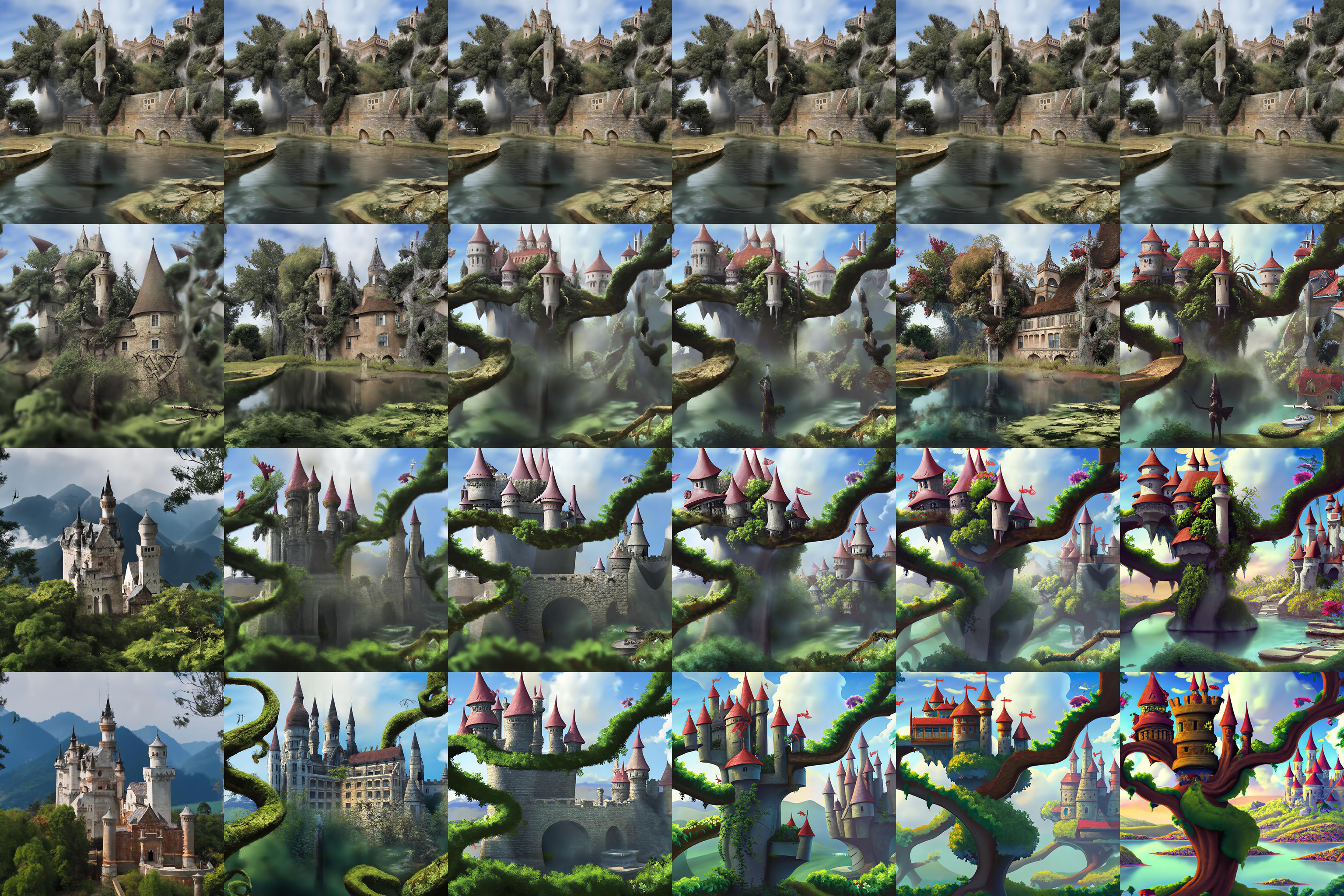}\hfill\hfill%
\begin{tikzpicture}
  \draw[use as bounding box, draw=none] (0,0) rectangle (0,8.5);
  \draw[arrows = {-Stealth[scale=1.25]}] (0,8.25) -- (0,0.4);
\end{tikzpicture}\hfill%
\rotatebox{90}{\hs{15}Increasing guidance weight for our method}\\%
\emph{``A fairytale castle floating in the sky, tethered by ancient vines''}\\[0.5mm]
\caption{\label{figDeepFloydAppendix}%
Additional results for DeepFloyd~IF~\cite{DeepFloyd}.
The rows correspond to guidance weights $w \in \{1, 2, 3, 4\}$.
CFG and our method (XL-sized model guided by M-sized one) on the leftmost and rightmost column, respectively.
The middle columns correspond to blending between the two.
}%
\end{figure}
}%

\definecolor{deepblue}{rgb}{0,0,0.5}
\definecolor{deepred}{rgb}{0.6,0,0}
\definecolor{deepgreen}{rgb}{0,0.5,0}
\DeclareFixedFont{\ttb}{T1}{txtt}{bx}{n}{6.6}
\DeclareFixedFont{\ttm}{T1}{txtt}{m}{n}{6.6}

\lstdefinestyle{styleReproEvalAppendix}{
  language=bash,
  basicstyle=\ttm,
  literate={-}{-}1 {.git}{.git}4,
  keywordstyle=\ttb\color{deepblue},
  stringstyle={},
  commentstyle=\color{deepgreen},
  showstringspaces=false,
  keywords={git, cd, rclone, sed, python, torchrun},
  numbers=left,
  numberstyle=\tiny\color{black!50},
}

\newsavebox{\codeReproEvalAppendix}
\begin{lrbox}{\codeReproEvalAppendix}%
\footnotesize%
\begin{lstlisting}[style=styleReproEvalAppendix]
# Download EDM2 source code.
git clone https://github.com/NVlabs/edm2.git
cd edm2
git checkout 32ecad3

# Patch the sampler so that class labels are passed in to the guiding network.
sed -i 's/gnet(x, t)/gnet(x, t, labels)/g' generate_images.py

# Download the necessary EDM2 models.
rclone copy --progress --http-url https://nvlabs-fi-cdn.nvidia.com/edm2 \
    :http:raw-snapshots/edm2-img512-s/ raw-snapshots/edm2-img512-s/
rclone copy --progress --http-url https://nvlabs-fi-cdn.nvidia.com/edm2 \
    :http:raw-snapshots/edm2-img512-xs/ raw-snapshots/edm2-img512-xs/

# Reconstruct the corresponding post-hoc EMA models.
python reconstruct_phema.py --indir=raw-snapshots/edm2-img512-s --outdir=autoguidance/phema \
    --outprefix=img512-s --outkimg=2147483 --outstd=0.070
python reconstruct_phema.py --indir=raw-snapshots/edm2-img512-xs --outdir=autoguidance/phema \
    --outprefix=img512-xs --outkimg=134217 --outstd=0.125

# Generate 50,000 images using 8 GPUs.
torchrun --standalone --nproc_per_node=8 generate_images.py --seeds=0-49999 --subdirs \
    --outdir=autoguidance/images/img512-s --net=autoguidance/phema/img512-s-2147483-0.070.pkl \
    --gnet=autoguidance/phema/img512-xs-0134217-0.125.pkl --guidance=2.10

# Calculate FID.
python calculate_metrics.py calc --images=autoguidance/images/img512-s \
    --ref=https://nvlabs-fi-cdn.nvidia.com/edm2/dataset-refs/img512.pkl
\end{lstlisting}%
\end{lrbox}

\newcommand{\algReproEvalAppendix}{%
\begin{algorithm}[t]%
\hs{5.6}\usebox{\codeReproEvalAppendix}%
\caption{\label{algReproEvalAppendix}\hs{1.5}\vphantom{$/$}%
Reproducing our FID result for the ``Autoguidance (XS, $T/16$)'' row in Table~\ref{tabMainResults}.
}%
\end{algorithm}
}%

\lstdefinestyle{styleReproTrainAppendix}{
  language=bash,
  basicstyle=\ttm,
  literate={-}{-}1,
  keywordstyle=\ttb\color{deepblue},
  stringstyle={},
  commentstyle=\color{deepgreen},
  showstringspaces=false,
  keywords={torchrun},
  numbers=left,
  numberstyle=\tiny\color{black!50},
}

\newsavebox{\codeReproTrainAppendix}
\begin{lrbox}{\codeReproTrainAppendix}%
\footnotesize%
\begin{lstlisting}[style=styleReproTrainAppendix]
# Unconditional EDM2-S with ImageNet-512, used as a main model in Table 1.
torchrun --nnodes=4 --nproc_per_node=8 train_edm2.py \
    --outdir=autoguidance/train/img512-s-uncond --data=datasets/img512-sd.zip --cond=0 \
    --preset=edm2-img512-s --duration=2048Mi

# Unconditional EDM2-XS with ImageNet-64, used as a guiding model in Table 1.
torchrun --nnodes=4 --nproc_per_node=8 train_edm2.py \
    --outdir=autoguidance/train/img64-xs-uncond --data=datasets/img64.zip --cond=0 \
    --preset=edm2-img64-s --channels=128 --lr=0.0120 --duration=2048Mi

# Conditional EDM2-XS with ImageNet-64, used as a guiding model in Table 1.
torchrun --nnodes=4 --nproc_per_node=8 train_edm2.py \
    --outdir=autoguidance/train/img64-xs --data=datasets/img64.zip --cond=1 \
    --preset=edm2-img64-s --channels=128 --lr=0.0120 --duration=512Mi

# Conditional EDM2-XXS with ImageNet-512, used as a guiding model in Figure 3b.
torchrun --nnodes=4 --nproc_per_node=8 train_edm2.py \
    --outdir=autoguidance/train/img512-xxs --data=datasets/img512-sd.zip --cond=1 \
    --preset=edm2-img512-xs --channels=64 --lr=0.0170 --duration=512Mi
\end{lstlisting}%
\end{lrbox}

\newcommand{\algReproTrainAppendix}{%
\begin{algorithm}[t]%
\hs{5.6}\usebox{\codeReproTrainAppendix}%
\caption{\label{algReproTrainAppendix}\hs{1.5}\vphantom{$/$}%
Training the additional EDM2 models needed in Section~\ref{sec:results}.
}%
\end{algorithm}
}%

\newcommand{\plotBonusGuidanceA}{%
\centering\footnotesize%
\begin{tikzpicture}
\def\colorA{C0}\def\cidA{img512-S-XSu}
\def\colorB{C1}\def\cidB{img512-S-XSu-glo0.28-ghi2.90}
\def\colorC{C2}\def\cidC{img512-S-XS-128Mi}
\begin{axis}[
  width={1.13\linewidth}, height={70mm}, grid={major},
  xmin={1.0}, xmax={3.5}, xmode={linear}, xtick={1.0, 1.5, 2.0, 2.5, 3.0}, xticklabels={\makebox[0mm][r]{$\guidance{=}$}$1.0$, $1.5$, $2.0$, $2.5$, $3.0$},
  ymin={1}, ymax={7}, ymode={linear}, ytick={2, 3, 4, 5, 6, 7}, yticklabels={$2$, $3$, $4$, $5$, $6$, $7$},
  legend pos={north east}, legend cell align={left},
]
\fillbetween[\colorA, opacity=0.2, forget plot]{coordinates {\rawdata{\cidA-fid-glo}}}{coordinates {\rawdata{\cidA-fid-ghi}}};
\fillbetween[\colorB, opacity=0.2, forget plot]{coordinates {\rawdata{\cidB-fid-glo}}}{coordinates {\rawdata{\cidB-fid-ghi}}};
\fillbetween[\colorC, opacity=0.2, forget plot]{coordinates {\rawdata{\cidC-fid-glo}}}{coordinates {\rawdata{\cidC-fid-ghi}}};
\addplot[\colorA, forget plot] coordinates {\rawdata{\cidA-fid-glo}};
\addplot[\colorB, forget plot] coordinates {\rawdata{\cidB-fid-glo}};
\addplot[\colorC, forget plot] coordinates {\rawdata{\cidC-fid-glo}};
\addplot[\colorA, mark=*, forget plot, nodes near coords align={south}, nodes near coords=\datalabel{\data{img512-S-XSu-edm2-fid}}] coordinates {(\rawdata{\cidA-fid-g}, \rawdata{\cidA-fid-precise})};
\addplot[\colorB, mark=*, forget plot, nodes near coords align={south}, nodes near coords=\datalabel{\data{img512-S-XSu-limit-tuomas-fid}}] coordinates {(\rawdata{\cidB-fid-g}, \rawdata{\cidB-fid-precise})};
\addplot[\colorC, mark=*, forget plot, nodes near coords align={north}, nodes near coords=\datalabel{\data{\cidC-fid}}] coordinates {(\rawdata{\cidC-fid-g}, \rawdata{\cidC-fid-precise})};
\addlegendimage{\colorA, mark=*}\addlegendentry{Classifier-free guidance}
\addlegendimage{\colorB, mark=*}\addlegendentry{Guidance interval}
\addlegendimage{\colorC, mark=*}\addlegendentry{Autoguidance (ours)}
\end{axis}
\end{tikzpicture}%
}%

\newcommand{\plotBonusGuidanceB}{%
\centering\footnotesize%
\begin{tikzpicture}
\def\colorA{C0}\def\cidA{img512-S-XSu}
\def\colorB{C1}\def\cidB{img512-S-XSu-glo0.60-ghi5.00}
\def\colorC{C2}\def\cidC{img512-S-XS-128Mi}
\begin{axis}[
  width={1.13\linewidth}, height={70mm}, grid={major},
  xmin={1.0}, xmax={3.5}, xmode={linear}, xtick={1.0, 1.5, 2.0, 2.5, 3.0}, xticklabels={\makebox[0mm][r]{$\guidance{=}$}$1.0$, $1.5$, $2.0$, $2.5$, $3.0$},
  ymin={30}, ymax={130}, ymode={linear}, ytick={40, 60, 80, 100, 120}, yticklabels={$40$, $60$, $80$, $100$, $120$},
  legend pos={north east}, legend cell align={left},
]
\fillbetween[\colorA, opacity=0.2, forget plot]{coordinates {\rawdata{\cidA-dino-glo}}}{coordinates {\rawdata{\cidA-dino-ghi}}};
\fillbetween[\colorB, opacity=0.2, forget plot]{coordinates {\rawdata{\cidB-dino-glo}}}{coordinates {\rawdata{\cidB-dino-ghi}}};
\fillbetween[\colorC, opacity=0.2, forget plot]{coordinates {\rawdata{\cidC-dino-glo}}}{coordinates {\rawdata{\cidC-dino-ghi}}};
\addplot[\colorA, forget plot] coordinates {\rawdata{\cidA-dino-glo}};
\addplot[\colorB, forget plot] coordinates {\rawdata{\cidB-dino-glo}};
\addplot[\colorC, forget plot] coordinates {\rawdata{\cidC-dino-glo}};
\addplot[\colorA, mark=*, forget plot, nodes near coords align={south}, nodes near coords=\datalabel{\data{img512-S-XSu-edm2-dino}}] coordinates {(\rawdata{\cidA-dino-g}, \rawdata{\cidA-dino-precise})};
\addplot[\colorB, mark=*, forget plot, nodes near coords align={south}, nodes near coords=\datalabel{\data{img512-S-XSu-limit-tuomas-dino}}] coordinates {(\rawdata{\cidB-dino-g}, \rawdata{\cidB-dino-precise})};
\addplot[\colorC, mark=*, forget plot, nodes near coords align={north}, nodes near coords=\datalabel{\data{\cidC-dino}}] coordinates {(\rawdata{\cidC-dino-g}, \rawdata{\cidC-dino-precise})};
\addlegendimage{\colorA, mark=*}\addlegendentry{Classifier-free guidance}
\addlegendimage{\colorB, mark=*}\addlegendentry{Guidance interval}
\addlegendimage{\colorC, mark=*}\addlegendentry{Autoguidance (ours)}
\end{axis}
\end{tikzpicture}%
}%

\newcommand{\figBonusGuidance}{%
\begin{figure}[t]%
\def\w{0.5\linewidth}%
\begin{subfigure}[b]{\w}\plotBonusGuidanceA\vspace*{-.5ex}\caption{\small FID}\end{subfigure}%
\begin{subfigure}[b]{\w}\plotBonusGuidanceB\vspace*{-.5ex}\caption{\small \DINO{}}\end{subfigure}%
\vspace*{-.25ex}%
\caption{\label{figBonusGuidance}%
Sweep over guidance weight $\guidance$ using EDM2-S on ImageNet-512.
The optimal EMA length was searched separately for the three methods and two metrics (FID and \DINO{}).
}%
\end{figure}
}%

\newcommand{\figVariationComparison}{%
\begin{figure}[t]%
\def\w{0.327\linewidth}%
\def\ww{34mm}%
\centering\footnotesize%
\rotatebox{90}{\makebox[\ww][c]{Ours}\makebox[\ww][c]{\ \ CFG}}\hfill%
\includegraphics[width=\w]{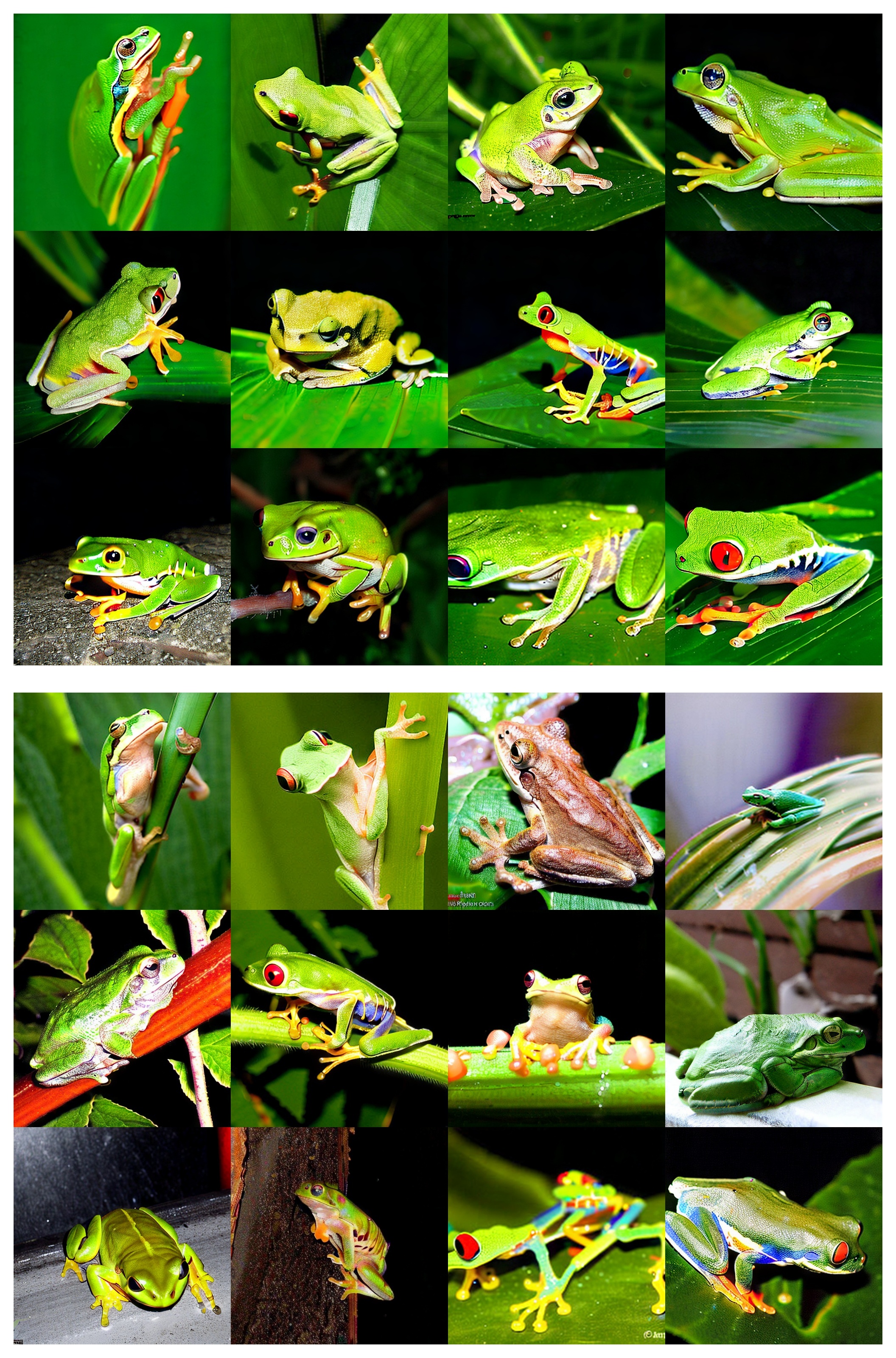}%
\includegraphics[width=\w]{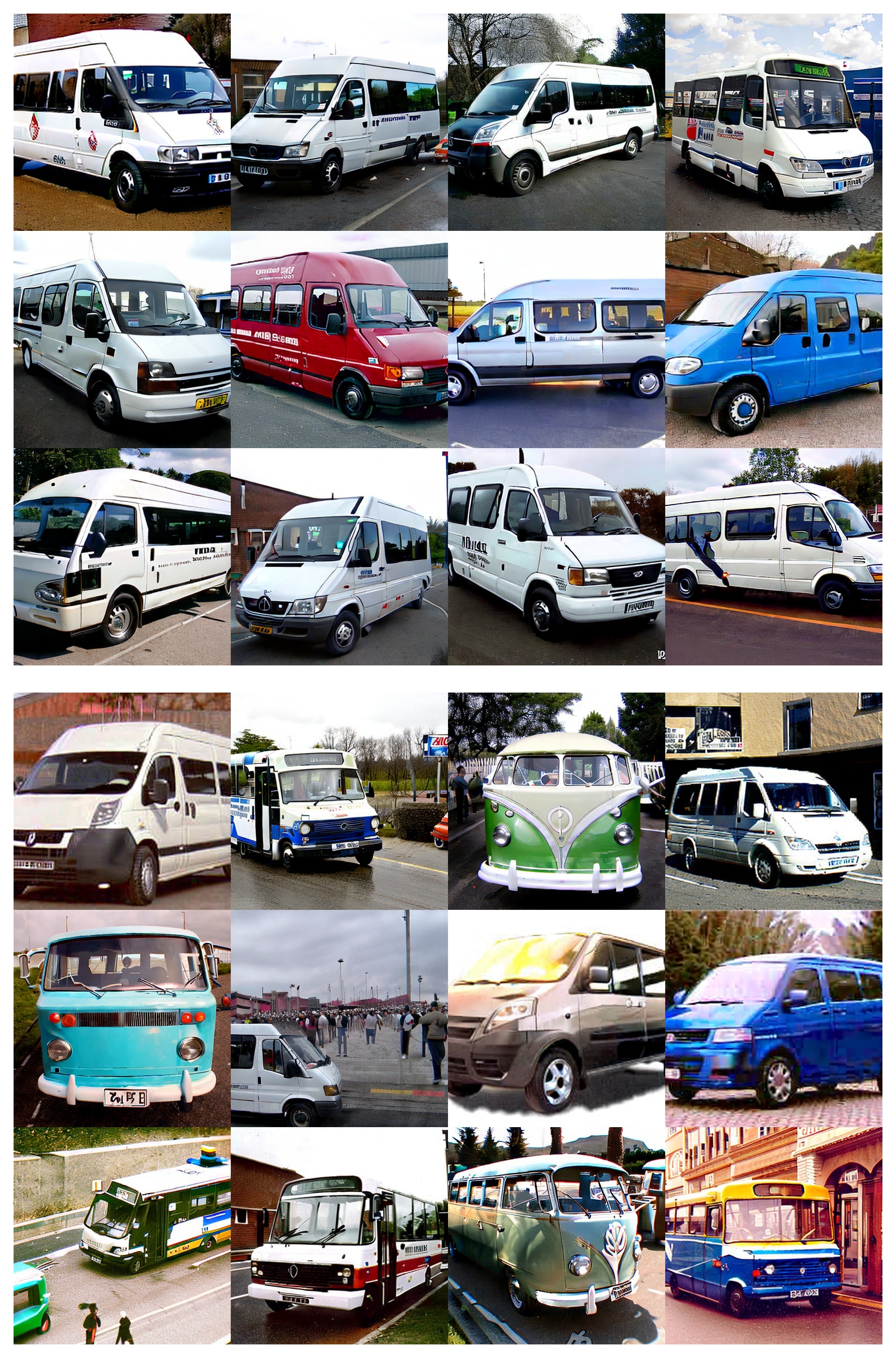}%
\includegraphics[width=\w]{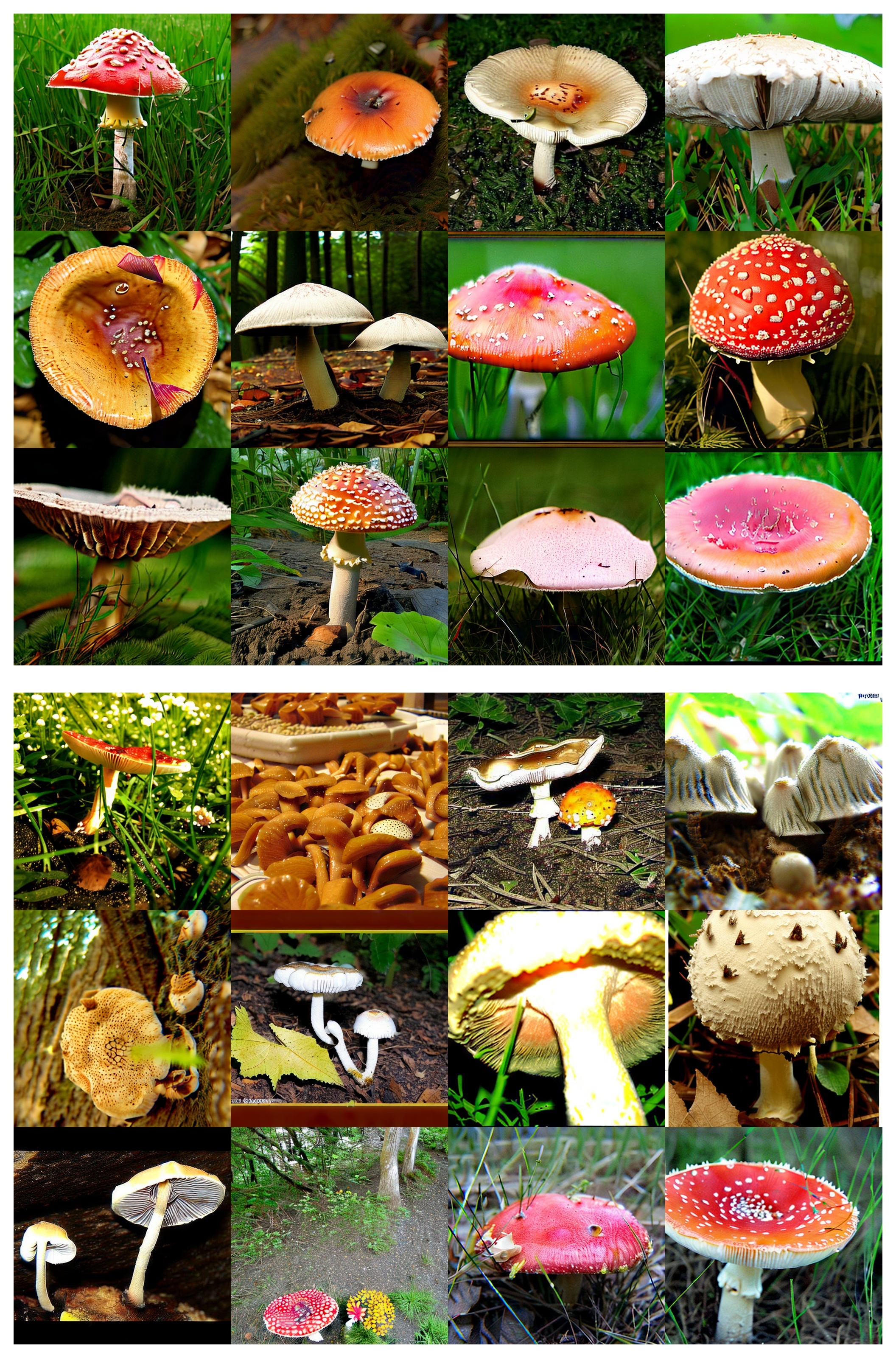}%
\caption{\label{figVariationComparison}%
Resulting variation for CFG and our method in \emph{Tree frog, Minibus, Mushroom} classes of ImageNet-512 using EDM2-S (\DINO{}-optimized).
In this image, we have exaggerated the amount of guidance ($w=4$) to make its effect on variation more clearly visible. 
This causes excessive saturation and other image artifacts, but clearly shows that CFG steers towards canonical templates, while our method preserves much greater variation.
}%
\end{figure}
}%

\begin{abstract}

The primary axes of interest in image-generating diffusion models are image quality, the amount of variation in the results, and how well the results align with a given condition, e.g., a class label or a text prompt.
The popular classifier-free guidance approach uses an unconditional model to guide a
conditional model, leading to simultaneously better prompt alignment and higher-quality images at the cost of reduced variation. 
These effects seem inherently entangled, and thus hard to control.
We make the surprising observation that it is possible to obtain disentangled control over image quality without compromising the amount of variation by guiding generation using a smaller, less-trained version of the model itself rather than an unconditional model.
This leads to significant improvements in ImageNet generation, setting record FIDs of 1.01 for $64{\times}64$ and 1.25 for $512{\times}512$, using publicly available networks.
Furthermore, the method is also applicable to unconditional diffusion models, drastically improving their quality.

\end{abstract}

\section{Introduction}

Denoising diffusion models~\cite{Ho2020,SohlDickstein2015,Song2020ddim,Song2019gradients,Song2021sde} generate synthetic
  images by reversing a stochastic corruption process.
Essentially, an image is revealed from pure noise by denoising it little by little in successive steps.
A neural network that implements the denoiser (equivalently~\cite{Vincent11}, the score function~\cite{Hyvarinen05})
  is a central design element, and various architectures have been
  proposed (e.g.,~\cite{Bao2022,Crowson2024,Dhariwal2021,Hoogeboom2023b,Jabri2023,Karras2024edm2,Peebles2022}).
Equally important are the details of the multi-step denoising process that corresponds mathematically to solving an
  ordinary~\cite{Nichol2021a,Song2020ddim} or a stochastic~\cite{Song2021sde} differential equation, for which
  many different parameterizations, solvers, and step schedules have been
  evaluated~\cite{Jolicoeur2021,Karras2022elucidating,Liu2022pseudo,Sabour2024,Zhang2022exp}.
To control the output image, the denoiser is typically conditioned on a class label,
  an embedding of a text prompt, or some other form of conditioning input~\cite{Nichol2021b,dalle2,imagen,Zhang2023}.

The training objective of a diffusion model aims to cover the entire (conditional) data distribution. This causes problems in low-probability regions: The model gets heavily penalized for not representing them, but it does not have enough data to learn to generate good images corresponding to them.
Classifier-free guidance (CFG)~\cite{Ho2021classifierfree} has become the standard method for
  ``lowering the sampling temperature'', i.e., focusing the generation on well-learned high-probability
  regions.
By training a denoiser network to operate in both conditional and unconditional setting,
  the sampling process can be steered away from the unconditional result\,---\,%
  in effect, the unconditional generation task specifies a result to \emph{avoid}.
This results in better prompt alignment and improved image quality, where the former effect
  is due to CFG implicitly raising the conditional part of the probability density to a power
  greater than one~\cite{guidance}.

However, CFG has drawbacks that limit its usage as a general low-temperature
  sampling method.
First, it is applicable only for conditional generation, as the guidance signal is based on
  the difference between conditional and unconditional denoising results.
Second, because the unconditional and conditional denoisers are trained to solve
  a different task, the sampling trajectory can overshoot the desired conditional distribution,
  which leads to skewed and often overly simplified image
  compositions~\cite{Kynkaanniemi2024guidance}.
Finally, the prompt alignment and quality improvement effects cannot be controlled separately,
  and it remains unclear how exactly they relate to each other.

In this paper, we provide new insights into why CFG improves image quality
  and show how this effect can be separated out into a novel method that we call \emph{autoguidance}.
Our method does not suffer from the task discrepancy problem because we use an
  inferior version of the main model itself as the guiding model, with unchanged conditioning.
This guiding model can be obtained by simply limiting, e.g., model capacity and/or training time.
We validate the effectiveness of autoguidance in various synthetic test cases as well as
  in practical image synthesis in class-conditional and text-conditional settings.
In addition, our method enables guidance for unconditional synthesis.
In quantitative tests, the generated image distributions are improved considerably when measured using
  FID~\cite{Heusel2017} and \DINO{}~\cite{stein2023exposing} metrics, setting new
  records in ImageNet-512 and ImageNet-64 generation.

Our implementation and pre-trained models are available at {\footnotesize\url{https://github.com/NVlabs/edm2}}

\section{Background}
\label{sec:background}

\paragraph{Denoising diffusion.}
Denoising diffusion generates samples from a distribution $p_\text{data}(\boldx)$ by iteratively denoising
  a sample of pure white noise, such that a noise-free random data sample is gradually revealed~\cite{Ho2020}.
The idea is to consider heat diffusion of $p_\text{data}(\boldx)$ into a sequence of increasingly smoothed
  densities \mbox{$p(\boldx; \sigma) = p_\text{data}(\boldx) \ast \mathcal{N}(\boldx; \boldzero, \sigma^2 \boldI)$}.
For a large enough $\sigma_\text{max}$, we have
  \mbox{$p(\boldx; \sigma_\text{max}) \approx \mathcal{N}(\boldx; \boldzero, \sigma_\text{max}^2 \boldI)$},
  from which we can trivially sample by drawing normally distributed white noise.
The resulting sample is then evolved backward towards low noise levels by a probability flow ODE~\cite{Karras2022elucidating,Song2020ddim,Song2021sde}
\begin{align}
  \mathrm{d}\boldx_\sigma &\,=\, -\sigma \nablaxs \log p(\boldx_\sigma; \sigma) ~ \mathrm{d}\sigma \label{eq:ode}
\end{align}
that maintains the property \mbox{$\boldx_\sigma \sim p(\boldx_\sigma; \sigma \big)$} for every $\sigma \in [0, \sigma_\text{max}]$.
Upon reaching \mbox{$\sigma = 0$}, we obtain
  $\boldx_0 \sim p(\boldx_0; 0) = p_\text{data}(\boldx_0)$ as desired.

In practice, the ODE is solved numerically by stepping along the trajectory defined by Equation~\ref{eq:ode}.
This requires evaluating the so-called score function~\cite{Hyvarinen05}
  \mbox{$\nablax \log p(\boldx; \sigma)$} for a given sample $\boldx$ and noise level $\sigma$ at each step.
Rather surprisingly, we can approximate this vector using a neural network $\dnet(\boldx; \sigma)$
  parameterized by weights $\theta$ trained for the denoising task
\begin{align}
  \theta &\,=\, \argmin_\theta \mathbb{E}_{\boldy \sim p_\text{data}, \sigma \sim p_\text{train}, \mathbf{n} \sim \mathcal{N}(\boldzero, \sigma^2 \boldI)} \lVert \dnet(\boldy+\boldn;\sigma) - \yy \rVert^2_2
  \text{,}
\end{align}
where $p_\text{train}$ controls the noise level distribution during training.
Given $\dnet$, we can estimate $\nablax \log p(\boldx; \sigma) \approx (\dnet(\boldx; \sigma) - \boldx) / \sigma^2$,
  up to approximation errors due to, e.g., finite capacity or training time~\cite{Karras2022elucidating,Vincent11}.
As such, we are free to interpret the network as predicting either a denoised sample or a score vector,
  whichever is more convenient for the analysis at hand.
Many reparameterizations and practical ODE solvers are possible,
  as enumerated by Karras et al.~\cite{Karras2022elucidating}.
We follow their recommendations, including the schedule \mbox{$\sigma(t) = t$}
  that lets us parameterize the ODE directly via noise level $\sigma$
  instead of a separate time variable $t$.

In most applications, each data sample $\boldx$ is associated with a label $\boldc$,
  representing, e.g., a class index or a text prompt.
At generation time, we control the outcome by choosing a label $\boldc$
  and seeking a sample from the conditional distribution $p(\boldx | \boldc; \sigma)$ with $\sigma = 0$.
In practice, this is achieved by training a denoiser network $\dnet(\boldx; \sigma, \boldc)$
  that accepts $\boldc$ as an additional conditioning input.

\paragraph{Classifier-free guidance.} For complex visual datasets, the generated images often fail to reproduce the clarity of the training images due to approximation errors made by finite-capacity networks. A broadly used trick called \emph{classifier-free guidance} (CFG)~\cite{Ho2021classifierfree} pushes the samples towards higher likelihood of the class label, sacrificing variety for ``more canonical'' images that the network appears to be better capable of handling.

In a general setting, guidance in a diffusion model involves two denoiser networks
  $\drud(\boldx; \sigma, \boldc)$ and $\dmain(\boldx; \sigma, \boldc)$.
The guiding effect is achieved by \emph{extrapolating} between the two denoising results by a factor $\guidance$:
\begin{align}
  \dguid(\boldx; \sigma, \boldc) &\,=\, \guidance \dmain(\boldx; \sigma, \boldc) + (1-\guidance) \drud(\boldx; \sigma, \boldc) \label{eq:generalized}
  \text{.}
\end{align}
Trivially, setting \mbox{$\guidance = 0$} or \mbox{$\guidance = 1$} recovers the output
  of $\drud$ and $\dmain$, respectively, while choosing \mbox{$\guidance > 1$} over-emphasizes the output of $\dmain$.
Recalling the equivalence of denoisers and scores~\cite{Vincent11}, we can write
\begin{align}
  \dguid(\boldx; \sigma, \boldc) &\,\approx\, \boldx + \sigma^2 \nablax \log \underbrace{\left( \prud(\boldx | \boldc; \sigma) \left[\frac{\pmain(\boldx | \boldc; \sigma)}{\prud(\boldx | \boldc; \sigma)}\right]^\guidance \right)}_{{\propto:~\pguid(\boldx|\boldc;\,\sigma)}} \label{eq:guidedpdf}
  \text{.}
\end{align}
Thus, guidance grants us access to the score of the density \mbox{$\pguid(\boldx | \boldc; \sigma)$} implied in the parentheses.
This score can be further written as~\cite{guidance,Ho2021classifierfree}
\begin{align}
  \nablax \log \pguid(\boldx | \boldc; \sigma) &\,=\, \nablax \log \pmain(\boldx | \boldc; \sigma)  + (\guidance-1) \nablax \log \frac{\pmain(\boldx | \boldc; \sigma)}{\prud(\boldx | \boldc; \sigma)} \label{eq:guidedscore}
  \text{.}
\end{align}
Substituting this expression into the ODE of Equation~\ref{eq:ode},
  this yields the standard evolution for generating images from $\pmain$,
  plus a perturbation that increases (for $\guidance > 1$) the ratio of $\pmain$ and $\prud$ as evaluated at the sample.
The latter can be interpreted as increasing the likelihood that
  a hypothetical classifier would attribute for the sample having come from density $\pmain$ rather than $\prud$.

In CFG, we train an auxiliary \emph{unconditional} denoiser $D_\theta(\boldx; \sigma)$ to denoise the
  distribution $p(\boldx; \sigma)$ marginalized over $\boldc$, and use this as $\drud$.
In practice, this is typically~\cite{Ho2021classifierfree} done using the same network $D_\theta$ with an empty conditioning label, setting
  $\drud \coloneqq D_\theta(\boldx; \sigma, \emptyset)$ and $\dmain \coloneqq D_\theta(\boldx; \sigma, \boldc)$.
By Bayes' rule, the extrapolated score vector becomes
  $%
  \nablax \log p(\boldx | \boldc; \sigma) + (\guidance-1) \nablax \log p(\boldc | \boldx; \sigma)$.
During sampling, this guides the image to more strongly align with the specified class $\boldc$.

It would be tempting to conclude that solving the diffusion ODE with the score function of Equation~\ref{eq:guidedscore}
  produces samples from the data distribution specified by $\pguid(\boldx | \boldc; 0)$.
Unfortunately this is \emph{not} the case, because
  $\pguid(\boldx | \boldc; \sigma)$ does not represent a valid heat diffusion of $\pguid(\boldx | \boldc; 0)$ \cite{Zheng2024}.
Therefore, solving the ODE does not, in fact, follow the density.
Instead, the samples are blindly pushed towards higher values of the implied density
  at each noise level during sampling.
This can lead to distorted sampling trajectories, greatly exaggerated truncation,
  and mode dropping in the results~\cite{Kynkaanniemi2024guidance},
  as well as over-saturation of colors~\cite{imagen}.
Nonetheless, the improvement in image quality is often remarkable,
  and high guidance values are commonly used despite the drawbacks (e.g.,~\cite{imagenvideo,Poole2023,dalle2,imagen}).

\section{Why does CFG improve image quality?}
\label{sec:whycfgimprove}

We begin by identifying the mechanism by which CFG improves image quality instead of only affecting prompt alignment.
To illustrate why unguided diffusion models often produce unsatisfactory images, and how CFG remedies the problem,
  we study a 2D toy example where a small-scale denoiser network is trained to perform conditional diffusion in a
  synthetic dataset (Figure~\ref{figToyExample}).
The dataset is designed to exhibit low local dimensionality (i.e., highly anisotropic and narrow support) and
  hierarchical emergence of local detail upon noise removal.
These are both properties that can be expected from the actual manifold of realistic images \cite{brown2023,pope2021}.
For details of the setup, see Appendix~\ref{app:toyDetails}.

\paragraph{Score matching leads to outliers.}
Compared to sampling directly from the underlying distribution (Figure~\ref{fig:toyGT}),
  the unguided diffusion in Figure~\ref{fig:toyNog} produces a large number of
  extremely unlikely samples outside the bulk of the distribution.
In the image generation setting, these would correspond to unrealistic and broken images.

We argue that the outliers stem from the limited capability of the score network combined
  with the score matching objective.
It is well known that maximum likelihood (ML) estimation leads to a ``conservative'' fit of the
  data distribution~\cite{Bishop1995book} in the sense that the model attempts to cover all training samples.
This is because the underlying Kullback--Leibler divergence incurs extreme penalties
  if the model severely underestimates the likelihood of any training sample.
While score matching is generally not equal to ML estimation, they are closely related~\cite{Ho2020,Lyu2009,Song2021sde}
  and appear to exhibit broadly similar behavior.
For example, it is known that for a multivariate Gaussian model,
  the optimal score matching fit coincides with the ML estimate~\cite{Hyvarinen05}.
Figures~\ref{fig:toyPMain} and~\ref{fig:toyPRudder} show the learned score field and implied density in our toy example for two models of different capacity at an intermediate noise level.
The stronger model envelops the data \mbox{more tightly, while the weaker model's density is more spread out.}

\figToyExample
\figToyDetail

From the perspective of image generation, a tendency to cover the entire training data becomes a problem:
  The model ends up producing strange and unlikely images from the data distribution's extremities
  that are not learnt accurately but included just to avoid the high loss penalties.
Furthermore, during training, the network has only seen real noisy images as inputs, and during sampling it
  may not be prepared to deal with the unlikely samples it is handed down from the higher noise levels.

\paragraph{CFG eliminates outliers.}
The effect of applying classifier-free guidance during generation is demonstrated in Figure~\ref{fig:toyCFG}.
As expected, the samples avoid the class boundary (i.e., there are no samples in the vicinity of the gray area), and entire branches of the distribution are dropped. We also observe a second phenomenon, where the samples have been pulled in towards the core of the manifold, and away from the low-probability intermediate regions.
Seeing that this eliminates the unlikely outlier samples, we attribute the image quality improvement to it.
However, mere boosting of the class likelihood does not explain this increased concentration.

We argue that this phenomenon stems from a quality difference between the
  conditional and unconditional denoiser networks.
The denoiser $\drud$ faces a more difficult task of the two:
  It has to generate from all classes at once,
  whereas $\dmain$ can focus on a single class for any specific sample.
Given the more difficult task,
  and typically only a small slice of the training budget,
  the network $\drud$ attains a worse fit to the data.%
\footnote{The visual quality difference is obvious if we simply inspect the
          unconditional images generated by current large-scale models.
          Furthermore, the unconditional case tends to work so poorly that the corresponding quantitative numbers are hardly ever reported.
          The EDM2-S model~\cite{Karras2024edm2} trained with ImageNet-512, for example, yields a FID of~\data{img512-S-nog-edm2-fid} in the class-conditional setting
          and \data{img512-Su-nog-fid} in the unconditional setting.}
This difference in accuracy is apparent in respective plots of the learned densities in Figures~\ref{fig:toyPMain} and~\ref{fig:toyPRudder}.

From our interpretation in Section~\ref{sec:background},
  it follows that CFG is not only boosting the likelihood of the sample having come from the
  class $\boldc$, but \emph{also} that of having come from the higher-quality implied distribution.
Recall that guidance boils down to an additional force (Equation~\ref{eq:guidedscore}) that pulls the samples towards higher values of
  $\log \left[\pmain(\boldx | \boldc; \sigma) / \prud(\boldx | \boldc; \sigma)\right]$.
Plotting this ratio for our toy example in Figure~\ref{fig:toyPRatio},
  along with corresponding gradients that guidance contributes to the ODE vector field,
  we see that the ratio generally decreases with distance from the manifold due to the denominator
  $p_0$ representing a more spread-out distribution, and hence falling off slower than the numerator $p_1$.
Consequently, the gradients point inward towards the data manifold.
Each contour of the density ratio corresponds to a specific likelihood that a
  hypothetical classifier would assign on a sample being drawn from $p_1$ instead of $p_0$.
Because the contours roughly follow the local orientation and branching of the data manifold,
  pushing samples deeper into the ``good side'' concentrates them at the manifold.%
\footnote{
  Discriminator guidance~\cite{Kim2023} trains an explicit classifier to discriminate between
    the generated samples and noisy training samples at each noise level and uses its log-gradient to guide the sampling.
  Our analysis is not applicable in this situation; in CFG the task is implicit and
    the distinction is between $p_1$ and $p_0$.
}

\paragraph{Discussion.}
We can expect the two models to suffer from inability to fit at similar places, but to a different degree.
The predictions of the denoisers will disagree more decisively in these regions. As such, CFG can be seen as a form of adaptive truncation that identifies when a sample is likely to be under-fit and pushes it towards the general direction of better samples.
Figures~\ref{fig:toyTMain} and~\ref{fig:toyTGuid} show the effect over the course of generation: The truncation ``overshoots'' the correction and produces a narrower distribution than the ground truth, but in practice this does not appear to have an adverse effect on the images.

In contrast, a naive attempt at achieving this kind of truncation\,---\,inspired by, e.g., the truncation trick in GANs~\cite{Brock2018,Marchesi2017} or lowering temperature in generative language models\,---\,would counteract the smoothing by uniformly lengthening the score vectors by a factor $\guidance > 1$.
This is illustrated in Figure~\ref{fig:toyNaive}, where the samples are indeed concentrated in high-probability regions,
  but in an isotropic fashion that leaves the outer branches empty.
In practice, images generated this way tend to show reduced variation, oversimplified details, and monotone texture.

\section{Our method}

We propose to isolate the image quality improvement effect by directly guiding a high-quality model $\dmain$ with a poor model $\drud$ trained on the \emph{same task, conditioning, and data distribution}, but suffering from certain additional degradations, such as low capacity and/or under-training. We call this procedure \emph{autoguidance}, as the model is guided with an inferior version of itself.

In the context of our 2D toy example, this turns out to work surprisingly well.
Figure~\ref{fig:toyOurs} demonstrates the effect of using a smaller $\drud$ with fewer training iterations.
As desired, the samples are pulled close to the distribution without systematically dropping any part of it.

To analyze why this technique works, recall that under limited model capacity, score matching tends to
  over-emphasize low-probability (i.e., implausible and under-trained) regions of the data distribution.
Exactly where and how the problems appear depend on various factors such as network architecture, dataset, training
  details, etc., and we cannot expect to identify and characterize the specific issues a priori.
However, we can expect a weaker version of the \emph{same} model to make broadly similar errors in
  the same regions, only stronger.
Autoguidance seeks to identify and reduce the errors made by the stronger model by measuring its difference
  to the weaker model's prediction, and boosting it.
When the two models agree, the perturbation is insignificant, but when they disagree, the difference
  indicates the general direction towards better samples.

As such, we can expect autoguidance to work if the two models suffer from degradations that are compatible
  with each other.
Since any $\dmain$ can be expected to suffer from, e.g., lack of capacity and lack of training\,---\,%
  at least to some degree\,---\,it makes sense to choose $\drud$ so that it further exacerbates these aspects.

In practice, models that are trained separately or for a different number of iterations differ not only in
  accuracy of fit, but also in terms of random initialization, shuffling of the training data, etc.
For guidance to be successful, the quality gap should be large enough to make the systematic spreading-out
  of the density outweigh these random effects.

\paragraph{Study on synthetic degradations.}
To validate our hypothesis that the two models must suffer from the same kind of degradations,
  we perform a controlled experiment using synthetic corruptions applied to a well-trained real-world image diffusion model. We create the main and guiding networks, $\dmain$ and $\drud$, by applying different degrees of a synthetic corruption to the base model. This construction allows us to use the untouched base model as grounding when measuring the FID effect of the various combinations of corruptions applied to $\dmain$ and $\drud$.
We find that as long as the degradations are compatible, autoguidance largely undoes the damage caused by the corruptions:

\begin{itemize}
\item \textbf{Base model:} As the base model, we use EDM2-S trained on ImageNet-512 without dropout (FID~=~\data{img512-S-nog-edm2-fid}).
\item \textbf{Dropout:}
  We construct $\dmain$ by applying 5\% dropout to the base model in a post-hoc fashion (FID~=~\data{\cidSynDropout-g1.00-fid}),
  and $\drud$ by applying 10\% dropout to the base model (FID~=~\data{\cidSynDropout-g0.00-fid}).
  Applying autoguidance, we reach the best result (FID~=~\data{\cidSynDropout-fid}) with $\guidance = \data{\cidSynDropout-fid-g}$,
    matching the base model's FID.
\item \textbf{Input noise:}
  We construct $\dmain$ by modifying the base model to add noise to the input images so that
  their noise level is increased by 10\% (FID~=~\data{\cidSynInputNoise-g1.00-fid}).
  The $\sigma$ conditioning input of the denoiser is adjusted accordingly.
  The guiding model $\drud$ is constructed similarly,
  but with a noise level increase of~20\% (FID~=~\data{\cidSynInputNoise-g0.00-fid}).
  Applying autoguidance, we reach the best result (FID~=~\data{\cidSynInputNoise-fid}) with $\guidance = \data{\cidSynInputNoise-fid-g}$,
    again matching the base model's FID.
\item \textbf{Mismatched degradations:}
  If we corrupt $\dmain$ by dropout and $\drud$ by input noise, or vice versa, 
  guidance does not improve the results at all;
  in these cases, the best FID is obtained by setting $\guidance = 1$, i.e., by disabling guidance and using the less corrupted $\dmain$ exclusively.
\end{itemize}

While this experiment corroborates our main hypothesis, we do not suggest that guiding with these
  synthetic degradations would be useful in practice.
A realistic diffusion model will not suffer from these particular degradations,
  so creating a guiding model by introducing them would not yield consistent truncation towards the data manifold.

\section{Results}
\label{sec:results}

\tabMainResults
\afterpage{\figMainPlots}

Our primary evaluation is carried out using ImageNet (ILSVRC2012) \cite{Deng2009imagenet} at two resolutions: $512{\times}512$ and $64{\times}64$.
For ImageNet-512 we use latent diffusion \cite{Rombach2021highresolution}, while ImageNet-64 works directly on RGB pixels.
We take the current state-of-the-art diffusion model EDM2~\cite{Karras2024edm2} as our baseline.\footnote{\url{https://github.com/NVlabs/edm2}}
We use the EDM2-S and EDM2-XXL models with default sampling parameters: 32 deterministic steps with a $2^{\text{nd}}$ order Heun sampler \cite{Karras2022elucidating}.
For most setups, a pre-trained model is publicly available, and in the remaining cases we train the models ourselves using the official implementation (Appendix~\ref{app:newModels}).

We use two degradations for the guiding model: shorter training time and reduced capacity compared to the main model.
We obtain the best results by having both of these enabled.
With EDM2-S, for example, we use an XS-sized guiding model that receives $1/16$\textsuperscript{th} of the training iterations of the main model.
We ablate the relative importance of the degradations as well as the sensitivity to these specific choices in Section~\ref{sec:ablation}.
As the EDM2 networks are known to be sensitive to the guidance weight and EMA length~\cite{Karras2024edm2}, we search the optimal values for each case using a grid search.

Table~\ref{tabMainResults} shows that our method improves FID~\cite{Heusel2017} and \DINO{} \cite{stein2023exposing} considerably. 
Using the small model (EDM2-S) in ImageNet-512, our autoguidance improves FID from \data{img512-S-nog-edm2-fid} to \data{\cidBestS-fid}.
This beats the \data{img512-S-XSu-limit-tuomas-fid} achieved by the concurrently proposed CFG + Guidance Interval \cite{Kynkaanniemi2024guidance}, and is the best result reported for this dataset regardless of the model size.
Using the largest model (EDM2-XXL) further improves the record to \data{\cidBestXXL-fid}.
The \DINO{} records are similarly improved, with the large model lowering the record from \data{img512-XXL-XSu-limit-tuomas-dino} to \data{\cidBestXXL-dino}.
In ImageNet-64, the improvement is even larger; in this dataset, we set the new record FID and \DINO{} 
  of \data{\cidBestSixtyFour-fid} and \data{\cidBestSixtyFour-dino}, respectively.

A particular strength of autoguidance is that it can be applied to unconditional models as well.
While conditional ImageNet generation may be getting close to saturation, the unconditional results remain surprisingly poor.
EDM2-S achieves a FID of \data{img512-Su-nog-fid} in the unconditional setting, indicating that practically none of the generated images are of presentable quality.
Enabling autoguidance lowers the FID substantially to \data{\cidBestUncond-fid}, and the improvement in \DINO{} is similarly significant.

\subsection{Ablations} 
\label{sec:ablation}

Table~\ref{tabMainResults} further shows that it is beneficial to allow independent EMA lengths for the main and guiding models.
When both are forced to use the same EMA, FID worsens from \data{\cidBestS-fid} to \data{\cidBestS-sameEMA-fid} in ImageNet-512 (EDM2-S).
We also measure the effect of each degradation (reduced training time, capacity) in isolation.
If we set the guiding model to the same capacity as the main model and only train it for a shorter time, FID worsens to \data{img512-S-S-128Mi-fid}.
If we instead train the reduced-capacity guiding model for as long as the main model, FID suffers a lot more, to \data{img512-S-XS-fid}.
We can thus conclude that both degradations are beneficial and orthogonal, but a majority of the improvement comes from reduced training of the guiding model.
Notably, all these ablations still outperform standard CFG in terms of FID.

Figure~\ref{figMainPlots} probes the sensitivity to various hyperparameters.
Our best result is obtained by training the guiding model $1/16$\textsuperscript{th} as much as the main model, in terms of images shown during training.
Further halving the training budget is almost equally good, while doubling the amount of training starts to slowly compromise the results.
The results are quite insensitive to the choice of the guidance weight.
In terms of the capacity of the guiding model, one step smaller (XS for EDM2-S) gave the best result.
Two steps smaller (XXS) was also better than no capacity reduction (S), but started to show excessive sensitivity to the guidance weight.
The results are also sensitive to the EMA length, similarly to the original EDM2.
Post-hoc EMA~\cite{Karras2024edm2} allows us to search the optimal parameters at a feasible cost.

We also explored several other degradations for the guiding model but did not find them to be beneficial.
First, we tried reducing the amount of training data used for the guiding model,
  but this did not seem to improve the results over the baseline.
Second, applying guidance interval \cite{Kynkaanniemi2024guidance} on top of our method reduced its benefits to some extent,
  suggesting that autoguidance is helpful at all noise levels.
Third, deriving the guiding model from the main model using synthetic degradations did not work at all,
  providing further evidence that the guiding model needs to exhibit the same kinds of degradations that
  the main model suffers from.
Fourth, we found that if the main model had been quantized, e.g., to improve inference speed,
  quantizing it to an even lower precision did not yield a useful guiding model.

One limitation of autoguidance is the need to train a separate guiding model. 
That said, the additional training cost can be quite modest when using a smaller model and shorter training time for the guiding model.
For example, the EDM2-M model trains approximately $2.7\times$ as fast as EDM2-XXL per iteration, and we train it for 1/3.5 of iterations, so the additional cost is around +11\%.
For the EDM2-S/XS pair used in most of our experiments, the added training cost is only +3.6\%.

\vspace{-0.5mm}
\subsection{Qualitative results}
\vspace{-0.5mm}
\figImageNetMain

Figure~\ref{figImageNetMain} shows examples of generated images for ImageNet-512.
Both CFG and our method tend to improve the perceptual quality of images,
  guiding the results towards clearer realizations as the guidance weight increases.
However, CFG seems to have a tendency to head towards a more limited number of
  canonical images~\cite{Kynkaanniemi2024guidance} per class, while
  our method produces a wider gamut of image compositions.
An example is the atypical image of a \emph{Palace} at \mbox{$\guidance=1$},
  which CFG converts to a somewhat idealized depiction as $\guidance$ increases.
Sometimes the unguided sample contains incompatible elements of multiple possible images,
  such as the \emph{Castle} image, which includes a rough sketch of two or three castles of unrelated styles.
In this instance, CFG apparently struggles to decide what to do,
  whereas our method first builds the large red element into a castle,
  and with increased guidance focuses on the red foreground object.
A higher number of possible output images is consistent with a lower FID, implying better
  coverage of the training data.

\figDeepFloydMain

In order to study our method in the context of large-scale image generators, we apply it to DeepFloyd IF \cite{DeepFloyd}.
We choose this baseline because multiple differently-sized models are publicly available. Ideally we could have also used an earlier snapshot as the guiding model, but those were not available.
DeepFloyd IF generates images as a cascade of three diffusion models: a base model and two super-resolution stages.
We apply our method to the base model only, while the subsequent stages always use CFG.
Figure~\ref{figDeepFloydMain} demonstrates the effect of CFG, our method, and their various combinations.
To combine autoguidance with CFG, we extend Equation 3 to cover multiple guiding models as
  proposed by Liu et al.~\cite{Liu2022} and distribute the total guidance weight among them using linear
  interpolation (see Appendix~\ref{app:deepFloydDetails} for details).
While CFG improves the image quality significantly, it also simplifies the style and layout of the image towards a canonical depiction.
Our method similarly improves the image quality, but it better preserves the image's style and visual complexity.
We hope that using both guiding methods \mbox{simultaneously will serve as a new, useful artistic control.}

\vspace*{-1mm}
\section{Discussion and Future work}
\label{sec:discussion}
\vspace*{-1mm}

We have shown that classifier-free guidance entangles several phenomena together,
  and that a different perspective together with simple practical changes opens up an entire new design space.
In addition to removing the superfluous connection to conditioning, this enables significantly better results.

Potential directions for future work include formally proving the conditions that allow autoguidance to be beneficial,
  and deriving good rules of thumb for selecting the best guiding model.
Our suggestion\,---\,an early snapshot of a smaller model\,---\,is easy to satisfy in principle,
  but these are not available for current large-scale image generators in practice.
Such generators are also often trained in successive stages where the training data may change at some point,
  causing potential distribution shifts between snapshots
  that would violate our assumptions.
Various modifications to guidance \cite{Ahn2024, Hong2024, Hong2023} can be seen as inducing degradations through perturbation of attention maps or denoiser inputs.
Whether these approaches could provide additional benefits in our setup remains an open question.
Autoguidance also bears conceptual similarity to contrastive decoding \cite{Li2023} used in large language models to reduce the repetitiveness of generations, and there may be opportunities for sharing observations between the two domains.

Recently, several studies \cite{Chang2023DynGuidance,Gao2023MaskedDiT,Kynkaanniemi2024guidance,Sadat2024DynCFG,wang2024} have reduced the downsides of CFG by making the guidance weight noise level-dependent.
A key benefit from these schedules appears to be the suppression of CFG at high noise levels, where its image quality benefit is overshadowed by the undesirable reduction in variation that is caused by large differences in the content of the differently conditioned distributions. In contrast, autoguidance is not expected to suffer from this problem at high noise levels, as both models target the same distribution. 
So far we have compared autoguidance only with the interval method \cite{Kynkaanniemi2024guidance}, which we did not find beneficial in combination.
A further study on the various possible combinations, in terms of quantitative performance as well as artistic control, is a natural next step. 
It could also be interesting to further isolate the origin of the improvement using alternative metrics, such as precision and recall \cite{Kynkaanniemi2019PR}, Human Preference Score~\cite{wu2023}, or PickScore~\cite{kirstain2023}.

\if@submission\else
\subsection*{Acknowledgments}
We thank David Luebke, Janne Hellsten, Ming-Yu Liu, and Alex Keller for discussions and comments, and
Tero Kuosmanen and Samuel Klenberg for maintaining our compute infrastructure.
\fi

{\small
  \bibliographystyle{ieee}
  \bibliography{paper}
}

\newpage
\appendix
{\LARGE\bf Appendices}
\vspace{-2mm}
\section{Additional results}
\vspace{-2mm}
\label{app:deepFloydExamples}

Figure~\ref{figDeepFloydAppendix} shows additional results using DeepFloyd IF, similar to Figure~\ref{figDeepFloydMain}.
\figDeepFloydAppendix

Figure~\ref{figBonusGuidance} plots FID and \DINO{} as functions of guidance weight, showing that autoguidance is less sensitive to the exact choice of $\guidance$ than CFG.
Figure~\ref{figVariationComparison} shows example grids that demonstrate that autoguidance retains much higher variation than CFG.
Figure~\ref{figToyProgress} provides additional visualizations from our toy example, showing how the implied densities evolve during inference with CFG and autoguidance.

\figBonusGuidance
\figVariationComparison
\figToyProgress\afterpage{\clearpage}

\section{Implementation details}
\label{app:repro}
\label{app:newModels}

We performed our main experiments on top of the publicly available EDM2~\cite{Karras2024edm2} codebase%
  \footnote{\url{https://github.com/NVlabs/edm2}}
  using NVIDIA A100 GPUs, Python 3.11.7, PyTorch 2.2.0, CUDA 11.8, and CuDNN 8.9.7.
Since our method only involves using a different guiding model during sampling, we were able to perform all measurements using the existing command-line scripts.
We only had to modify one line of code in the sampling loop to pass in the class label to the guiding network in addition to the main network.
Algorithm~\ref{algReproEvalAppendix} demonstrates the steps needed to reproduce one of our results from Table~\ref{tabMainResults};
  the rest can be reproduced by repeating the same steps with different models and hyperparameters as indicated in the table.
For the four cases where a pre-trained model was unavailable, we trained the models ourselves as detailed in Algorithm~\ref{algReproTrainAppendix}.

\subsection{Hyperparameter search}
\label{app:gridSearch}
\label{app:compute}

We optimized the autoguidance parameters for each configuration in Table~\ref{tabMainResults} and Figure~\ref{figMainPlots} using automated grid search.
We performed the search separately for FID and \DINO{} across the space of five parameters:
\begin{itemize}
\item \textbf{Model capacity:} All available model capacities (EDM2-XXS, EDM2-XS, etc.) up to and including the capacity of the main model.
\item \textbf{Training time:} Number of training images in powers of two. We found that the sweet spot was always within the range $\{ 2^{26}, 2^{27}, 2^{28}, 2^{29} \}$\,---\,there was no need to go for lower or higher values.
\item \textbf{Guidance weight:} All values within $[1.00, 3.50]$ at regular intervals of $0.05$.
\item \textbf{EMA lengths:} All values within $[0.010, 0.250]$ at regular intervals of $0.005$. We treated the EMA length of the main model and the guiding model as two separate parameters.
\end{itemize}
In order to reduce the computational workload, we pruned the search space by considering only a local neighborhood of parameters around the best FID or \DINO{} found thus far.
Whenever the result improved, we placed a new local grid around the corresponding parameters, resulting in gradual convergence towards the global optimum.
Once we reached the optimum, we further re-evaluated the nearby parameter choices two more times to account for the effect of random noise.
As such, each result reported in Table~\ref{tabMainResults} and Figure~\ref{figMainPlots} represents the best of three evaluations.
The typical range of random variation is indicated by the shaded regions in Figure~\ref{figMainPlots}.

In total, the grid search resulted in roughly 30,000 metric evaluations across all of our configurations.
In each metric evaluation, the main cost comes from generating 50,000 random images, which takes around 30--60 minutes using eight A100 GPUs and consumes approximately 2--5 kWh of energy, depending on model size.
The overall energy consumption of our entire project was thus in the ballpark of 60--150 MWh.

\subsection{DeepFloyd IF experiments}
\label{app:deepFloydDetails}

In order to apply CFG and autoguidance simultaneously, we extend Equation~\ref{eq:generalized} to cover multiple guiding models as proposed by Liu~et~al.~\cite{Liu2022}.
In this case, we have three models: the main model $D_\text{m}$, an unconditional model $D_\text{u}$, and a reduced-capacity conditional model $D_\text{c}$.
The guided denoising result is then defined as
\begin{align}
  \dguid(\boldx; \sigma, \boldc) &\,\coloneqq\, D_\text{m}(\boldx; \sigma, \boldc) \,+ \!\!\sum_{i \in \{\text{u}, \text{c}\}}\! (\guidance_i - 1) \big( D_\text{m}(\boldx; \sigma, \boldc) -  D_i(\boldx; \sigma, \boldc) \big)
  \text{,}
\end{align}
where $\guidance_\text{u}$ and $\guidance_\text{c}$ correspond to the guidance weights for CFG and autoguidance, respectively.
To interpolate between the two methods, we further define $\guidance_\text{u} \coloneqq (1 - \alpha) (\guidance - 1) + 1$ and \mbox{$\guidance_\text{c} \coloneqq \alpha (\guidance - 1) + 1$},
  where $\guidance$ indicates the desired total amount of guidance and $\alpha \in [0, 1]$ is a linear interpolation factor.

DeepFloyd IF~\cite{DeepFloyd} uses a three-stage cascade: a base model followed by two super-resolution stages. We apply our method only to the base model, while the super-resolution stages always use CFG with their default weights (4 and 9).
We use their Stochastic DDPM sampler with default settings: 100, 75, 50 steps for the base model and subsequent super-resolution stages, respectively. 
Dynamic thresholding \cite{imagen} is used for all stages.

\algReproEvalAppendix

\section{Details of the 2D toy example}
\label{app:toyDetails}

In this section, we describe the construction of the 2D toy dataset used in the analysis of Section~\ref{sec:whycfgimprove},
  as well as the associated model architecture, training setup, and sampling parameters.
The related code is available at {\footnotesize\url{https://github.com/NVlabs/edm2}}

\paragraph{Dataset.}

For each of the two classes $\boldc$, we model the fractal-like data distribution as a mixture of Gaussians
  \mbox{$\mathcal{M}_\boldc = \big( \{\phi_i\}, \{\mathbf{\mu}_i\}, \{\mathbf{\Sigma}_i\} \big)$},
  where $\phi_i$, $\mathbf{\mu}_i$, and $\mathbf{\Sigma}_i$ represent the weight, mean, and $2{\times}2$ covariance matrix of each component $i$, respectively.
This lets us calculate the ground truth scores and probability densities analytically and, consequently, to visualize them without making any additional assumptions.
The probability density for a given class is given by
\begin{align}
  p_\text{data}(\boldx | \boldc) &\,=\, \sum_{i \in \mathcal{M}_\boldc} \!\phi_i \,\NN(\boldx; \mathbf{\mu}_i, \mathbf{\Sigma}_i)
  \text{,} \hs{3} \text{where} \\
  \NN(\boldx; \mathbf{\mu}, \mathbf{\Sigma}) &\,=\, \frac{1}{\sqrt{(2 \pi)^2 \det(\mathbf{\Sigma})}} \exp \bigg( \!-\frac{1}{2} (\boldx - \mathbf{\mu})^\top \mathbf{\Sigma}^{-1} (\boldx - \mathbf{\mu}) \bigg)
  \text{.}
\end{align}
Applying heat diffusion to $p_\text{data}(\boldx | \boldc)$, we obtain a sequence of increasingly smoothed densities $p(\boldx | \boldc; \sigma)$ parameterized by noise level $\sigma$:
\begin{align}
  p(\boldx | \boldc; \sigma) &\,=\, \sum_{i \in \mathcal{M}_\boldc} \!\phi_i \,\NN\big( \boldx; \mathbf{\mu}_i, \mathbf{\Sigma}^\ast_{i,\sigma} \big) \label{eq:GT}
  \text{,} \hs{3} \text{where} \hs{2}
  \mathbf{\Sigma}^\ast_{i,\sigma} =\, \mathbf{\Sigma}_i + \sigma^2 \boldI
  \text{.}
\end{align}
The score function of $p(\boldx | \boldc; \sigma)$ is then given by
\begin{align}
  \nablax \log p(\boldx | \boldc; \sigma) &\,=\, \frac{
    \sum_{i \in \mathcal{M}_\boldc} \phi_i \,\NN\big( \boldx; \mathbf{\mu}_i, \mathbf{\Sigma}^\ast_{i,\sigma} \big) \,\big( \mathbf{\Sigma}^\ast_{i,\sigma} \big)^{-1} \big( \mathbf{\mu}_i - \boldx \big)}{
    \sum_{i \in \mathcal{M}_\boldc} \phi_i \,\NN\big( \boldx; \mathbf{\mu}_i, \mathbf{\Sigma}^\ast_{i,\sigma} \big)}
  \text{.}
\end{align}

\algReproTrainAppendix

We construct $\mathcal{M}_\boldc$ to represent a thin tree-like structure by starting with one main ``branch'' and recursively subdividing it into smaller ones.
Each branch is represented by 8 anisotropic Gaussian components and the subdivision is performed 6 times, decaying $\phi$ after each subdivision and slightly randomizing the lengths and orientations of the two resulting sub-branches.
This yields \mbox{$127 {\times} 8 = 1016$} components per class and $1016 {\times} 2 = 2032$ components in total.
We define the coordinate system so that the mean and standard deviation of $p_\text{data}$, marginalized over $\boldc$,
  are equal to~$0$ and $\sigma_\text{data} = 0.5$ along each axis, respectively, matching the recommendations by Karras~et~al.~\cite{Karras2022elucidating}.

\paragraph{Models.}

We implement the denoiser models $D_0$ and $D_1$ as simple multi-layer perceptrons, utilizing the magnitude-preserving design principles from EDM2~\cite{Karras2024edm2}.
To be able to visualize the implied probability densities in Figure~\ref{figToyDetail}, we design the model interface so that for a given noisy sample,
  each model outputs a single scalar representing the logarithm of the corresponding unnormalized probability density,
  as opposed to directly outputting the denoised sample or the score vector.
Concretely, let us denote the output of a given model by $G_\theta(\boldx; \sigma, \boldc)$.
The corresponding normalized probability density is then given by
\begin{align}
  p_\theta(\boldx | \boldc; \sigma) &\,=\, \exp\big( G_\theta(\boldx; \sigma, \boldc) \big) \,\Big/ \int \exp\big( G_\theta(\boldx; \sigma, \boldc) \big) \,\diff\boldx
  \text{.}
\end{align}
By virtue of defining $G_\theta$ this way, we can derive the score vector, and by extension, the denoised sample, from $G_\theta$ through automatic differentiation:
\begin{align}
  \nablax \log p_\theta(\boldx | \boldc; \sigma) &\,=\, \nablax \,G_\theta(\boldx; \sigma, \boldc) \label{eq:EBM} \\[1mm]
  D_\theta(\boldx; \sigma, \boldc) &\,=\, \boldx + \sigma^2 \nablax \,G_\theta(\boldx; \sigma, \boldc)
  \text{.}
\end{align}
Besides Equation~\ref{eq:EBM}, we also tried out the alternative formulations where the model outputs the score vector or the denoised sample directly.
The results produced by all these variants were qualitatively more or less identical; we chose to go with the formulation above purely for convenience.

To connect the above definition of $G_\theta$ to the raw network layers, we apply preconditioning using the same general principles as in EDM~\cite{Karras2022elucidating}.
Denoting the function represented by the raw network layers as $F_{\theta}$, we define $G_\theta$ as
\begin{align}
  G_\theta(\boldx; \sigma, \boldc) &\,=\, -\frac{1}{2} \norm{\boldx^\ast}_2^2
    - \frac{g_\theta}{\sigma n} \,\sum_{i=1}^n F_{\theta,i}\Big(\boldx^\ast; \,\frac{1}{4} \log \sigma, \,\boldc\Big)^2 \label{eq:precond}
  \text{,} \hs{3} \text{where} \hs{2}
  \boldx^\ast \,=\, \frac{\boldx}{\sqrt{\sigma^2 + \sigma_\text{data}^2}}
\end{align}
and the sum is taken over the $n$ output features of $F_\theta$.
We scale the output of $F_\theta$ by a learned scaling factor $g_\theta$ that we initialize to zero.

The goal of Equation~\ref{eq:precond} is to satisfy the following three requirements:
\begin{itemize}
\item The input of $F_\theta$ should have zero mean and unit magnitude. This is achieved through the division by $\sqrt{\sigma^2 + \sigma_\text{data}^2}$.
\item After initialization, $G_\theta$ should represent the best possible first-order approximation of the correct solution. This is achieved through the $-\tfrac{1}{2}\norm{\boldx^\ast}_2^2$ term, as well as the fact that $g_\theta = 0$ after initialization.
\item After training, $\sqrt{g_\theta} \cdot F_\theta$ should have approximately unit magnitude. This is achieved through the division by $\sigma n$.
\end{itemize}

In practice, we use an MLP with one input layer and four hidden layers, interspersed with SiLU~\cite{Hendrycks2016} activation functions and implemented using the magnitude-preserving primitives from EDM2~\cite{Karras2024edm2}.
The input is a 4-dimensional vector $\big[ \boldx^\ast_x; \boldx^\ast_y; \tfrac{1}{4} \log \sigma; 1 \big]$ and the output of each hidden layer has $n$ features, where $n = 64$ for $D_1$ and $32$ for $D_0$.

\paragraph{Training.}

Given that we have the exact score function of the ground truth distribution readily available (Equation~\ref{eq:GT}), we train the models using exact score matching~\cite{Hyvarinen05} for simplicity and increased robustness.
We thus define the loss function as
\begin{align}
  \mathcal{L}(\theta) &\,=\, \mathbb{E}_{\sigma \sim p_\text{train}, \boldx \sim p(\boldx; \,\sigma)} \,\sigma^2 \big\lVert \nablax \log p_\theta(\boldx; \sigma) - \nablax \log p(\boldx; \sigma) \big\rVert^2_2
  \text{,}
\end{align}
where $\sigma \sim p_\text{train}$ is realized as $\log(\sigma) \sim \mathcal{N}(P_\text{mean}, P_\text{std})$ \cite{Karras2022elucidating}.
As an alternative to exact score matching, we also experimented with the more commonly used denoising score matching, but did not observe any noticeable differences in model behavior or training dynamics.

We train $D_1$ for 4096 iterations using a batch size of 4096 samples, and $D_0$ for 512 iterations. We set
  $P_\text{mean} \!= -2.3$ and $P_\text{std} \!= 1.5$,
  and use $\alpha_\text{ref} / \sqrt{\max(t / t_\text{ref}, 1)}$ learning rate decay schedule with $\alpha_\text{ref} \!= 0.01$ and $t_\text{ref} \!= 512$ iterations,
  along with a power function EMA profile~\cite{Karras2024edm2} with $\sigma_\text{rel} \!= 0.010$.
Overall, the setup is robust with respect to the hyperparameters;
  the phenomena illustrated in Figures~\ref{figToyExample} and~\ref{figToyDetail} remain unchanged across a wide range of parameter choices.

\paragraph{Sampling.}

We use the standard EDM sampler~\cite{Karras2022elucidating} with \mbox{$N = 32$} Heun steps (\mbox{$\text{NFE} = 63$}), \mbox{$\sigma_\text{min} \!= 0.002$}, \mbox{$\sigma_\text{max} \!= 5$}, and \mbox{$\rho = 7$}.
We chose the values of $N$ and $\sigma_\text{max}$ to be much higher than what is actually needed for this dataset in order to avoid potential discretization errors from affecting our conclusions.
In Figure~\ref{figToyExample}, we set \mbox{$w = 4$} for CFG and \mbox{$w = 3$} for autoguidance, and multiply the score vectors (Equation~\ref{eq:EBM}) by $1.40$ for naive truncation.
In Figure~\ref{figToyDetail}, we set \mbox{$w = 4$} and \mbox{$\sigma_\text{mid} \!= 0.03$}.

\section{Broader societal impact}
\label{app:broaderImpact}
Generative modeling, including images and videos, has significant misuse potential.
It can trigger negative consequences within the society in several ways.
The primary concerns include various types of disinformation, but also the potential to amplify stereotypes and unwanted biases~\cite{Mishkin2022risks}.
Our improvements to the sample quality can make the results even more believable, even when used for disinformation.
That said, we do not unlock any novel uses of the technology.

\section{Licenses}
\label{app:licenses}

\begin{itemize}
\item \makebox[50mm][l]{EDM2 models \cite{Karras2024edm2}:}                           Creative Commons BY-NC-SA 4.0 license
\item \makebox[50mm][l]{DeepFloyd IF models \cite{DeepFloyd}:}                        Modified MIT license
\item \makebox[50mm][l]{Stable Diffusion VAE model \cite{Rombach2021highresolution}:} CreativeML Open RAIL++-M license
\item \makebox[50mm][l]{InceptionV3 model \cite{Szegedy2016inception}:}               Apache 2.0 license
\item \makebox[50mm][l]{DINOv2 model \cite{oquab2023dinov2}:}                         Apache 2.0 license
\item \makebox[50mm][l]{ImageNet dataset \cite{Deng2009imagenet}:}                    Custom non-commercial license
\end{itemize}

\end{document}